\documentclass{article}

    \usepackage[preprint]{neurips_2021}

\usepackage{microtype}
\usepackage{graphicx}
\usepackage{booktabs} 
\usepackage{multirow}
\usepackage{float}

\usepackage{hyperref}

\usepackage{amsmath,amsfonts,bm}

\def\eqref#1{equation~\ref{#1}}

\def\1{\bm{1}}

\DeclareMathAlphabet{\mathsfit}{\encodingdefault}{\sfdefault}{m}{sl}
\SetMathAlphabet{\mathsfit}{bold}{\encodingdefault}{\sfdefault}{bx}{n}

\usepackage{hyperref}
\usepackage{url}
\usepackage[utf8]{inputenc} 
\usepackage[T1]{fontenc}    
\usepackage{url}            
\usepackage{booktabs}       
\usepackage{amsmath, amsfonts,amsthm,amssymb}     
\usepackage{nicefrac}       
\usepackage{microtype}      
\usepackage{caption}
\usepackage{amssymb}
\usepackage{adjustbox}
\usepackage{mathtools}

\usepackage{graphicx}
\usepackage{xargs}  
\usepackage[colorinlistoftodos,prependcaption, textwidth=3cm]{todonotes}

\usepackage{booktabs}
\usepackage{wrapfig}
\usepackage{needspace}
\newcommand{\bbbbbb}[1]{{\color{green} { #1} --
\textsc{bbbb}}}

\usepackage{xspace}
\newcommand{\simgan}{\textsc{MS-VAE}\xspace}
\newcommand{\simgangan}{\textsc{MS-GAN}\xspace}
\newcommand{\simganflow}{\textsc{MS-FLOW}\xspace}
\newcommand{\DR}{DR\xspace}

\usepackage{subfig}
\usepackage[utf8]{inputenc}
\usepackage{pgfplots}
\pgfplotsset{compat=newest}
\usepgfplotslibrary{groupplots}
\usepgfplotslibrary{dateplot}
\usepackage{afterpage}
\usepackage[compact]{titlesec}

\usepackage{tikz}
\usetikzlibrary{bayesnet}

\usepackage{pgfplots}
\pgfplotsset{compat=newest}
\usepgfplotslibrary{groupplots}
\usepgfplotslibrary{dateplot}

\title{Improving the Reconstruction of Disentangled Representation Learners via Multi-Stage Modeling}

\author{
  Akash Srivastava, Yamini Bansal,  Yukun Ding, Cole Lincoln Hurwitz \\ Equal Contribution \\
  MIT-IBM Watson AI Lab\\
  Cambridge, MA \\
  \\
  \And
  Kai Xu, Bernhard Egger, Prasanna Sattigeri, Joshua B. Tenenbaum, Kristjan Greenewald, \\
  \textbf{David D. Cox, Dan Gutfreund} \\
  MIT-IBM Watson AI Lab, IBM Research, MIT\\
  Cambridge, MA \\
  \\
   \And
   Agus Sudjianto,
    Phuong Le, Arun Prakash R, Nengfeng Zhou, Joel Vaughan; Yaqun Wang,  \\ \textbf{Anwesha Bhattacharyya } \thanks{ The authors from Wells Fargo express their gratitude to Yueyang Shen and Maorong Rao for their valuable contributions to the discussion. Appreciation is also extended to the corporate risk - model risk team at Wells Fargo for their support in this project. The views expressed in the paper are those of the authors and do not represent the views of Wells Fargo. }   \\
   Wells Fargo, US \\
}

\begin{document}
\maketitle
\begin{abstract}
Current autoencoder-based disentangled representation learning methods achieve disentanglement by penalizing the (aggregate) posterior to encourage statistical independence of the latent factors. This approach introduces a trade-off between disentangled representation learning and reconstruction quality since the model does not have enough capacity to learn correlated latent variables that capture detail information present in most image data. To overcome this trade-off, we present a novel multi-stage modeling approach where the disentangled factors are first learned using a penalty-based disentangled representation learning method; then, the low-quality reconstruction is improved with another deep generative model that is trained to model the missing correlated latent variables, adding detail information while maintaining conditioning on the previously learned disentangled factors. Taken together, our multi-stage modelling approach results in a single, coherent probabilistic model that is theoretically justified by the principal of D-separation and can be realized with a variety of model classes including likelihood-based models such as variational autoencoders, implicit models such as generative adversarial networks, and tractable models like normalizing flows or mixtures of Gaussians. We demonstrate that our multi-stage model has higher reconstruction quality than current state-of-the-art methods with equivalent disentanglement performance across multiple standard benchmarks. In addition, we apply the multi-stage model to generate synthetic tabular datasets, showcasing an enhanced performance over benchmark models across a variety of metrics. The interpretability analysis further indicates that the multi-stage model can effectively uncover distinct and meaningful features of variations from which the original distribution can be recovered. 
\end{abstract}

\section{Introduction}
\label{sec:intro}

Deep generative models (DGMs) such as variational autoencoders (VAEs) \citep{vae, rezende2014stochastic} and generative adversarial networks (GANs) \citep{gan} have recently enjoyed great success at modeling high dimensional data such as natural images. As the name suggests, DGMs leverage deep learning to model a data generating process. The underlying assumption is that the high dimensional observations $X \in \mathbb{R}^D$ can be meaningfully described by a small set of latent factors $H \in \mathbb{R}^K$, where $K < D$. More precisely, the observation $(X=x)$
is assumed to be generated by first sampling a set of low dimensional factors $h$ from a simple prior distribution $p(H)$ and then sampling $x \sim p_\theta(X|h)$. DGMs realize $p_\theta$ through a deep neural network also known as the decoder or the generative network. VAE-based DGMs use another deep neural network (called the encoder or the inference network) to parameterize an approximate posterior $q_\phi(H|x)$. Learning the variational posterior parameters is done by maximizing an evidence lower bound (ELBO) to the log-marginal likelihood of the data under the model $p_\theta(X)$. 

Learning \textit{disentangled} factors $h \sim q_\phi(H|x)$ that are semantically meaningful representations of the observation $x$ is highly desirable because such interpretable representations can arguably be advantageous for a variety of downstream tasks \citep{icmlbest}, including classification, detection, reinforcement learning, transfer learning and image synthesis from textual descriptions \citep{bengio2013representation, lecun2015deep, lake2017building, van2019disentangled, DBLP:conf/icml/ReedAYLSL16, DBLP:journals/corr/ZhangXLZHWM16}. While a formal definition of disentangled representation (\DR) remains elusive, it is understood to mean that by manipulating only one of the factors while holding the rest constant, only one semantically meaningful aspect of the observation (e.g. the pose of an object in an image) changes. 
Prior work in unsupervised \DR learning focuses on the objective of learning statistically independent latent factors as means for obtaining \DR. The underlying assumption is that the latent variables $H$ can be partitioned into independent components $C$ (i.e. the disentangled factors) and correlated components $Z$. An observation $(X=x)$
is assumed to be generated by sampling the low dimensional factors $h=(c,z)$ from $p(H)$ and then sampling $x \sim p_\theta(X|c,z)$ (Figure \ref{fig:cgan}). A series of works starting from \citep{beta} enforce statistical independence of the latent factors $H$ via regularization, up-weighting certain terms in the ELBO which penalize the (aggregate) posterior to be factorized over all or some of the latent dimensions \citep{kumar2017variational,factor,mig} (see Section \ref{sec:related} for details).

While the aforementioned models show promising results, they suffer from a trade-off between \DR learning and reconstruction quality. If the latent space is heavily regularized -- not allowing enough capacity for the correlated variables -- then the reconstruction quality will be diminished, signaling that the learned representation is sub-optimal. As the correlated variables are functionally ignored with high levels of regularization, an observation $(X=x)$ can be thought to be generated by only sampling independent latent factors $c$ from $p(C)$ and then sampling $x \sim p_\theta(X|c)$ (Figure \ref{fig:btcvae}). This is a departure from the original data generating hypothesis that $x$ is sampled from a distribution dependent on both the independent and correlated latent variables. On the other hand, if the correlated variables are not well-constrained, the model can use them to achieve a high quality reconstruction while ignoring the independent variables  (the disentangled latent factors). This phenomena is referred to as the "shortcut problem" and has been discussed in previous works \citep{DBLP:conf/iclr/SzaboHPZF18,lezama2018overcoming}. Overcoming the aforementioned trade-off is essential for using these models in real world applications such as realistic, controlled image synthesis \citep{lee2020high,lezama2018overcoming}.

\begin{figure}[t] 
\centering
\subfloat[Target]{
    {%
    \setlength{\fboxsep}{-0.0pt}%
    \setlength{\fboxrule}{0.3pt}%
    \fbox{\includegraphics[width=0.2\linewidth]{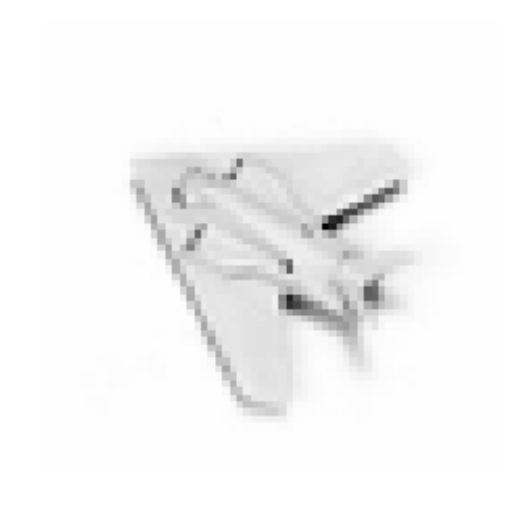}}%
    }%
    \label{fig:panel-target}
}
\subfloat[$\beta$-TCVAE]{
    {%
    \setlength{\fboxsep}{-0.0pt}%
    \setlength{\fboxrule}{0.3pt}%
    \fbox{\includegraphics[width=0.2\linewidth]{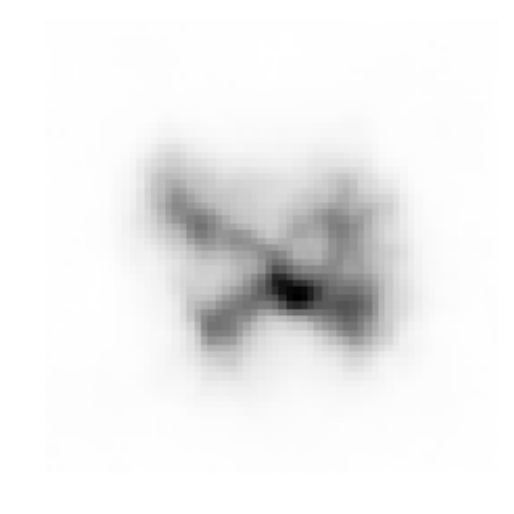}}%
    }%
    \label{fig:panel-tcvae}
}
\subfloat[\simgan]{
    {%
    \setlength{\fboxsep}{-0.0pt}%
    \setlength{\fboxrule}{0.3pt}%
    \fbox{\includegraphics[width=0.2\linewidth]{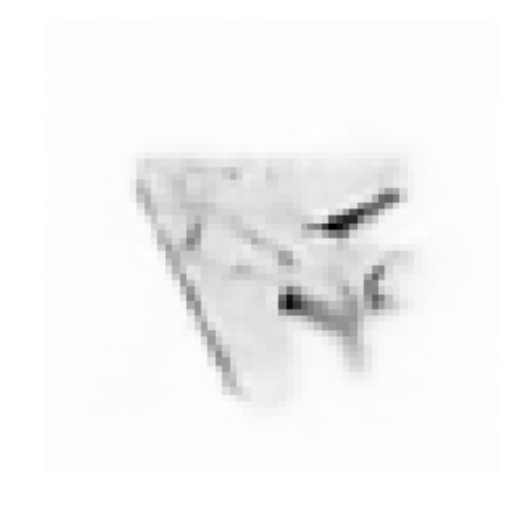}}%
    }%
    \label{fig:panel-dsvae}
}
\caption{
Image reconstruction using $\beta$-TCVAE (Figure~\ref{fig:panel-tcvae}) and \simgan (Figure~\ref{fig:panel-dsvae}). \simgan is able to take the blurry output of the underlying $\beta$-TCVAE model and learn to render a much better approximation of the target while maintaining the pose of the original image (Figure~\ref{fig:panel-target}). 
}
\label{fig:three}
\end{figure}

In this paper, we propose a new graphical model for \DR learning (Figure \ref{fig:ds_vae}) that allows for learning disentangled factors while also correctly realizing the data generating hypothesis that an observation is generated from independent and correlated factors. Importantly, the graphical model in Figure \ref{fig:ds_vae} is D-separated, meaning that any changes in the correlated latent variables $Z$ will not influence the independent latent variables $C$. Generating an observation $(X=x)$ from this model can then be done by sampling the independent factors $c$ from $p(C)$, sampling a low-quality reconstruction $y \sim p_\theta(Y|c)$, sampling the correlated factors $z$ from $p(Z)$, and then finally sampling $x \sim p_\theta(X|z,Y)$. The final reconstruction $x$ depends both on $z$ and $c$, however, any regularization needed to extract the independent factors $c$ no longer diminishes the model capacity for the correlated factors $z$. 

To realize our proposed graphical model, we introduce \textbf{\simgan}, a multi-stage DGM that is implemented as follows: first, the disentangled representation $C$ is learned using an existing \DR learning method such as $\beta$-TCVAE \citep{mig}. Since the learned factors $C$ are regularized to be statistically independent -- not allowing enough capacity for correlated factors -- the final reconstruction $Y$ will have diminished reconstruction quality. Then, we train another DGM to improve the low-quality reconstruction $Y$ by learning the missing correlated factors $Z$. This is achieved during training by inputting the reconstruction $Y$ into the decoder of the second DGM and then modulating the hidden activation of each layer as function of latent factor $Z$ (using Feature-wise Linear Modulation \citep{perez2018film}). Through this training paradigm, \simgan is able to preserve conditioning on the disentangled factors while dramatically improving the reconstruction quality. A schematic of \simgan is shown in Figure~\ref{fig:schematic} and example images from each stage are shown in Figure \ref{fig:three}. The reconstruction from $\beta$-TCVAE (\ref{fig:panel-tcvae}) is improved by \simgan (\ref{fig:panel-dsvae}) to better approximate the target (\ref{fig:panel-target}) while maintaining conditioning on the disentangled factors, e.g. azimuth. 

To summarize our contributions:
\begin{itemize}
  \item We propose a new graphical model for \DR learning (Figure \ref{fig:ds_vae}) that, due to the D-separation between its independent and correlated latent variables, alleviates the disentanglement vs. reconstruction trade-off.
  \item We introduce \simgan, a multi-stage DGM that implements our proposed graphical model, achieving state-of-the-art reconstruction quality while maintaining the same level of disentanglement as preexisting methods. \simgan can be implemented using a variety of DGMs such as VAEs, GANs, or FLOW-based models and does not depend on hand-crafted loss functions or additional hyperparameters (unlike current state-of-the-art \citep{lezama2018overcoming})
  \item Following \citep{icmlbest}, we test our framework with a wide-range of qualitative and quantitative tests to demonstrate the efficacy of the framework. We provide all code used for the experiments. 
\end{itemize}

\section{Background}
\label{sec:related}
VAEs, first introduced in \citep{vae}, represent a class of likelihood-based, DGMs comprised of an encoder and a decoder that are trained using amortized variational inference to maximize the expected ELBO under the data distribution,
\begin{align}
\label{eq:elbo}
  &
  \int p(x) \log p_\theta(x) dx \ge \mathbb{E}_{p(x)} \left[ \mathbb{E}_{q_\phi(h|x)}[\log p_\theta(x|h)] - \mathrm{KL}[q_{\phi}(h|x)\Vert p(h)] \right] 
\end{align}
The first term is the expected reconstruction loss under the variational posterior
$q_\phi$. The second term -- the Kullback–Leibler (KL) divergence -- underpins most of the recent \DR learning methods. The KL term regularizes the variational posterior to be similar (in expectation) to a simple spherical Gaussian prior over $H$.   
\citet{beta} attempts to factorize the variational aggregate posterior by up-weighting this KL-divergence term with a Lagrange multiplier $\beta$. Follow-up works decompose the KL-term in \eqref{eq:elbo} and up-weight only the KL-divergence between the aggregate posterior $q_{\phi_H}(H) = \int_X p(x) q_\phi(H|x)dx$ and the factorized aggregate posterior \citep{factor, mig, jeong2019learning}. \citet{kumar2017variational} simply uses a factorized prior. 
These models introduce a trade-off between unsupervised \DR learning and reconstruction quality. When the latent variables are heavily regularized to be independent, an observation $(X=x)$ can be thought to be generated by sampling independent latent factors $c$ from $p(C)$ and then sampling $x \sim p_\theta(X|c)$ (Figure \ref{fig:btcvae}). This leads to low-quality reconstructions as the correlated factors are not utilized. \citet{lezama2018overcoming} attempts to tackle this issue with a teacher-student learning technique. First, the teacher model is trained to extract disentangled factors at the cost of reconstruction quality (similar to the first stage of our model). Then, a student autoencoder (AE) model with a larger latent dimension size is trained with additional losses that force the Jacobian of the final reconstruction with respect to the disentangled factors to remain the same (a.k.a Jacobian supervision). While this model achieves promising results on MNIST \citep{deng2012mnist}, it suffers from a few drawbacks including having to relearn the disentangled factors in the second stage and having no theoretical guarantees that changes in the correlated factors $Z$ will not affect the independent factors $C$ in the student model. Other works have also been proposed to overcome the trade-off using adversarial training \citep{sup5, DBLP:conf/nips/LampleZUBDR17, DBLP:journals/corr/PerarnauWRA16, DBLP:conf/iclr/SzaboHPZF18, DBLP:conf/cvpr/HuSPFZ18}.   

\section{Method}
\label{sec:simgan}

\subsection{Graphical Model}
\label{sec:d-separate}
\paragraph{Lower Bound.} To alleviate the trade-off between reconstruction and disentanglement, we propose a new graphical model for \DR learning (Figure \ref{fig:ds_vae}). To understand this model, let us assume that the data come as pairs of images $\{y_i,x_i\}_{i=1}^N$.
Let the $x$'s be the ground truth images (generated using both the independent factors $c$ and correlated factors $z$) and let the $y$'s be the approximations of those images generated using only 
$c$. 
Given this paired data, the graphical model depicted in Figure \ref{fig:ds_vae} represents a single coherent model of the data that defines a joint distribution over $Y$ and $X$. The log-marginal likelihood of the observations under this model can be lower-bounded as follows,

\begin{align}
    \label{eq:sim_elbo} 
    &\log p_{\{\phi,\theta,\theta_Z,\Theta\}}(x,y) \geq \\ &- \underbrace{\mathrm{KL}[q_\phi(c|y) \Vert p(c)] + \mathbb{E}_{q_{\phi}}[\log p_\theta(y|c)]}_{\text{(a) }C \rightarrow Y} \nonumber -\underbrace{\mathrm{KL}[q_{\theta_Z}(z|x,y) \Vert p(z)] + \mathbb{E}_{q_{\theta_Z}}[\log p_{\Theta}(x|z,y)] }_{\text{(b) }(Y,Z) \rightarrow X}. \nonumber
\end{align}
In practice, we do not have access to $Y$. However, by learning the disentangled factors $C$ from $X$, we can generate $Y$. Therefore, we first use state-of-the-art \DR learning method to learn the sub-graph $C \rightarrow Y$ by maximizing term (a) in Equation \ref{eq:sim_elbo} to produce $Y$. Then, we train another DGM to learn the sub-graph $(Y,Z) \rightarrow X$ by maximizing term (b) in Equation \ref{eq:sim_elbo}. 
This two-stage training procedure approximates the graphical model from Figure \ref{fig:ds_vae}. 
\paragraph{D-separation.} Now  that we introduced the graphical model from Figure \ref{fig:ds_vae}, it is important to understand the theoretical motivation behind it. Recall that our goal is to take an existing method for learning the disentangled factors C and learn additional residual information in Z without changing the conditioning on C.
In a standard VAE graphical model (Figure \ref{fig:cgan}), this could occur because there is no guarantee that $p(Z,C|X) = p(Z|X)p(C|X)$.
By introducing an observed node Y (i.e. D-Separation) in our graphical model, we ensure that changes in Z will not influence C. This is because the joint distribution $p(Z,C|X,Y)$ can now be broken down as $p(C|Y)p(Z|X,Y)$. We illustrate this property of the graphical model with a toy example (Appendix \ref{sec:illustrative_example}). In practice, realizing the graphical model in Figure \ref{fig:ds_vae} using a deep neural network requires care as we want to ensure that the conditioning on Y is stronger than that of Z so that Z only captures the residual correlated information. Also, simply training a single neural network end-to-end to approximate this graphical model is infeasible as Y will no longer act as an observed node, leading to the entanglement of C and Z. 
We demonstrate this phenomenon by performing an ablation study of our proposed method (Appendix \ref{app:ablation}).

\subsection{Implementation (\simgan)}
\begin{figure}[h!]
    \centering
    \subfloat[VAE]{
        \resizebox{0.2\linewidth}{2.5cm}{
            \begin{tikzpicture}
            
              \node[obs]                               (x) {$X$};
              \node[latent, above=of x, xshift=-1.2cm] (c) {$C$};
              \node[latent, above=of x, xshift=1.2cm]  (z) {$Z$};
            
              \edge {z,c} {x} ; %
            
              \plate {} {(c)(z)(x)} {$N$} ;
            
            \end{tikzpicture}
            \label{fig:cgan}
        }
    }
    \subfloat[$\beta$-TCVAE]{
        \resizebox{0.2\linewidth}{2.5cm}{
            \begin{tikzpicture}
            
              \node[obs]                               (x) {$X$};
              \node[latent, above=of x, xshift=0cm] (c) {$C$};
            
              \edge {c} {x} ; %
            
              \plate[xscale=3]  {} {(c)(x)} {$N$} ;
            
            \end{tikzpicture}
            
        }
        \label{fig:btcvae}
    }
    \subfloat[\simgan]{
    \resizebox{0.2\linewidth}{2.5cm}{%
\begin{tikzpicture}

  \node[obs]                               (x) {$X$};
  \node[obs,    above=of x, yshift=-0.6cm] (y) {$Y$};
  \node[latent, above=of x, xshift=1.2cm]  (z) {$Z$};
  \node[latent, above=of x, xshift=-1.2cm] (c) {$C$};

  \edge {z,y} {x} ; %
  \edge {c}   {y} ; %

  \plate {} {(c)(y)(z)(x)} {$N$} ;

\end{tikzpicture}
        \label{fig:ds_vae}
    }
    }
    \subfloat[\simgan Architecture]{
    \includegraphics[width=.2\linewidth]{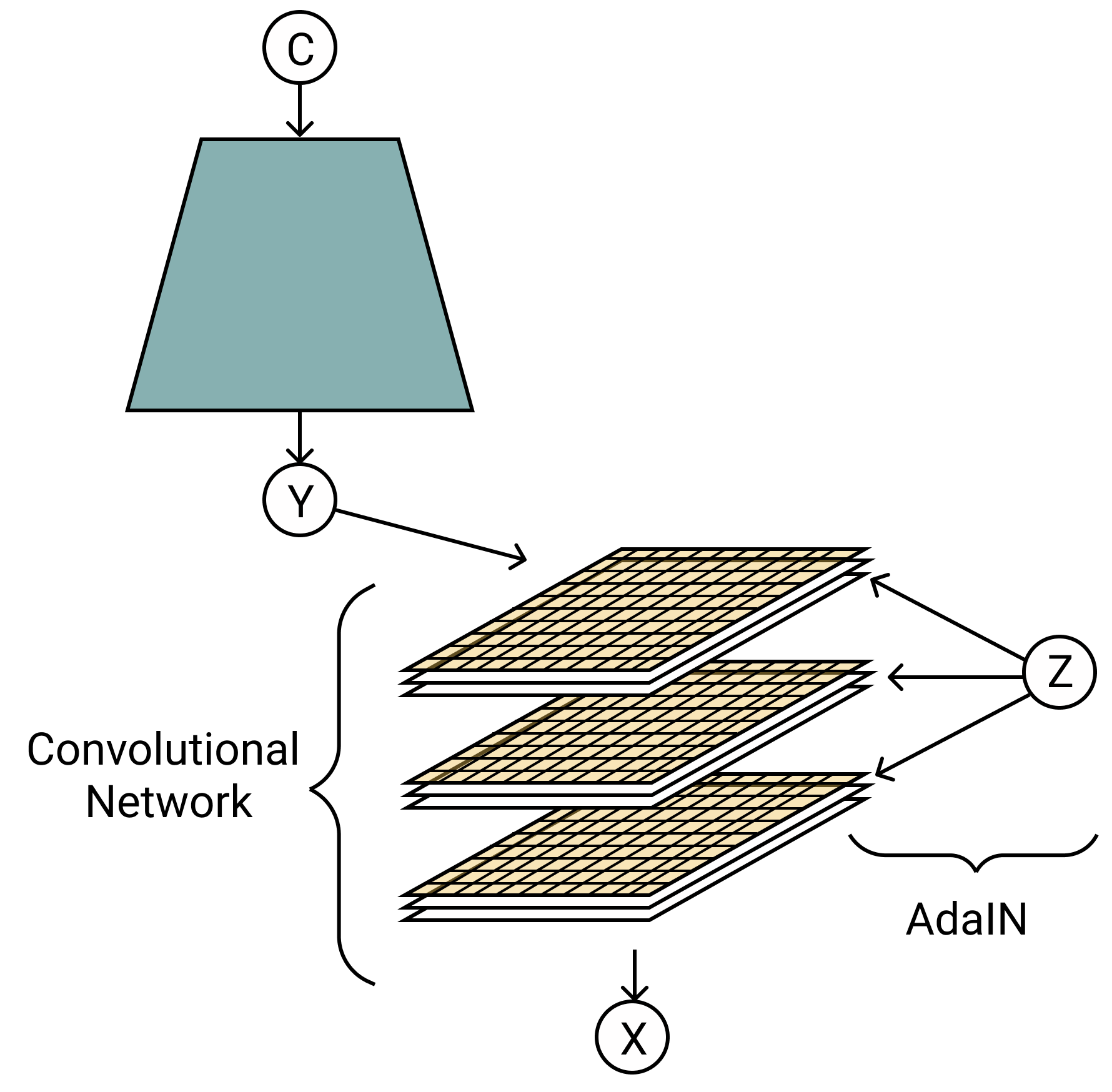}
    \label{fig:schematic}
    }
    \caption{(a) Graphical model of a standard VAE where $C$ and $Z$ are not independent conditioned on $X$. (b) Graphical model of $\beta$-TCVAE where the reconstruction only depends on the independent latent factors $C$. (c) \simgan graphical model where $C$ and $Z$ are independent conditioned on $Y$. (d) Schematic for \simgan when implemented as a convolutional architecture. Both $Y$ and $X$ are the reconstructions of the same image.}\label{fig:simgan_}
\end{figure}
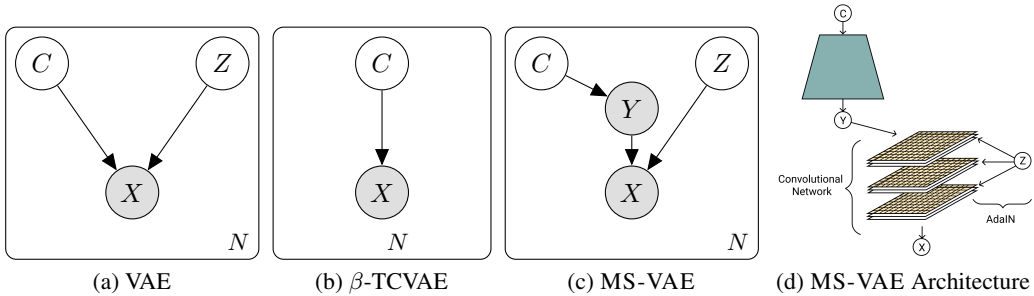

In the following sections, we describe how the graphical model in Figure \ref{fig:ds_vae} can be implemented as variational autoencoder using a multi-stage modelling approach, i.e. \simgan.
\label{sec:implement}
\paragraph{Stage one: Learning a Disentangled Representation.} In stage one, we train a penalty-based \DR learning method to infer the independent latent factors $C$. Although this stage can be realized with any penalty-based method, we utilize $\beta$-TCVAE \citep{mig} to learn the independent latent factors because it has been shown to perform well across all standard benchmarks and its total correlation penalty is simpler to compute in comparison to FactorVAE's \citep{factor}. We use the standard convolution and transposed-convolutions-based realization of $\beta$-TCVAE as provided in the \textit{disentanglement\_lib} \citep{icmlbest} package. As $\beta$-TCVAE heavily regularizes the latent space to be statistically independent -- not allowing enough capacity for the correlated factors -- the final reconstruction $Y$ will be low-quality.

\paragraph{Stage Two: Improving the Reconstruction.} In stage two, we learn the correlated factors $Z$ with another DGM and then use them to improve the reconstruction $Y$ while maintaining the conditioning on the independent factors $C$. To this end, we train a conditional VAE that models the data $X$ given $Y$ and $Z$. The encoder network $R_{\theta_Z}$ learns the posterior distribution over $Z$ as a function of the residual of $X$ and $Y$. The observation model $G_{\Theta}$ then reconstructs $X$ given $Y$ and $Z$. 
Incorporating $Z$ into $G_{\Theta}$ requires care as simply inputting the concatenation of $Y$ and $Z$ into the network at the start of training will result in a non-linear entanglement between $Y$ and $Z$. Once entangled this way, $G_{\Theta}$ will fail to condition on $Y$ sufficiently and use the entangled representation as a whole towards the reconstruction of $X$. To overcome this, we induce an architectural bias with a Feature-wise Linear Modulation (FiLM) technique \citep{perez2017film} prevalent in style transfer literature \citep{adaIN}, introducing $Z$ only through the adaptive instance normalization (AdaIN) layers of $G_{\Theta}$. As demonstrated in \citet{dumoulin2018feature}, inputting $Y$ into the decoder and then incorporating $Z$ through AdaIN at each layer allows the network to use $Z$ later in the generative process to model the residual information. This allows the network to better preserve the information in $Y$ while improving the reconstruction with $Z$. We demonstrate the importance of AdaIN by performing an ablation study of \simgan (Appendix \ref{app:ablation}).

\paragraph{FiLM and AdaIN.}
Recently, FiLM have been proposed as a general-purpose conditioning technique for deep neural networks \citep{perez2018film}. 
A FiLM layer learns an affine transformation of the intermediate statistics of a deep neural network, conditioned on an external input. FiLM has found great success in style transfer literature where the goal is to render content from a source domain in the style of a target domain. One such method, Instance Normalization \citep{instancenorm}, incorporates the style and the content information through the normalization layers in a feed-forward network, learning style-specific affine parameters from the data that are used to shift and scale the feature statistics (per instance) extracted from the source image across spatial dimensions. 
\citet{adaIN} extended this method to address arbitrary styles by taking the affine parameters to be the feature statistics extracted from the style domain, arriving at the $AdaIN(y,x) = \sigma(x) \Big( \frac{y-\mu(y)}{\sigma(y)} \Big) + \mu(x)$ which is applied for each normalization layer. 
Here, $y$ are the incoming activations for a normalization layer in the network that is receiving an image from the source domain and $x$ are the style features computed from the target domain. The features statistics for $y$ (as well as for for $x$) are computed as follows: $\mu(y) = \frac{1}{HW}\sum_{h, w} a_{nhwc}$ and $\sigma(y) = \sqrt{\frac{1}{HW}\sum_{h, w} (x_{nhwc}-\mu(y))^2 + \epsilon}$, where $h,w,c$ stand for the height, width and channel, respectively. In \simgan, the equivalent of style features is the representation of the residual information between $X$ and $Y$ captured in the correlated variables $Z$. The goal is similar to style transfer: incorporate the information stored in $Z$ while maintaining the semantic content stored in $Y$. Therefore, we slightly modify the AdaIN formulation as follows: $AdaIN(y,z) = \gamma(z) \Big( \frac{y-\mu(y)}{\sigma(y)} \Big) + \beta(z)$.
where $\mu(y)$ and $\sigma(y)$ are the same as before and $\gamma(z)$ and $\beta(z)$ are learned functions of $z$ parameterized by a fully-connected neural network. As mentioned before, incorporating $Z$ through AdaIN at each layer allows the network to use $Z$ later in the generative process to model the residual information, thus preserving the information in $Y$ \citep{dumoulin2018feature}.


\paragraph{\simgangan.} Equation \ref{eq:sim_elbo} also allows for realizing subgraph $(Y,Z) \rightarrow X$ using a GAN, with a little manipulation. Simply by adding and subtracting the log of the true conditional density $\log p(X|Y)$ in \eqref{eq:sim_elbo}, we get the lower bound,
\begin{align}
    \label{eq:sim_elbo_gan} 
    &\log p_{\{\phi,\theta,\theta_Z,\Theta\}}(x,y) \geq - \mathcal{KL}[q_\phi(c|y) \Vert p(c)] +  \\ 
    &\int q_\phi(c|y)\log p_\theta(y|c) dc  
    - \mathcal{KL}[q_{\theta_Z}(z|x,y) \Vert p(z)] +\nonumber \underbrace{ \int q_{\theta_Z}(z|x,y)\log \frac{p_{\Theta}(x|z,y)}{p(x|y)} dz}_{\text{CGAN-based }(Y,Z) \rightarrow X} + \log p(x|y)
\end{align}
Here, the log density ratio $\log \frac{p_{\Theta}(x|z,y)}{p(x|y)}$ can be estimated using adversarial learning with a binary discriminator function \citep{sugiyama2012density,srivastava2017veegan}. In practice, since we are learning conditional distributions, the decoder is implemented as a conditional GAN (CGAN) \cite{mirza2014conditional}. We refer to this implementation as \simgangan. We show that \simgangan leads to even higher fidelity generation than \simgan while delivering better disentanglement than conditional GAN. We also qualitatively compare \simgangan to the InfoGAN model \citep{info} for the unsupervised DR learning. Due to space constraints, we defer the details and all the results of this study to the Appendix \ref{sec:gan_impl}.

\paragraph{\simganflow.} We also present a simple a FLOW-based version of our framework, \simganflow, that utilizes a mixture of Gaussians model to realize the sub-graph $C \rightarrow Y$ and a flow-based model to realize the subgraph $(Y,Z) \rightarrow X$. The flow-based model in \simgan can be implemented as a conditional normalizing flow that conditions on $Y$ \cite{dinh14_nice,huang18_neural_autor_flows,durkan19_neural_splin_flows}.
As we are limited for space, we defer the details of the model and the experiment results to the Appendix \ref{sec:flow_impl}.

\section{Experiments}
\label{sec:experiments}
\paragraph{Baseline Models}
\citet{icmlbest} demonstrated that most state-of-the-art \DR learning methods are able to effectively learn a factorized posterior distribution (i.e. extract independent latent variables $C$). Therefore, 
we chose $\beta$-TCVAE 
as the baseline \DR learning method for our paper. 
For fair comparison with our work, we use $\beta$-TCVAE to implement the sub-graph $C \rightarrow Y$ in \simgan. For our main experiments, we thoroughly evaluate three models: (i) \simgan with latent dimensions $C$ for the $C \rightarrow Y$ network and $Z$ for the $(Y,Z) \rightarrow X$ network, (ii) $\beta$-TCVAE with latent dimension size $|C|$, and (iii) $\beta$-TCVAE-L with latent dimension size $|C + Z|$. In line with previous work \citep{icmlbest}, we use a 10-dimensional $C$. We use a 5-dimensional and 10-dimensional $Z$ for \textit{Cars3D} and \textit{Small}Norb, respectively. In Appendix \ref{sec:c0}, we sweep over the dimensionalities of $C$ and $Z$ to evaluate its effect on the reported results. We train $\beta$-TCVAE with $\beta$ values ranging from 1 to 10. In addition to the main models, we include comparisons to a recent state-of-the-art method, Jacobian Supervision \citep{lezama2018overcoming}. Although Jacobian supervision can be implemented with multiple student autoencoders, we only compare to the single student model version as it is most directly comparable to \simgan. We utilize the same architecture and $C, Z$ dimensionalities as \simgan. For their teacher model, we also use a pre-trained $\beta$-TCVAE model as we did for \simgan, but we only evaluate their method for $\beta=1,10$. The hyperparameters and loss functions are as specified in the paper and are available in the linked code repository.

\paragraph{Measuring Disentanglement Performance}
\citet{icmlbest} concluded that most metrics for measuring disentanglement performance correlate relatively well with each other. Therefore, we only use MIG \citep{mig} to quantitatively measure disentanglement performance. Please see Appendix \ref{sec:mig_desc} for how this is computed. 
\paragraph{Measuring Reconstruction Quality.}
Following \citep{razavi2019generating}, we report the Frechet Inception Distance (FID) \citep{fid} between the dataset and each models' reconstructions. We choose FID instead of reconstruction error for our main metric as L1/L2 metrics are known to not correlate well with perceived image quality \citep{wang2003multiscale, wang2004image}. We still report both L1 and L2 results in Appendix \ref{sec:pw_recon}. Additionally, FID does not require a well-defined likelihood function and can, therefore, be used to evaluate \simgan when the conditional model is implemented using a GAN.
\paragraph{Measuring the Conditioning on $C$.}
To quantitatively measure how well the improved reconstructions are conditioned on the independent latent variables $C$, we use the following procedure. Let the mean parameter of the encoder of the subgraph $C \rightarrow Y$ be $E_\phi$. Further, let $C_R = E_\phi(X)$, $C_Y = E_\phi(Y)$, $C_X = E_\phi(G_\Theta(Y,Z))$, $C_{z_\epsilon} = E_\phi(G_\Theta(Y,\epsilon)))$, and $C_{Y_\epsilon} = E_\phi(G_\Theta(\epsilon,Z)))$ where $X$ is an input image, $\epsilon \sim \mathcal{N}(0,I)$, and $G_\Theta$ is the decoder of the second stage of \simgan. We report the following mutual information terms: $M_1 = \text{MI}[ C_R; C_Y ], M_2 = \text{MI}[ C_R; C_X ]$, $M_3 = \text{MI}[ C_R; C_{z_\epsilon} ]$, and $M_4 = \text{MI}[ C_R; C_{Y_\epsilon} ]$. If $Z$ is encoding meaningful information for the reconstruction process then we should see $M_2 > M_1$ as the reconstruction should better approximate the ground truth image X. More importantly, if $M_1 > M_3 >> M_4$ then the conditioning on $Y$ (and therefore $C$) can be assumed to be maintained. This follows because if $G_\Theta$ was only using $Z$ for the final reconstruction, $M_3$ would be close to zero and $M_4$ would be high (since Y is not being used in the final reconstruction). In addition to this quantitative analysis, we provide latent traversal plots for all the datasets to qualitatively show that the conditioning on the disentangled factors $C$ is preserved.

\paragraph{Downstream Classification}
To demonstrate that $Z$ is learning meaningful residual information that is \textit{complimentary} to $C$, we perform a set of downstream classification tasks using both $Z$ and $C$. In these tasks, we predict attributes of objects in the \textit{Small}NORB dataset using an MLP that is trained with three different inputs: $C$, $Z$, and both $C$ and $Z$ concatenated together. Ideally, the MLP trained with both $C$ and $Z$ would achieve the highest accuracy as the inferred latent variables contain complementary information about the object attributes.

\paragraph{Ablation Studies}
\simgan design employs four major modelling choices. In order to explain their individual impact on the model, we carried out a set of empirical analysis. Specifically, we performed ablation studies to understand the contribution of: 
(1) Our two-step training paradigm (vs end-to-end training).
(2) Our use of $Y$ instead of $C$ for the second stage (vs the  underlying graphical model of ID-GAN \cite{lee2020high}).
(3) Our use of AdaIN to incorporate $Z$.
(4) Our decision to learn $C$ before learning $Z$.
To evaluate our two-step training process, we compare \simgan to a model where both the $C$ and $Z$ factors are learned together in an end-to-end manner (by penalizing only $C$) and $Z$ is incorporated using AdaIN. 
To evaluate how using $Y$ improves reconstruction performance over using $C$, we compare \simgan to a version of \simgan where $Y$ is replaced with $C$. 
To evaluate the contribution of AdaIN for disentanglement, we compare \simgan to a version of \simgan where $C$ and $Z$ are concatenated together as input to the second stage DGM. Finally, to evaluate our decision to learn $C$ before learning $Z$, we train a version of \simgan where $Z$ is learned in the first stage and then $C$ is learned in the second stage ($C$ and $Z$ are concatenated together to reconstruct $X$). Due to space constraints, all experiments are reported in the Appendix \ref{app:ablation}.

\paragraph{Code and Datasets.}
We use the \textit{disentanglement\_lib} package \citep{icmlbest} to train all $\beta$-TCVAE models and to evaluate MIG. The datasets we use for our quantitative benchmarks are \textit{Cars3D} \citep{reed2015deep} and \textit{Small}NORB \citep{lecun2004learning}. These datasets are a subset of those used in the large scale study of \citep{icmlbest} and are available (with ground truth factors) from the \textit{disentanglement\_lib} package.  Additionally, we use the CelebA \citep{liu2018large} dataset of real face images for qualitative evaluations since the other datasets consist of only simulated images (see Appendix \ref{sec:celeba}). 
For FID evaluations, we use the standard Tensorflow \citep{abadi2016tensorflow} implementation for all the models. We provide all the hyperparameter and architectural details for the experiments in the appendix. A reference Tensorflow implementation is also available at \url{https://github.com/AnonymousAuthors000/DS-VAE} in the form of a Jupyter notebook.

\section{Results on Computer Vision Datasets}
\label{sec:results}
\begin{figure*}[t!]
\centering
\subfloat[\textit{Cars3D} dataset]{%
  \centering
  \label{fig:cars_main}
    \scalebox{.5}{
\begin{tikzpicture}

\definecolor{color0}{rgb}{0.12156862745098,0.466666666666667,0.705882352941177}
\definecolor{color1}{rgb}{1,0.498039215686275,0.0549019607843137}
\definecolor{color2}{rgb}{0.172549019607843,0.627450980392157,0.172549019607843}
\definecolor{color3}{rgb}{0.83921568627451,0.152941176470588,0.156862745098039}
\definecolor{color4}{rgb}{0.580392156862745,0.403921568627451,0.741176470588235}
\definecolor{color5}{rgb}{0.549019607843137,0.337254901960784,0.294117647058824}

\begin{groupplot}[
    group style={
        group size= 2 by 1,
        horizontal sep=0.1\linewidth
    }, 
    width=\linewidth,
]

    \nextgroupplot[
    width=0.48\linewidth,
    title={(i)},
    title style={at={(1,-0.065)},anchor=south west,draw=black,fill=white},
    legend cell align={left},
    legend columns=10,
    legend style={fill opacity=0.8, draw opacity=1, text opacity=1,
    /tikz/every even column/.append style={column sep=0.5cm},
    at={(-0.15,1.08)}, anchor=south west, draw=white!80!black},
    tick pos=left,
    x grid style={white!69.0196078431373!black},
    xlabel={\(\displaystyle \beta\)},
    xmajorgrids,
    xmin=0.55, xmax=10.45,
    xtick style={color=black},
    y grid style={white!69.0196078431373!black},
    ylabel={FID},
    ymajorgrids,
    ymin=4.19083251229766, ymax=159.048744614286,
    ytick style={color=black}
    ]
    \path [fill=color0, fill opacity=0.4]
    (axis cs:1,147.042667013777)
    --(axis cs:1,140.536027461007)
    --(axis cs:2,140.170643725401)
    --(axis cs:4,136.409110627301)
    --(axis cs:6,139.205515764288)
    --(axis cs:8,142.122533979246)
    --(axis cs:10,144.215322498909)
    --(axis cs:10,150.634732035311)
    --(axis cs:10,150.634732035311)
    --(axis cs:8,149.99642938141)
    --(axis cs:6,150.807897042358)
    --(axis cs:4,139.882676392937)
    --(axis cs:2,145.331236501172)
    --(axis cs:1,147.042667013777)
    --cycle;
    
    \path [fill=color1, fill opacity=0.4]
    (axis cs:1,134.951965498164)
    --(axis cs:1,130.071259524577)
    --(axis cs:2,135.710741458477)
    --(axis cs:4,142.042356902519)
    --(axis cs:6,146.369278904421)
    --(axis cs:8,148.856097312104)
    --(axis cs:10,146.523563169138)
    --(axis cs:10,152.00974860965)
    --(axis cs:10,152.00974860965)
    --(axis cs:8,151.346705547402)
    --(axis cs:6,147.692792973249)
    --(axis cs:4,143.867658478011)
    --(axis cs:2,137.274909509279)
    --(axis cs:1,134.951965498164)
    --cycle;
    
    
    \path [fill=color3, fill opacity=0.4]
    (axis cs:1,11.9353681511386)
    --(axis cs:1,11.2298285169335)
    --(axis cs:2,15.2149682534959)
    --(axis cs:4,22.1220951094503)
    --(axis cs:6,27.1193465043186)
    --(axis cs:8,30.7303880928467)
    --(axis cs:10,36.201427079351)
    --(axis cs:10,37.4574454498709)
    --(axis cs:10,37.4574454498709)
    --(axis cs:8,32.7116789668796)
    --(axis cs:6,30.0104079602281)
    --(axis cs:4,23.1923239800145)
    --(axis cs:2,15.9382296258151)
    --(axis cs:1,11.9353681511386)
    --cycle;
    
    \addplot [semithick, color0]
    table {%
    1 143.789347237392
    2 142.750940113286
    4 138.145893510119
    6 145.006706403323
    8 146.059481680328
    10 147.42502726711
    };
    \addlegendentry{$\beta$-TCVAE}
    \addplot [semithick, color1]
    table {%
    1 132.511612511371
    2 136.492825483878
    4 142.955007690265
    6 147.031035938835
    8 150.101401429753
    10 149.266655889394
    };
    \addlegendentry{$\beta$-TCVAE-L}
    
    \addplot [semithick, color3]
    table {%
    1 11.582598334036
    2 15.5765989396555
    4 22.6572095447324
    6 28.5648772322733
    8 31.7210335298631
    10 36.829436264611
    };
    \addlegendentry{\simgan}
    
\addplot [only marks, mark=*, draw=color2, fill=color2, colormap/viridis]
table{%
x                      y
1 59.93050371177986
10 71.0353058574579
};
    \addlegendentry{\cite{lezama2018overcoming}}
    
    \addplot [only marks, mark=*, draw=color4, fill=color4, colormap/viridis]
table{%
x                      y
1 22.5
};
    \addlegendentry{Big-VAE}
    

    \nextgroupplot[
    width=0.48\linewidth,
    title={(ii)},
    title style={at={(1,-0.065)},anchor=south west,draw=black,fill=white},
    legend cell align={left},
    legend columns=2,
    legend style={fill opacity=0.8, draw opacity=1, text opacity=1, at={(-0.6,1.1)}, anchor=south west, draw=white!80!black},
    tick pos=left,
    x grid style={white!69.0196078431373!black},
    xlabel={\(\displaystyle \beta\)},
    xmajorgrids,
    xmin=0.55, xmax=10.45,
    xtick style={color=black},
    y grid style={white!69.0196078431373!black},
    ylabel={MIG},
    ymajorgrids,
    ymin=-0.00442773861792501, ymax=0.226869294601003,
    ytick style={color=black}
    ]

    \path [fill=color0, fill opacity=0.4]
    (axis cs:1,0.0912759788311178)
    --(axis cs:1,0.00608576289202625)
    --(axis cs:2,0.0363861443453772)
    --(axis cs:4,0.0699225280910173)
    --(axis cs:6,0.149326663854955)
    --(axis cs:8,0.156419052178519)
    --(axis cs:10,0.171801357909799)
    --(axis cs:10,0.216355793091051)
    --(axis cs:10,0.216355793091051)
    --(axis cs:8,0.165919885467228)
    --(axis cs:6,0.168103207162912)
    --(axis cs:4,0.124816827813815)
    --(axis cs:2,0.117905017008345)
    --(axis cs:1,0.0912759788311178)
    --cycle;
    
    \path [fill=color1, fill opacity=0.4]
    (axis cs:1,0.0671145397335569)
    --(axis cs:1,0.0278369920618631)
    --(axis cs:2,0.0451243692745419)
    --(axis cs:4,0.0417723821717762)
    --(axis cs:6,0.11713608571454)
    --(axis cs:8,0.110657809280169)
    --(axis cs:10,0.115827073345479)
    --(axis cs:10,0.13519522871584)
    --(axis cs:10,0.13519522871584)
    --(axis cs:8,0.140163024506039)
    --(axis cs:6,0.140127902301989)
    --(axis cs:4,0.14461534728505)
    --(axis cs:2,0.100793443842011)
    --(axis cs:1,0.0671145397335569)
    --cycle;

    \addplot [semithick, color0]
    table {%
    1 0.048680870861572
    2 0.0771455806768613
    4 0.0973696779524163
    6 0.158714935508934
    8 0.161169468822873
    10 0.194078575500425
    };
    \addplot [semithick, color1]
    table {%
    1 0.04747576589771
    2 0.0729589065582766
    4 0.0931938647284133
    6 0.128631994008264
    8 0.125410416893104
    10 0.125511151030659
    };

    \addplot [only marks, mark=*, draw=color2, fill=color2, colormap/viridis]
    table{%
    x                      y
    1 0.01214418196300139
    10 0.22038522631341714
    };

\end{groupplot}

\end{tikzpicture}}
}
\subfloat[\textit{Small}NORB dataset]{%
  \centering
  \label{fig:sn_main}
  \scalebox{.5}{
\begin{tikzpicture}

\definecolor{color0}{rgb}{0.12156862745098,0.466666666666667,0.705882352941177}
\definecolor{color1}{rgb}{1,0.498039215686275,0.0549019607843137}
\definecolor{color2}{rgb}{0.172549019607843,0.627450980392157,0.172549019607843}
\definecolor{color3}{rgb}{0.83921568627451,0.152941176470588,0.156862745098039}
\definecolor{color4}{rgb}{0.580392156862745,0.403921568627451,0.741176470588235}
\definecolor{color5}{rgb}{0.549019607843137,0.337254901960784,0.294117647058824}
\definecolor{color6}{rgb}{0.890196078431372,0.466666666666667,0.76078431372549}

\begin{groupplot}[
    group style={
        group size= 2 by 1,
        horizontal sep=0.1\linewidth
    }, 
    width=\linewidth,
]

    \nextgroupplot[
    width=0.48\linewidth,
    title={(i)},
    title style={at={(1,-0.065)},anchor=south west,draw=black,fill=white},
    legend cell align={left},
    legend columns=4,
    legend style={fill opacity=0.8, draw opacity=1, text opacity=1, at={(0,1.3)}, anchor=south west, draw=white!80!black},
    tick pos=left,
    x grid style={white!69.0196078431373!black},
    xlabel={\(\displaystyle \beta\)},
    xmajorgrids,
    xmin=0.55, xmax=10.45,
    xtick style={color=black},
    y grid style={white!69.0196078431373!black},
    ylabel={FID},
    ymajorgrids,
    ymin=28.0758438885456, ymax=258.281180843513,
    ytick style={color=black}
    ]
    \path [fill=color0, fill opacity=0.4]
    (axis cs:1,157.273616737868)
    --(axis cs:1,149.572595349467)
    --(axis cs:2,173.990858647709)
    --(axis cs:4,200.641802667413)
    --(axis cs:6,225.283694568451)
    --(axis cs:8,236.079422305354)
    --(axis cs:10,239.838549559433)
    --(axis cs:10,247.817301891015)
    --(axis cs:10,247.817301891015)
    --(axis cs:8,242.24280161665)
    --(axis cs:6,228.663922705672)
    --(axis cs:4,205.437901585536)
    --(axis cs:2,178.252844270938)
    --(axis cs:1,157.273616737868)
    --cycle;
    
    \path [fill=color1, fill opacity=0.4]
    (axis cs:1,158.378121474805)
    --(axis cs:1,153.258393831605)
    --(axis cs:2,178.819210059525)
    --(axis cs:4,203.320364902088)
    --(axis cs:6,225.599053386352)
    --(axis cs:8,237.515096053798)
    --(axis cs:10,245.275382469037)
    --(axis cs:10,247.201029801204)
    --(axis cs:10,247.201029801204)
    --(axis cs:8,243.446471237899)
    --(axis cs:6,226.515578052797)
    --(axis cs:4,208.498306471652)
    --(axis cs:2,182.255777894978)
    --(axis cs:1,158.378121474805)
    --cycle;
    
    
    \path [fill=color3, fill opacity=0.4]
    (axis cs:1,46.8727948663412)
    --(axis cs:1,44.8050835198068)
    --(axis cs:2,62.6395388671849)
    --(axis cs:4,54.8665475642999)
    --(axis cs:6,40.3855470704)
    --(axis cs:8,40.0300089271083)
    --(axis cs:10,38.5397228410441)
    --(axis cs:10,39.3033777102516)
    --(axis cs:10,39.3033777102516)
    --(axis cs:8,40.8448967217297)
    --(axis cs:6,47.0155384942035)
    --(axis cs:4,63.4435025260928)
    --(axis cs:2,64.3457121756779)
    --(axis cs:1,46.8727948663412)
    --cycle;
    
    \addplot [semithick, color0]
    table {%
    1 153.423106043668
    2 176.121851459323
    4 203.039852126475
    6 226.973808637062
    8 239.161111961002
    10 243.827925725224
    };
    \addplot [semithick, color1]
    table {%
    1 155.818257653205
    2 180.537493977252
    4 205.90933568687
    6 226.057315719574
    8 240.480783645848
    10 246.238206135121
    };
    \addplot [semithick, color3]
    table {%
    1 45.838939193074
    2 63.4926255214314
    4 59.1550250451964
    6 43.7005427823017
    8 40.437452824419
    10 38.9215502756479
    };
    \addplot [only marks, mark=*, draw=color2, fill=color2, colormap/viridis]
table{%
x                      y
1 59.44100583673139
10 56.0944150815759
};
   \addplot [only marks, mark=*, draw=color4, fill=color4, colormap/viridis]
    table{%
    x                      y
    1 86.8
    };
    \nextgroupplot[
    width=0.48\linewidth,
    title={(ii)},
    title style={at={(1,-0.065)},anchor=south west,draw=black,fill=white},
    legend cell align={left},
    legend columns=2,
    legend style={fill opacity=0.8, draw opacity=1, text opacity=1, at={(-0.1\linewidth,1.1)}, anchor=south west, draw=white!80!black},
    tick pos=left,
    x grid style={white!69.0196078431373!black},
    xlabel={\(\displaystyle \beta\)},
    xmajorgrids,
    xmin=0.55, xmax=10.45,
    xtick style={color=black},
    y grid style={white!69.0196078431373!black},
    ylabel={MIG},
    ymajorgrids,
    ymin=0.0755906415082958, ymax=0.266143807502697,
    ytick style={color=black}
    ]

    \path [fill=color0, fill opacity=0.4]
    (axis cs:1,0.244953462635574)
    --(axis cs:1,0.240995901927481)
    --(axis cs:2,0.199603604326742)
    --(axis cs:4,0.211248883525098)
    --(axis cs:6,0.172788844926542)
    --(axis cs:8,0.160699848457918)
    --(axis cs:10,0.147313512346739)
    --(axis cs:10,0.171512044081615)
    --(axis cs:10,0.171512044081615)
    --(axis cs:8,0.180798998591248)
    --(axis cs:6,0.190625679502397)
    --(axis cs:4,0.221179470479708)
    --(axis cs:2,0.224875537274634)
    --(axis cs:1,0.244953462635574)
    --cycle;
    
    \path [fill=color1, fill opacity=0.4]
    (axis cs:1,0.257482299957497)
    --(axis cs:1,0.244177601246938)
    --(axis cs:2,0.200045676692978)
    --(axis cs:4,0.193036854114443)
    --(axis cs:6,0.193648916360918)
    --(axis cs:8,0.154108530226388)
    --(axis cs:10,0.141284703367676)
    --(axis cs:10,0.169394782060336)
    --(axis cs:10,0.169394782060336)
    --(axis cs:8,0.17002411792992)
    --(axis cs:6,0.216280854831646)
    --(axis cs:4,0.222488510768487)
    --(axis cs:2,0.212590420099501)
    --(axis cs:1,0.257482299957497)
    --cycle;
    
    \addplot [semithick, color0]
    table {%
    1 0.242974682281528
    2 0.212239570800688
    4 0.216214177002403
    6 0.18170726221447
    8 0.170749423524583
    10 0.159412778214177
    };
    \addplot [semithick, color1]
    table {%
    1 0.250829950602217
    2 0.206318048396239
    4 0.207762682441465
    6 0.204964885596282
    8 0.162066324078154
    10 0.155339742714006
    };
    
    \addplot [only marks, mark=*, draw=color2, fill=color2, colormap/viridis]
    table{%
    x                      y
    1 0.18764354950477138
    10 0.17365575198137675
    };

\end{groupplot}

\end{tikzpicture}}
}

\caption{FID (lower is better) and MIG (higher is better) comparison of $\beta$-TCVAE, $\beta$-TCVAE-L, and \simgan models. On both datasets, \simgan is able to consistently improve the reconstruction quality of its underlying $\beta$-TCVAE model while achieving a better MIG than $\beta$-TCVAE-L . We also provide FID and MIG results for Lezama's model \citep{lezama2018overcoming} with $\beta=1,10$ as well as FID for a vanilla VAE model of the same capacity as \simgan (denoted Big-VAE).}
\label{fig:fid}
\end{figure*}

\begin{figure*}[h!]
\centering
\subfloat[\textit{Cars3D} dataset]{%
  \centering
  \label{fig:cars_MI}
    \scalebox{.4}{
\begin{tikzpicture}
\tikzstyle{every node}=[font=\fontsize{18}{30}\selectfont]
\pgfplotsset{
every tick label/.append style={scale=1.0},
every axis label/.append style={scale=1.0},
}
\definecolor{color0}{rgb}{0.12156862745098,0.466666666666667,0.705882352941177}
\definecolor{color1}{rgb}{1,0.498039215686275,0.0549019607843137}
\definecolor{color2}{rgb}{0.172549019607843,0.627450980392157,0.172549019607843}
\definecolor{color3}{rgb}{0.83921568627451,0.152941176470588,0.156862745098039}
\definecolor{color4}{rgb}{0.580392156862745,0.403921568627451,0.741176470588235}
\definecolor{color5}{rgb}{0.549019607843137,0.337254901960784,0.294117647058824}

\begin{axis}[
width=.48\linewidth,
legend cell align={left},
legend columns=2,
legend style={fill opacity=0.8, draw opacity=1, text opacity=1, at={(0.12,1.15)}, anchor=center,align=center, draw=white!80!black},
    /tikz/every even column/.append style={column sep=0.5cm},
tick pos=left,
x grid style={white!69.0196078431373!black},
xlabel={\(\displaystyle \beta\)},
xmajorgrids,
xmin=0.55, xmax=10.45,
xtick style={color=black},
y grid style={white!69.0196078431373!black},
ylabel={MI},
ymajorgrids,
ymin=0.0, ymax=1.00593678868124,
ytick style={color=black}
]
    \path [fill=color0, fill opacity=0.4]
    (axis cs:1,0.709721742797889)
    --(axis cs:1,0.703350100048022)
    --(axis cs:2,0.677458529338209)
    --(axis cs:4,0.631310005333881)
    --(axis cs:6,0.592128038702002)
    --(axis cs:8,0.571862711043809)
    --(axis cs:10,0.5574706207826)
    --(axis cs:10,0.575773590291703)
    --(axis cs:10,0.575773590291703)
    --(axis cs:8,0.580166860717607)
    --(axis cs:6,0.604689838718549)
    --(axis cs:4,0.638375019412633)
    --(axis cs:2,0.683209403987982)
    --(axis cs:1,0.709721742797889)
    --cycle;
    
    
    \path [fill=color2, fill opacity=0.4]
    (axis cs:1,0.498299938561123)
    --(axis cs:1,0.474410468053202)
    --(axis cs:2,0.449334433643528)
    --(axis cs:4,0.420362814635308)
    --(axis cs:6,0.390079552104446)
    --(axis cs:8,0.373316449237438)
    --(axis cs:10,0.348259741691998)
    --(axis cs:10,0.37052855338393)
    --(axis cs:10,0.37052855338393)
    --(axis cs:8,0.379193029882554)
    --(axis cs:6,0.400621746079426)
    --(axis cs:4,0.422084094964967)
    --(axis cs:2,0.461194870791895)
    --(axis cs:1,0.498299938561123)
    --cycle;
    
    \path [fill=color3, fill opacity=0.4]
    (axis cs:1,0.792861919168781)
    --(axis cs:1,0.78229048024884)
    --(axis cs:2,0.742273437469228)
    --(axis cs:4,0.68269555002313)
    --(axis cs:6,0.645729577452759)
    --(axis cs:8,0.618800930412712)
    --(axis cs:10,0.600625075738941)
    --(axis cs:10,0.60828174384887)
    --(axis cs:10,0.60828174384887)
    --(axis cs:8,0.627543252479977)
    --(axis cs:6,0.650845541781928)
    --(axis cs:4,0.69410195148763)
    --(axis cs:2,0.74576093714405)
    --(axis cs:1,0.792861919168781)
    --cycle;

    \path [fill=color4, fill opacity=0.4]
    (axis cs:1,0.0790667515100361)
    --(axis cs:1,0.0702184178520322)
    --(axis cs:2,0.0735525827928724)
    --(axis cs:4,0.0916033806254232)
    --(axis cs:6,0.107801349222112)
    --(axis cs:8,0.122090863094621)
    --(axis cs:10,0.129457575277303)
    --(axis cs:10,0.143924209935821)
    --(axis cs:10,0.143924209935821)
    --(axis cs:8,0.129302520719975)
    --(axis cs:6,0.135845631885447)
    --(axis cs:4,0.120269221362959)
    --(axis cs:2,0.0879407020235312)
    --(axis cs:1,0.0790667515100361)
    --cycle;
    
    \addplot [semithick, color0]
    table {%
    1 0.706535921422955
    2 0.680333966663096
    4 0.634842512373257
    6 0.598408938710275
    8 0.576014785880708
    10 0.566622105537152
    };
    \addlegendentry{$\beta$-TCVAE}
    \addplot [semithick, color2]
    table {%
    1 0.486355203307163
    2 0.455264652217711
    4 0.421223454800138
    6 0.395350649091936
    8 0.376254739559996
    10 0.359394147537964
    };
    \addlegendentry{\simgan}
    \addplot [semithick, color3]
    table {%
    1 0.78757619970881
    2 0.744017187306639
    4 0.68839875075538
    6 0.648287559617344
    8 0.623172091446345
    10 0.604453409793905
    };
    \addplot [semithick, color4]
    table {%
    1 0.0746425846810341
    2 0.0807466424082018
    4 0.105936300994191
    6 0.12182349055378
    8 0.125696691907298
    10 0.136690892606562
    };
\end{axis}

\end{tikzpicture}}
}
\subfloat[\textit{Small}NORB dataset]{%
  \centering
  \label{fig:norb_MI}
  \scalebox{.4}{
\begin{tikzpicture}
\tikzstyle{every node}=[font=\fontsize{18}{30}\selectfont]
\pgfplotsset{
every tick label/.append style={scale=1.0},
every axis label/.append style={scale=1.0},
}

\definecolor{color0}{rgb}{0.12156862745098,0.466666666666667,0.705882352941177}
\definecolor{color1}{rgb}{1,0.498039215686275,0.0549019607843137}
\definecolor{color2}{rgb}{0.172549019607843,0.627450980392157,0.172549019607843}
\definecolor{color3}{rgb}{0.83921568627451,0.152941176470588,0.156862745098039}
\definecolor{color4}{rgb}{0.580392156862745,0.403921568627451,0.741176470588235}
\definecolor{color5}{rgb}{0.549019607843137,0.337254901960784,0.294117647058824}

\begin{axis}[
width=0.48\linewidth,
legend cell align={left},
legend columns=2,
legend style={fill opacity=0.8, draw opacity=1, text opacity=1, at={(0.12,1.15)}, anchor=center,align=center, draw=white!80!black},
    /tikz/every even column/.append style={column sep=0.5cm},
tick pos=left,
x grid style={white!69.0196078431373!black},
xlabel={\(\displaystyle \beta\)},
xmajorgrids,
xmin=0.55, xmax=10.45,
xtick style={color=black},
y grid style={white!69.0196078431373!black},
ylabel={MI},
ymajorgrids,
ymin=0.0, ymax=1.00593678868124,
ytick style={color=black}
]
    \path [fill=color0, fill opacity=0.4]
    (axis cs:1,0.445140572600982)
    --(axis cs:1,0.4265002910694)
    --(axis cs:2,0.39505615692591)
    --(axis cs:4,0.328573893608853)
    --(axis cs:6,0.283640005180642)
    --(axis cs:8,0.26285112968067)
    --(axis cs:10,0.268767182523572)
    --(axis cs:10,0.283148991710752)
    --(axis cs:10,0.283148991710752)
    --(axis cs:8,0.300134805835506)
    --(axis cs:6,0.288577710732918)
    --(axis cs:4,0.341384505371743)
    --(axis cs:2,0.419339254816194)
    --(axis cs:1,0.445140572600982)
    --cycle;
    
    
    \path [fill=color2, fill opacity=0.4]
    (axis cs:1,0.459096273216278)
    --(axis cs:1,0.404965305354976)
    --(axis cs:2,0.363754896921226)
    --(axis cs:4,0.246586738086392)
    --(axis cs:6,0.208192745910629)
    --(axis cs:8,0.192750020637533)
    --(axis cs:10,0.188727553037793)
    --(axis cs:10,0.198176607486956)
    --(axis cs:10,0.198176607486956)
    --(axis cs:8,0.223916728696539)
    --(axis cs:6,0.244137892273527)
    --(axis cs:4,0.251265841441947)
    --(axis cs:2,0.372060534849737)
    --(axis cs:1,0.459096273216278)
    --cycle;
    
    \path [fill=color3, fill opacity=0.4]
    (axis cs:1,0.490988079872732)
    --(axis cs:1,0.471924301841508)
    --(axis cs:2,0.415151039099657)
    --(axis cs:4,0.42599624534879)
    --(axis cs:6,0.458930511857604)
    --(axis cs:8,0.490889629567135)
    --(axis cs:10,0.499935561740744)
    --(axis cs:10,0.513667924729582)
    --(axis cs:10,0.513667924729582)
    --(axis cs:8,0.505676820311253)
    --(axis cs:6,0.478626432125433)
    --(axis cs:4,0.43520626850189)
    --(axis cs:2,0.442102856525755)
    --(axis cs:1,0.490988079872732)
    --cycle;
    
    \path [fill=color4, fill opacity=0.4]
    (axis cs:1,0.00963774926327792)
    --(axis cs:1,0.00637228426850185)
    --(axis cs:2,0.0085257568623931)
    --(axis cs:4,0.0155374617671502)
    --(axis cs:6,0.00677971368226869)
    --(axis cs:8,0.00783782087688204)
    --(axis cs:10,0.00936111165718604)
    --(axis cs:10,0.0120413459130083)
    --(axis cs:10,0.0120413459130083)
    --(axis cs:8,0.0106103311309925)
    --(axis cs:6,0.0146275889395218)
    --(axis cs:4,0.0158423691250162)
    --(axis cs:2,0.0115190177555731)
    --(axis cs:1,0.00963774926327792)
    --cycle;
    
    \addplot [semithick, color0]
    table {%
    1 0.435820431835191
    2 0.407197705871052
    4 0.334979199490298
    6 0.28610885795678
    8 0.281492967758088
    10 0.275958087117162
    };
    \addplot [semithick, color2]
    table {%
    1 0.432030789285627
    2 0.367907715885482
    4 0.248926289764169
    6 0.226165319092078
    8 0.208333374667036
    10 0.193452080262375
    };
    \addplot [semithick, color3]
    table {%
    1 0.48145619085712
    2 0.428626947812706
    4 0.43060125692534
    6 0.468778471991519
    8 0.498283224939194
    10 0.506801743235163
    };
    \addlegendentry{\simgan (Random $Z$)}
    \addplot [semithick, color4]
    table {%
    1 0.00800501676588988
    2 0.0100223873089831
    4 0.0156899154460832
    6 0.0107036513108953
    8 0.00922407600393725
    10 0.0107012287850972
    };
    \addlegendentry{\simgan (Random Y)}

\end{axis}

\end{tikzpicture}}
}
\caption{Mutual Information (MI) between inferred independent factors from the true image X (using $\beta$-TCVAE) and independent factors from various reconstructions of X. Please note that Blue $=M_1$, Red $=M_2$, Green $=M_3$, and Purple $=M_4$ (see Section \ref{sec:experiments} for their definitions).}
\label{fig:mutual_information}
\end{figure*}

\begin{figure*}[h!]
\centering
\subfloat[Category]{%
  \centering
  \begin{minipage}[b]{0.2\textwidth}
  \centering
   \scalebox{0.3}{
\begin{tikzpicture}
\tikzstyle{every node}=[font=\fontsize{18}{30}\selectfont]
\pgfplotsset{
every tick label/.append style={scale=1.0},
every axis label/.append style={scale=1.0},
}

\definecolor{color0}{rgb}{0.12156862745098,0.466666666666667,0.705882352941177}
\definecolor{color1}{rgb}{1,0.498039215686275,0.0549019607843137}
\definecolor{color2}{rgb}{0.172549019607843,0.627450980392157,0.172549019607843}

\begin{axis}[
legend cell align={left},
legend columns=3,
legend style={fill opacity=0.8, draw opacity=1, text opacity=1, at={(0.12,1.11)}, anchor=south west, draw=white!80!black},
    /tikz/every even column/.append style={column sep=0.5cm},
tick pos=left,
x grid style={white!69.0196078431373!black},
xlabel={\(\displaystyle \beta\)},
xmajorgrids,
xmin=0.55, xmax=10.45,
xtick style={color=black},
y grid style={white!69.0196078431373!black},
ylabel={Accuracy},
ymajorgrids,
ymin=0.0, ymax=1.00593678868124,
ytick style={color=black}
]
\path [fill=color0, fill opacity=0.4]
(axis cs:1,0.971671137531143)
--(axis cs:1,0.971192668439006)
--(axis cs:2,0.94164176501592)
--(axis cs:4,0.915509269911182)
--(axis cs:6,0.826320830250019)
--(axis cs:8,0.77927689250006)
--(axis cs:10,0.760875694996638)
--(axis cs:10,0.764388608485949)
--(axis cs:10,0.764388608485949)
--(axis cs:8,0.794417137350686)
--(axis cs:6,0.830939741889284)
--(axis cs:4,0.92396523257638)
--(axis cs:2,0.951082115581095)
--(axis cs:1,0.971671137531143)
--cycle;

\path [fill=color1, fill opacity=0.4]
(axis cs:1,0.491967378722387)
--(axis cs:1,0.42050152675025)
--(axis cs:2,0.463260507523559)
--(axis cs:4,0.468819825603518)
--(axis cs:6,0.495983371505079)
--(axis cs:8,0.519991779234789)
--(axis cs:10,0.541470008839241)
--(axis cs:10,0.551347155339864)
--(axis cs:10,0.551347155339864)
--(axis cs:8,0.534112698377151)
--(axis cs:6,0.522129190683976)
--(axis cs:4,0.501562637083049)
--(axis cs:2,0.496472079541118)
--(axis cs:1,0.491967378722387)
--cycle;

\path [fill=color2, fill opacity=0.4]
(axis cs:1,0.97805891906548)
--(axis cs:1,0.972112100835018)
--(axis cs:2,0.953933904066135)
--(axis cs:4,0.966953711300311)
--(axis cs:6,0.938514672119673)
--(axis cs:8,0.934632811649592)
--(axis cs:10,0.932200787251511)
--(axis cs:10,0.937746352051972)
--(axis cs:10,0.937746352051972)
--(axis cs:8,0.937879626161354)
--(axis cs:6,0.945658213452466)
--(axis cs:4,0.96976892551561)
--(axis cs:2,0.961100299913965)
--(axis cs:1,0.97805891906548)
--cycle;

\addplot [semithick, color0]
table {%
1 0.971431902985075
2 0.946361940298508
4 0.919737251243781
6 0.828630286069652
8 0.786847014925373
10 0.762632151741294
};
\addlegendentry{C}
\addplot [semithick, color1]
table {%
1 0.456234452736318
2 0.479866293532338
4 0.485191231343284
6 0.509056281094527
8 0.52705223880597
10 0.546408582089552
};
\addlegendentry{Z}
\addplot [semithick, color2]
table {%
1 0.975085509950249
2 0.95751710199005
4 0.96836131840796
6 0.94208644278607
8 0.936256218905473
10 0.934973569651741
};
\addlegendentry{C+Z}
\end{axis}

\end{tikzpicture}}
  \end{minipage}
}
\subfloat[Instance]{%
  \begin{minipage}[b]{0.2\textwidth}
  \centering
  \scalebox{0.3}{
\begin{tikzpicture}
\tikzstyle{every node}=[font=\fontsize{18}{30}\selectfont]
\pgfplotsset{
every tick label/.append style={scale=1.0},
every axis label/.append style={scale=1.0},
}

\definecolor{color0}{rgb}{0.12156862745098,0.466666666666667,0.705882352941177}
\definecolor{color1}{rgb}{1,0.498039215686275,0.0549019607843137}
\definecolor{color2}{rgb}{0.172549019607843,0.627450980392157,0.172549019607843}

\begin{axis}[
legend cell align={left},
legend columns=3,
legend style={fill opacity=0.8, draw opacity=1, text opacity=1, at={(0.12,1.11)}, anchor=south west, draw=white!80!black},
    /tikz/every even column/.append style={column sep=0.5cm},
tick pos=left,
x grid style={white!69.0196078431373!black},
xlabel={\(\displaystyle \beta\)},
xmajorgrids,
xmin=0.55, xmax=10.45,
xtick style={color=black},
y grid style={white!69.0196078431373!black},
ymajorgrids,
ymin=0.0, ymax=1.00593678868124,
ymajorticks=false
]
\path [fill=color0, fill opacity=0.4]
(axis cs:1,0.588283878733011)
--(axis cs:1,0.548998459575446)
--(axis cs:2,0.468893223117932)
--(axis cs:4,0.365710386389322)
--(axis cs:6,0.253664087001568)
--(axis cs:8,0.221950729451735)
--(axis cs:10,0.215249395758457)
--(axis cs:10,0.220229459962936)
--(axis cs:10,0.220229459962936)
--(axis cs:8,0.241901882488563)
--(axis cs:6,0.269190390610372)
--(axis cs:4,0.396571951919135)
--(axis cs:2,0.491228045538784)
--(axis cs:1,0.588283878733011)
--cycle;

\path [fill=color1, fill opacity=0.4]
(axis cs:1,0.176793196380978)
--(axis cs:1,0.155529564813052)
--(axis cs:2,0.151977880965405)
--(axis cs:4,0.265200046754666)
--(axis cs:6,0.309866431482772)
--(axis cs:8,0.32034613507025)
--(axis cs:10,0.328451001258203)
--(axis cs:10,0.344279098244284)
--(axis cs:10,0.344279098244284)
--(axis cs:8,0.327663815178506)
--(axis cs:6,0.326638543641606)
--(axis cs:4,0.272502067673195)
--(axis cs:2,0.154303213561958)
--(axis cs:1,0.176793196380978)
--cycle;

\path [fill=color2, fill opacity=0.4]
(axis cs:1,0.608124377157952)
--(axis cs:1,0.560874379060954)
--(axis cs:2,0.492151041523523)
--(axis cs:4,0.504684099160114)
--(axis cs:6,0.461070201268597)
--(axis cs:8,0.429182484094289)
--(axis cs:10,0.439140228143124)
--(axis cs:10,0.448530791757373)
--(axis cs:10,0.448530791757373)
--(axis cs:8,0.466573112920636)
--(axis cs:6,0.481327186791104)
--(axis cs:4,0.532162915765259)
--(axis cs:2,0.507693485839661)
--(axis cs:1,0.608124377157952)
--cycle;

\addplot [semithick, color0]
table {%
1 0.568641169154229
2 0.480060634328358
4 0.381141169154229
6 0.26142723880597
8 0.231926305970149
10 0.217739427860697
};
\addplot [semithick, color1]
table {%
1 0.166161380597015
2 0.153140547263682
4 0.26885105721393
6 0.318252487562189
8 0.324004975124378
10 0.336365049751244
};
\addplot [semithick, color2]
table {%
1 0.584499378109453
2 0.499922263681592
4 0.518423507462687
6 0.471198694029851
8 0.447877798507463
10 0.443835509950249
};
\end{axis}

\end{tikzpicture}}
  \end{minipage}
}
\subfloat[Elevation]{%
  \begin{minipage}[b]{0.2\textwidth}
  \centering
  \scalebox{0.3}{
\begin{tikzpicture}
\tikzstyle{every node}=[font=\fontsize{18}{30}\selectfont]
\pgfplotsset{
every tick label/.append style={scale=1.0},
every axis label/.append style={scale=1.0},
}

\definecolor{color0}{rgb}{0.12156862745098,0.466666666666667,0.705882352941177}
\definecolor{color1}{rgb}{1,0.498039215686275,0.0549019607843137}
\definecolor{color2}{rgb}{0.172549019607843,0.627450980392157,0.172549019607843}

\begin{axis}[
legend cell align={left},
legend columns=6,
legend style={fill opacity=0.8, draw opacity=1, text opacity=1,
font=\small,
at={(0.12,1.11)}, anchor=south west, draw=white!80!black},
    /tikz/every even column/.append style={column sep=0.5cm},
tick pos=left,
x grid style={white!69.0196078431373!black},
xlabel={\(\displaystyle \beta\)},
xmajorgrids,
xmin=0.55, xmax=10.45,
xtick style={color=black},
y grid style={white!69.0196078431373!black},
ymajorgrids,
ymin=0.0, ymax=1.00593678868124,
ymajorticks=false
]
\path [fill=color0, fill opacity=0.4]
(axis cs:1,0.448932489309389)
--(axis cs:1,0.391241640043844)
--(axis cs:2,0.419248072355079)
--(axis cs:4,0.385425864012492)
--(axis cs:6,0.351733528253185)
--(axis cs:8,0.322118192989843)
--(axis cs:10,0.305979545403222)
--(axis cs:10,0.317621200865435)
--(axis cs:10,0.317621200865435)
--(axis cs:8,0.338873722433043)
--(axis cs:6,0.38015390955776)
--(axis cs:4,0.396912444445219)
--(axis cs:2,0.432275559485717)
--(axis cs:1,0.448932489309389)
--cycle;

\path [fill=color1, fill opacity=0.4]
(axis cs:1,0.185203899979598)
--(axis cs:1,0.171761274149755)
--(axis cs:2,0.160580083557074)
--(axis cs:4,0.167652692225641)
--(axis cs:6,0.181180413704514)
--(axis cs:8,0.205413982705409)
--(axis cs:10,0.210696713014348)
--(axis cs:10,0.220973063105055)
--(axis cs:10,0.220973063105055)
--(axis cs:8,0.207054922767227)
--(axis cs:6,0.191720705698471)
--(axis cs:4,0.172987855038041)
--(axis cs:2,0.173141931368299)
--(axis cs:1,0.185203899979598)
--cycle;

\path [fill=color2, fill opacity=0.4]
(axis cs:1,0.486346477762299)
--(axis cs:1,0.460559616765064)
--(axis cs:2,0.461885531726109)
--(axis cs:4,0.43514153935428)
--(axis cs:6,0.398992241299373)
--(axis cs:8,0.3608605904825)
--(axis cs:10,0.365600345939729)
--(axis cs:10,0.377869803314002)
--(axis cs:10,0.377869803314002)
--(axis cs:8,0.402743265238893)
--(axis cs:6,0.407366589546398)
--(axis cs:4,0.440791296466616)
--(axis cs:2,0.476391831457971)
--(axis cs:1,0.486346477762299)
--cycle;

\addplot [semithick, color0]
table {%
1 0.420087064676617
2 0.425761815920398
4 0.391169154228856
6 0.365943718905473
8 0.330495957711443
10 0.311800373134328
};
\addplot [semithick, color1]
table {%
1 0.178482587064677
2 0.166861007462687
4 0.170320273631841
6 0.186450559701493
8 0.206234452736318
10 0.215834888059701
};
\addplot [semithick, color2]
table {%
1 0.473453047263682
2 0.46913868159204
4 0.437966417910448
6 0.403179415422886
8 0.381801927860697
10 0.371735074626866
};
\end{axis}

\end{tikzpicture}}
  \end{minipage}
}
\subfloat[Azimuth]{%
  \begin{minipage}[b]{0.2\textwidth}
  \centering
  \scalebox{0.3}{
\begin{tikzpicture}
\tikzstyle{every node}=[font=\fontsize{18}{30}\selectfont]
\pgfplotsset{
every tick label/.append style={scale=1.0},
every axis label/.append style={scale=1.0},
}

\definecolor{color0}{rgb}{0.12156862745098,0.466666666666667,0.705882352941177}
\definecolor{color1}{rgb}{1,0.498039215686275,0.0549019607843137}
\definecolor{color2}{rgb}{0.172549019607843,0.627450980392157,0.172549019607843}

\begin{axis}[
legend cell align={left},
legend columns=3,
legend style={fill opacity=0.8, draw opacity=1, text opacity=1, at={(0.12,1.11)}, anchor=south west, draw=white!80!black},
    /tikz/every even column/.append style={column sep=0.5cm},
tick pos=left,
x grid style={white!69.0196078431373!black},
xlabel={\(\displaystyle \beta\)},
xmajorgrids,
xmin=0.55, xmax=10.45,
xtick style={color=black},
y grid style={white!69.0196078431373!black},
ymajorgrids,
ymin=0.0, ymax=1.00593678868124,
ymajorticks=false
]
\path [fill=color0, fill opacity=0.4]
(axis cs:1,0.47962767122151)
--(axis cs:1,0.465257279027246)
--(axis cs:2,0.416209362314752)
--(axis cs:4,0.299136908135264)
--(axis cs:6,0.236669398744293)
--(axis cs:8,0.213804052187517)
--(axis cs:10,0.211076588647478)
--(axis cs:10,0.219893560606253)
--(axis cs:10,0.219893560606253)
--(axis cs:8,0.230070325921936)
--(axis cs:6,0.256723014191031)
--(axis cs:4,0.35314853962593)
--(axis cs:2,0.425830438680273)
--(axis cs:1,0.47962767122151)
--cycle;

\path [fill=color1, fill opacity=0.4]
(axis cs:1,0.129774738617272)
--(axis cs:1,0.10872028625835)
--(axis cs:2,0.110336509931461)
--(axis cs:4,0.167175225715812)
--(axis cs:6,0.206971179882418)
--(axis cs:8,0.221436765320311)
--(axis cs:10,0.230105210726903)
--(axis cs:10,0.238489316636282)
--(axis cs:10,0.238489316636282)
--(axis cs:8,0.231299553087649)
--(axis cs:6,0.217780063898677)
--(axis cs:4,0.183726515577721)
--(axis cs:2,0.126370579620777)
--(axis cs:1,0.129774738617272)
--cycle;

\path [fill=color2, fill opacity=0.4]
(axis cs:1,0.490795536281619)
--(axis cs:1,0.468937050783057)
--(axis cs:2,0.410826758757181)
--(axis cs:4,0.429148204412172)
--(axis cs:6,0.41197707814532)
--(axis cs:8,0.392249226131608)
--(axis cs:10,0.401607584875894)
--(axis cs:10,0.419909828059429)
--(axis cs:10,0.419909828059429)
--(axis cs:8,0.406180500236551)
--(axis cs:6,0.42532080742682)
--(axis cs:4,0.45378090006544)
--(axis cs:2,0.426160181541327)
--(axis cs:1,0.490795536281619)
--cycle;

\addplot [semithick, color0]
table {%
1 0.472442475124378
2 0.421019900497512
4 0.326142723880597
6 0.246696206467662
8 0.221937189054726
10 0.215485074626866
};
\addplot [semithick, color1]
table {%
1 0.119247512437811
2 0.118353544776119
4 0.175450870646766
6 0.212375621890547
8 0.22636815920398
10 0.234297263681592
};
\addplot [semithick, color2]
table {%
1 0.479866293532338
2 0.418493470149254
4 0.441464552238806
6 0.41864894278607
8 0.39921486318408
10 0.410758706467662
};
\end{axis}

\end{tikzpicture}}
  \end{minipage}
}
\subfloat[Lighting]{%
  \begin{minipage}[b]{0.2\textwidth}
  \centering
  \scalebox{0.3}{
\begin{tikzpicture}
\tikzstyle{every node}=[font=\fontsize{18}{30}\selectfont]
\pgfplotsset{
every tick label/.append style={scale=1.0},
every axis label/.append style={scale=1.0},
}

\definecolor{color0}{rgb}{0.12156862745098,0.466666666666667,0.705882352941177}
\definecolor{color1}{rgb}{1,0.498039215686275,0.0549019607843137}
\definecolor{color2}{rgb}{0.172549019607843,0.627450980392157,0.172549019607843}

\begin{axis}[
legend cell align={left},
legend columns=3,
legend style={fill opacity=0.8, draw opacity=1, text opacity=1, at={(0.12,1.11)}, anchor=south west, draw=white!80!black},
    /tikz/every even column/.append style={column sep=0.5cm},
tick pos=left,
x grid style={white!69.0196078431373!black},
xlabel={\(\displaystyle \beta\)},
xmajorgrids,
xmin=0.55, xmax=10.45,
xtick style={color=black},
y grid style={white!69.0196078431373!black},
ymajorgrids,
ymin=0.0, ymax=1.00593678868124,
ymajorticks=false
]
\path [fill=color0, fill opacity=0.4]
(axis cs:1,0.950513433007749)
--(axis cs:1,0.945553109280808)
--(axis cs:2,0.942215587459139)
--(axis cs:4,0.935286836064273)
--(axis cs:6,0.926363033301239)
--(axis cs:8,0.901257896948602)
--(axis cs:10,0.901418515754251)
--(axis cs:10,0.906650514096495)
--(axis cs:10,0.906650514096495)
--(axis cs:8,0.920648197578761)
--(axis cs:6,0.929358359733587)
--(axis cs:4,0.93730333806508)
--(axis cs:2,0.950586029456284)
--(axis cs:1,0.950513433007749)
--cycle;

\path [fill=color1, fill opacity=0.4]
(axis cs:1,0.67116640177495)
--(axis cs:1,0.552636458921567)
--(axis cs:2,0.65121659769383)
--(axis cs:4,0.525726172341197)
--(axis cs:6,0.439617661858782)
--(axis cs:8,0.575640387768201)
--(axis cs:10,0.613424845472306)
--(axis cs:10,0.636886099801325)
--(axis cs:10,0.636886099801325)
--(axis cs:8,0.616057373425829)
--(axis cs:6,0.611843780927288)
--(axis cs:4,0.531021340096614)
--(axis cs:2,0.678307656037514)
--(axis cs:1,0.67116640177495)
--cycle;

\path [fill=color2, fill opacity=0.4]
(axis cs:1,0.94901912512277)
--(axis cs:1,0.9461923176633)
--(axis cs:2,0.939834465930184)
--(axis cs:4,0.935478844123483)
--(axis cs:6,0.928066192984333)
--(axis cs:8,0.908097879729195)
--(axis cs:10,0.90714190318477)
--(axis cs:10,0.914064564476922)
--(axis cs:10,0.914064564476922)
--(axis cs:8,0.929666423753392)
--(axis cs:6,0.930298234876364)
--(axis cs:4,0.939909837468557)
--(axis cs:2,0.947059190786233)
--(axis cs:1,0.94901912512277)
--cycle;

\addplot [semithick, color0]
table {%
1 0.948033271144279
2 0.946400808457711
4 0.936295087064677
6 0.927860696517413
8 0.910953047263682
10 0.904034514925373
};
\addplot [semithick, color1]
table {%
1 0.611901430348259
2 0.664762126865672
4 0.528373756218905
6 0.525730721393035
8 0.595848880597015
10 0.625155472636816
};
\addplot [semithick, color2]
table {%
1 0.947605721393035
2 0.943446828358209
4 0.93769434079602
6 0.929182213930348
8 0.918882151741294
10 0.910603233830846
};
\end{axis}

\end{tikzpicture}}
  \end{minipage}
}
\caption{Object property prediction for \textit{Small}NORB using inferred representations $C$, $Z$, and $C+Z$ as input to a MLP. The accuracy is highest for the MLP trained with $C+Z$.}
\label{fig:predicting_label}
\end{figure*}
\vspace{-10pt}
\paragraph{Disentanglement Performance.} 
As mentioned above, \simgan utilizes $\beta$-TCVAE to learn disentangled factors in the first stage of its training. Therefore, we report the MIG values for the $\beta$-TCVAE used with \simgan and for $\beta$-TCVAE-L (a $\beta$-TCVAE with the same latent dimension as \simgan). For the \textit{Cars3D} dataset, we find that with both models, the MIG scores increase as $\beta$ is increased (Figure \ref{fig:cars_main}(ii)). However, $\beta$-TCVAE appears to consistently have a higher MIG than $\beta$-TCVAE-L, potentially reflecting the entangling effect of an overly large latent space. For the \textit{Small}NORB dataset, MIG decreases as the value of $\beta$ increases as shown in Figure \ref{fig:sn_main}(ii). This confirms the findings of \citet{icmlbest} on this particular dataset. Overall, these results illustrate that \simgan is able to extract disentangled factors from image data at the same level as state-of-the-art methods and that naively increasing the latent dimensionality of $\beta$-TCVAE may diminish disentanglement performance. For the Jacobian supervision baseline \citep{lezama2018overcoming}, we use $\beta$-TCVAE at the teacher model with two values of $\beta$ (1 and 10). After training, the MIG values for the student network are approximately the same as teacher model on both datasets (i.e. it maintains disentanglement).

\paragraph{Reconstruction Quality.}
As shown in Figures \ref{fig:cars_main}(i) and \ref{fig:sn_main}(i), \simgan drastically improves the reconstruction quality compared to both $\beta$-TCVAE models, achieving a much lower FID across all values of $\beta$. We also compare \simgan to Jacobian supervision and find that \simgan has lower FID on both datasets.
To ensure that the improvement in reconstruction quality over $\beta$-TCVAE is not just because \simgan has more parameters, we trained several $\beta$-TCVAE models with the same number of parameters as \simgan on the \textit{Small}NORB dataset and reported the results in Appendix \ref{sec:big-tcvae}. Although we found marginal improvements in the average FID for these bigger $\beta$-TCVAEs ($145.79\pm2.98$), they are still far higher than \simgan's FID ($59.16\pm4.52$). Also, increasing the parameters of $\beta$-TCVAE led to a decreased MIG ($0.163\pm0.026$) which indicates that increasing the parameters of $\beta$-TCVAE may increase reconstruction quality at the cost of disentanglement.
Finally, for reference, we provide FID scores for a standard VAE with same number of parameters as \simgan (we refer to this model as Big-VAE). Surprisingly, \simgan has a lower FID than Big-VAE at low values of $\beta$ suggesting that the residual modeling in \simgan leads to better image quality in general (see Figures \ref{fig:cars_main}(i) and \ref{fig:sn_main}(i)). 
\paragraph{Conditioning on $C$}
Consider Figures \ref{fig:cars_MI} and \ref{fig:norb_MI} where we report the mutual information terms $M_1, M_2$, $M_3$, and $M_4$ for both datasets (see Section \ref{sec:experiments} for their definitions). As per our expectation, we find that $M_2 > M_1 > M_3 >> M_4$ and that $M_3$ tracks $M_1$ without dropping significantly. This suggests that $Z$ is encoding meaningful information for the reconstruction of $X$ and that the model maintains the conditioning on $C$ in the final reconstruction. We further illustrate the conditioning on the $C$ with qualitative examples. In Figures \ref{fig:qual}(c), \ref{fig:qual}(d) and \ref{fig:qual}(e) in the Appendix \ref{sec:trf}, we show latent traversal of $Y$ and $X$ pairs. Notice, \simgan improves the quality of the generated output without diminishing the ability of the model to manipulate single factors of variation by traversing $C$.

\paragraph{Downstream Classification}
The results of the downstream attribute prediction tasks are shown in Figure~\ref{fig:predicting_label}. As can be seen, the MLP that was trained with both $C$ and $Z$ attains a higher accuracy on all five attribute prediction tasks than the MLPs trained with only $C$ or $Z$. Notice that as the regularization of the latent space increases (larger $\beta$) and the accuracy of the classier trained on $C$ decreases, the accuracy of the classifier trained on $Z$ increases. This highlights that $Z$ contains complementary information to that of $C$ and sheds light on why simply increasing the size of the latent space (i.e. the $\beta$TCVAE-L model) does not improve the reconstruction significantly. 

\section{Experiments with tabular data generation}
\label{sec:ex tab}
Synthetic data generation is an important topic at the intersection of machine learning and finance. The generated data can be used for critical downstream tasks such as fraud detection or customer propensity prediction. Moreover, such generated data is especially useful when the real data is limited or too sparse and irregular to be used for machine learning training.

Variational autoencoder (VAE) and its variations proved to be highly effective in data generation tasks on many types of data ranging from images to structured data such as molecules. Nonetheless, applying VAE to tabular data, which frequently appears in financial domains, is a non-trivial task. Many VAE models used for tabular data-related work tend to be in the healthcare domain (\cite{healthcare1, healthcare2}).

Often, the mechanism for data generation of VAE from the latent space is hidden and can only be accessed via the weights of the neural network. To allow interpretability for financial compliance and regulations, one needs to provide meaningful descriptions of each coordinate of the latent space. Moreover, we also expect that those coordinates can reflect the core factors underlying the original input data.

As a result, we are motivated to consider VAE models that can handle these two problems: TVAE (\cite{tvae}), $\beta$-TCVAE \citep{mig}, and MS-VAE. In this section, we apply these three models to generate synthetic credit tabular data, evaluate their performance, and analyze the corresponding latent features learned. Our goals are to explore and detect appropriate latent representation models for data generation tasks in the financial domain. The details of the experiments on credit tabular data are provided together with comprehensive model evaluation and analysis. Furthermore, we dedicate Section 7 to investigating how the two models $\beta$-TCVAE and MS-VAE allow interpretability for data transparency purposes. 
\subsection{Baseline modes TVAE}
Variational autoencoder was first developed to work on real-valued data, often image data. However, modeling the probability distribution of rows in tabular data and generating realistic synthetic data with VAE is a non-trivial task. \cite{tvae} develop TVAE framework to handle this task where tabular data contain a mixture of discrete and continuous variables. For discrete variables, \cite{tvae} uses the standard one-hot encoding, while the authors use mode-specific normalization to handle continuous feature columns. More specifically, TVAE models each continuous variable by a Gaussian mixture model so that each variable is represented by a pair of number $(\alpha, \beta)$. Here, $\beta$ indicates which Gaussian mode the variable belongs to and $\alpha$ is the normalized value of this variable w.r.t the chosen mode. TVAE's neural network architectures for encoder and decoder are then modified to represent the real variable $\alpha$, the distribution vectors for discrete variables, and $\beta$'s values of continuous variables.

\subsection{Experimental setup}
\textbf{Datasets.} We apply 3 methods TVAE, $\beta$-TCVAE and MS-VAE on a simulated tabular credit dataset. The simulated data has 20000 observations which are generated using several underlying factors represented by 11 variables: 
\begin{enumerate}
    \item Two demographic variables: Variable $L$ for gender, and variable $M$ for race.
    \item Four Utilization variables named $Ub, Ub_1, Ub_2, Ub_3$.
    \item Four Debt stress variables $S, S_1, S_2, S_3$ and debt stress variable threshold $T$.
\end{enumerate}

These 11 factors induces a credit tabular dataset with 19 features:
\begin{enumerate}
    \item Mortgage, Utilization, Amount Past Due
    \item Four balance features with different periods: Balance
    \item Four credit inquiry features
    \item Four open trade features
    \item Four delinquency status features
\end{enumerate}

\textbf{Data processing.} To improve performances across 3 VAE models, we perform several pre-processing steps consisting of log transform and normalization. Particularly, prior to training the model, we apply log transform on continuous variables of the original dataset, and apply min-max normalization on both continuous and categorical variables. Once the models are trained and the synthetic data is generated, we perform post-processing steps including: clipping the generated data to range [0,1], de-normalizing categorical variables and rounding off to the nearest integer, and de-logging and de-normalizing continuous variables.

\subsection{Evaluation metrics}\label{metrics}
The synthetic tabular data is assessed on two aspects: Classifier 2-sample test (C2ST) or Detection test and statistical similarity. The first metric, proposed by \cite{ctab_gan}, evaluates the faithfulness of the synthetic data to the original data. Unlike ML-efficacy, which measures the (test) performance gap between predictive models trained on real versus synthetic data, C2ST evaluates the ability of a discriminant classifier to detect real data from fake data. Combined with statistical similarity, these metrics thus quantify the realism of the synthetic data set and whether it can be used as a proxy for the original data.

\subsubsection{Statistical Similarity}
We use two metrics to measure the statistical similarity between the artificial and original datasets.

\textbf{Jensen-Shannon divergence (JSD)} (\cite{jensen_shannon}). The JSD, a symmetric version of KL divergence, calculates the relative entropy in information represented by distributions between individual features of real versus fake data. The metric is applied to the (normalized) synthetic and real data, and is bounded between 0 and 1. To allow easy interpretation, we provide the averaged JSD across all features as the final measure.

\textbf{Correlation Difference (CD)}. We use CD to capture how well the feature relations are maintained in the synthetic data. To calculate CD, we first calculate the pairwise correlation matrices for real and synthetic data separately. The final CD is the difference between these two correlation matrices calculated using Frobenious norm. We remark that for each correlation matrix, three cases are considered: (1) correlation between two continuous features is measured via Pearson correlation coefficient, (2) correlation between two categorical features is measured using Theil uncertainty coefficient, and (3) correlation between categorical and continuous features is measured with correlation ratio.

\subsubsection{Detection Test/Classifier 2-sample test (C2ST)}\label{c2st}
The original and the generated data should be indistinguishable not only from a statistical viewpoint but also from a machine-learning viewpoint. This is important for downstream tasks such as fraud detection because we don't want the fraud classifiers to make a distinction between real and generated data, leading to a possibility that the classifiers flag the generated data as fraud. 

Thus, to evaluate how closely the synthetic data follows the real data, we perform the Classifier 2-sample test (C2ST). The evaluation process consists of the following steps: (1) Split the original data into two datasets $D_{train}$ and $D_{test}$, (2) Use $D_{train}$ to train the generative models and generate a fake data set $D_{syn}$ with the same size as $D_{test}$, (3) Create a new dataset $\mathcal{T}$ with $D_{test}$ (labeled 1) and $D_{syn}$ (labeled 0) for real and fake data respectively, (4) Split $\mathcal{T}$ into $\mathcal{T}_{train}$ and $\mathcal{T}_{test}$, and train a binary classifier on $\mathcal{T}_{train}$ to produce the probability of a datapoint being fake, (5) Use \textit{Area Under the ROC Curve (AUC)} to measure the performance of this classifier on test data $\mathcal{T}_{test}$.

\subsection{Results analysis}
In this section, we present the empirical results evaluating the synthetic credit data generated by our main method MS-VAE. This process computes the metrics specified in Section \ref{metrics}, and the results are compared with two benchmark models: TVAE and $\beta$-TCVAE. Across three models, we perform hyper-parameters tuning for latent dimension and $\beta$ values. Specifically, each model was trained for 3000 epochs, with latent dimensions ranging from 2 to 14, and $\beta$ values ranging from 2 to 9.

\begin{figure}[h!]
\caption{JSD results for different $\beta$ at latent dim = 9 }
\label{beta_tuning}
\centering
\includegraphics[scale=0.8]{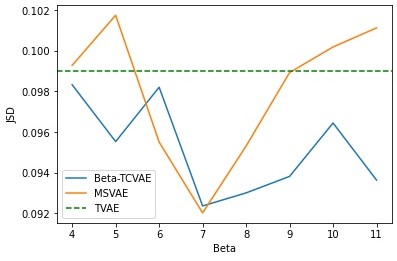}
\end{figure}

The tuning results are included in Figure \ref{beta_tuning}. The graph shows that MS-VAE generates synthetic data with the highest JSD at $\beta=7$ and latent 9. In the following parts, we provide experimental results across three VAE models at these specified hyper-parameters.

\textbf{Statistical similarity}. Results on JSD and CD performance for MS-VAE and benchmark models are presented in Table \ref{table:statistical_similarity_performance}. Here, we can see that while MS-VAE outperforms benchmark models on JSD, CD metric indicates that $\beta$-TCVAE best preserves the features interactions. This is perhaps due to its focus on latent disentanglement, while MS-VAE has a balance between reconstruction quality and disentanglement outcomes. The JSD performance comparison for each individual feature of the three models are provided in Appendix \ref{additional}.

\textbf{Detection Test (C2ST)}. We follow the steps detailed in Section \ref{c2st} to perform Detection Test for MS-VAE's generated data. Two binary classifiers used are Logistic Regression and Random Forest. To make the task more challenging, we also include the strongest-performing classifier from our experiments: XGBoost. As shown in Table \ref{table:detection_test}, MS-VAE outperforms the other two benchmarks on C2ST metric. We remark that an AUC of 0.5 means that the fake data is indistinguishable from the original data. Thus, XGBoost having AUC of 0.738 when detecting fake data implies that the MS-VAE-generated synthetic data is relatively faithful and similar (average JSD = 0.071) to the original dataset. We want to remark that an AUC of 0.7 corresponds to a very good performance as indicated in \cite{fool_xgboost}. $\beta$-TCVAE, while having better CD performance, has the worst performance among the 3 models in this test. This suggests that $\beta$-TCVAE's relatively lower reconstruction quality makes it vulnerable to fake data classifiers.

\begin{table}[h!]
\centering
\caption{Statistical Similarity performance for latent dim = 9, $\beta = 7$}
\label{table:statistical_similarity_performance}
  \renewcommand{\arraystretch}{1.2}
  \begin{tabular}{|p{3cm}|c|c|c|c|}
    \hline
    \multirow{2}{5cm}{\textbf{Models/Metrics}} & \multicolumn{2}{c|}{\textbf{JSD} $\downarrow$} & \multicolumn{2}{c|}{\textbf{CD}$\downarrow$}\\
    \cline{2-3} \cline{4-5}
    & \textbf{Train} & \textbf{Test}
    & \textbf{Train} & \textbf{Test}\\
    \hline
    $\beta$-TCVAE & \textbf{0.068} & 0.0923 & \textbf{0.855} & \textbf{0.926}\\ \hline
    MS-VAE & 0.071 & \textbf{0.092} & 0.868 & 0.99\\ \hline
    TVAE & 0.066 & 0.099 & 1.259 & 1.183  \\ \hline
  \end{tabular}
\end{table}

\begin{table}[h!]
\centering
\caption{Detection Test (C2ST) performance for latent dim = 9, $\beta = 7$}
\label{table:detection_test}
  \renewcommand{\arraystretch}{1.2}
  \begin{tabular}{|p{3cm}|c|c|c|c|c|c|}
    \hline
    \multirow{2}{5cm}{\textbf{Models/AUC $\downarrow$}} & \multicolumn{2}{c|}{\textbf{XGBoost}} & \multicolumn{2}{c|}{\textbf{Logistic Regression}} & \multicolumn{2}{c|}{\textbf{Random Forest}}\\
    \cline{2-3} \cline{4-5} \cline{6-7}
    & \textbf{Train} & \textbf{Test}
    & \textbf{Train} & \textbf{Test}
    & \textbf{Train} & \textbf{Test}\\
    \hline
    $\beta$-TCVAE & 0.826 & 0.81 & 0.588 & 0.595 & 0.8 & 0.809\\ \hline
    MS-VAE &\textbf{0.74} & \textbf{0.738} & 0.581 & 0.592 & \textbf{0.749} & \textbf{0.741}\\ \hline
    TVAE & 0.745 & 0.743 &  \textbf{0.557} &\textbf{0.577} &0.754 & 0.75 \\ \hline
  \end{tabular}
\end{table}

\subsection{Further discussions}
In this section, we provide several examples of the individual feature performance of synthetic data generated by MS-VAE, $\beta$-TCVAE, and TVAE and their corresponding analyses. The four types of variables discussed below are relatively representative of the attributes available within the original credit tabular dataset.

\begin{figure}[h!]
\caption{Original data distribution ($1^{st}$ plot) vs distribution for generated data from MS-VAE ($2^{nd}$ plot), $\beta$-TCVAE ($3^{rd}$ plot), and TVAE ($4^{th}$ plot) for column \textbf{Amount Past Due}}
\label{fig:reconstructed_data_amount_past_due}
\centering
\includegraphics[scale=0.2]{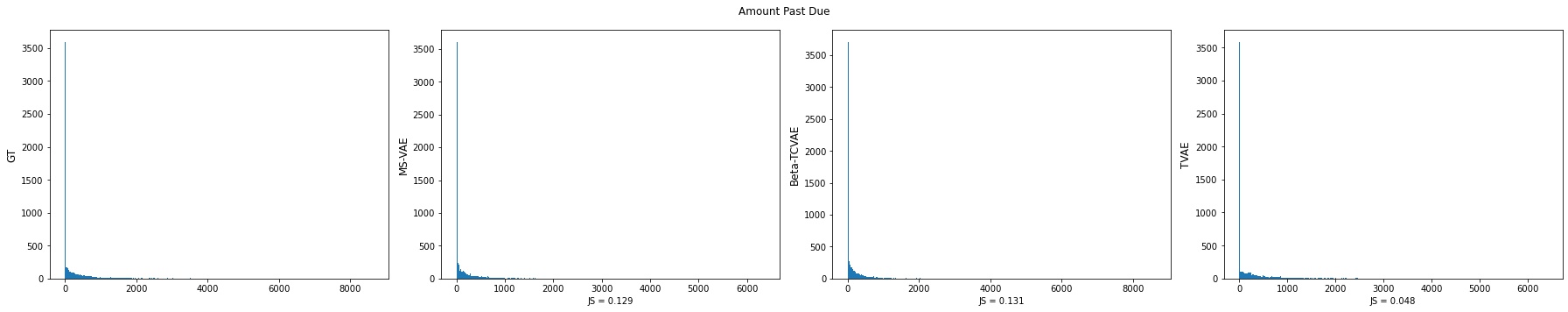}
\end{figure}

\textbf{Discrete variables.} The categorical (discrete) feature \textbf{Credit Inquiry\_3m} has 7 possible values. Figure \ref{fig:reconstructed_data_amount_past_due} shows the distribution plot of the original distribution for this feature against the distribution of the generated data produced by 3 VAE models. All models do very well in terms of reproducing the density for the most frequent value (value 0). Overall all three models do well in this case. However, there is a slight mismatch for the densities of values 1 and 2. Even though in original data, value 1 occurs less frequently than 2, both MS-VAE and $\beta$-TCVAE couldn't capture this property. TVAE even misses the underrepresented value $6$. Hence, additional care is required to handle the class imbalance problem for categorical features.
\begin{figure}[h!]
\caption{Original data distribution ($1^{st}$ plot) vs distribution for generated data from MS-VAE ($2^{nd}$ plot), $\beta$-TCVAE ($3^{rd}$ plot), and TVAE ($4^{th}$ plot) for column \textbf{Credit Inquiry\_3m}}
\label{fig:reconstructed_data_credit_inquire_3m}
\centering
\includegraphics[scale=0.2]{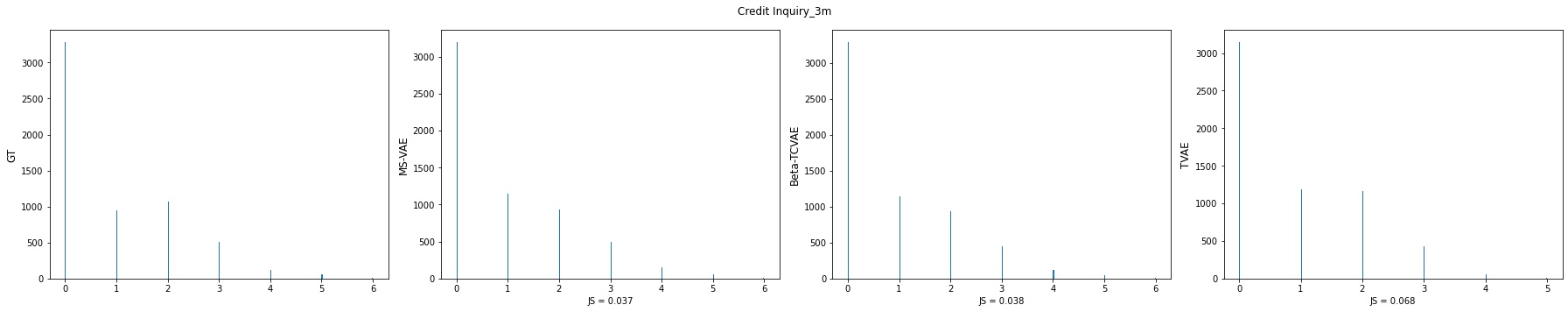}
\end{figure}

\textbf{Mixed data type variables.} The \text{Amount Past Due} feature contains both categorical value (denoted by $0$) and continuous values. For this special mixed-data type distribution, MS-VAE reconstructs the original data sufficiently well. On the other hand, $\beta$-TCVAE produces a somewhat over-concentrated histogram, while continuous values generated by TVAE are too spread out. From Figure \ref{fig:reconstructed_data_amount_past_due}, we observe that TVAE indicates a lack of special treatment to handle this feature type, while the distortion from $\beta$-TCVAE might come from its large reconstruction error when compared to MS-VAE.

\begin{figure}[h!]
\caption{Original data distribution ($1^{st}$ plot) vs distribution for generated data from MS-VAE ($2^{nd}$ plot), $\beta$-TCVAE ($3^{rd}$ plot), and TVAE ($4^{th}$ plot) for column \textbf{Balance\_6m}}
\label{fig:reconstructed_data_balance_6m}
\centering
\includegraphics[scale=0.2]{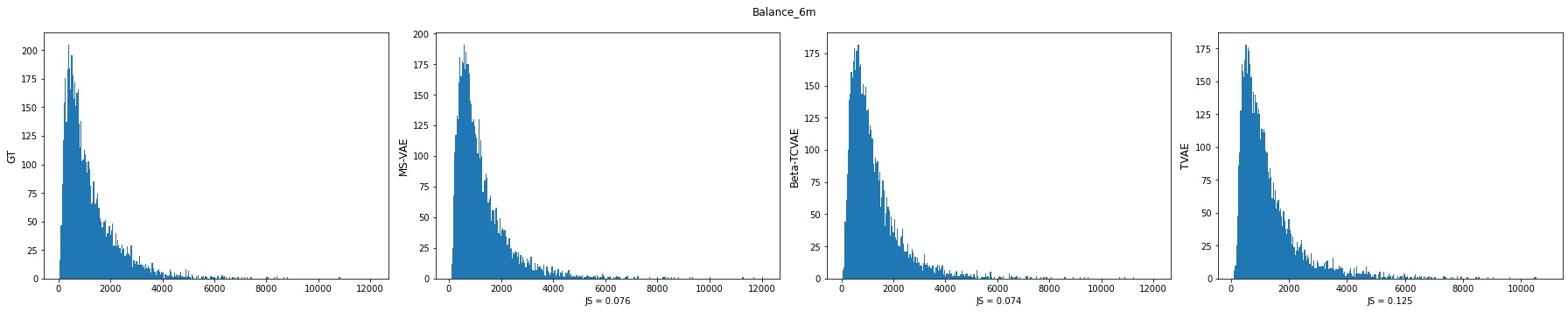}
\end{figure}

\begin{figure}[h!]
\caption{Original data distribution ($1^{st}$ plot) vs distribution for generated data from MS-VAE ($2^{nd}$ plot), $\beta$-TCVAE ($3^{rd}$ plot), and TVAE ($4^{th}$ plot) for column \textbf{Mortgage}}
\label{fig:reconstructed_data_mortgage}
\centering
\includegraphics[scale=0.2]{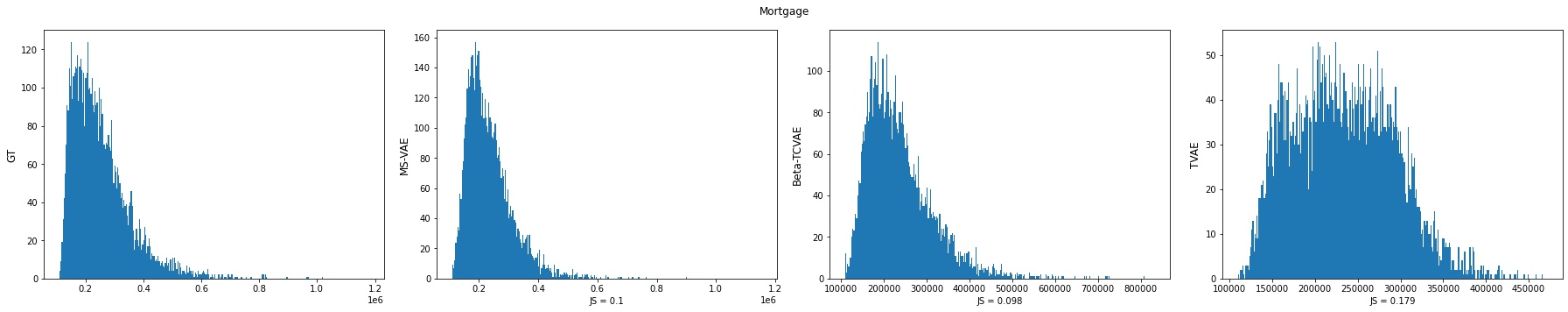}
\end{figure}

\textbf{Skewed continuous variables.} \textbf{Balance\_6m} and \textbf{Mortgage} are two features with skewed distribution. Even though their general distribution shapes are not widely different, the result is quite surprising. For \textbf{Balance\_6m}, all models capture the original data sufficiently well, while for \textbf{Mortgage}, only MS-VAE provides a decent performance(See Figure \ref{fig:reconstructed_data_balance_6m} and \ref{fig:reconstructed_data_mortgage}). A possible explanation for this phenomenon is the non-trivial irregularity appearing in \textbf{Mortgage} feature (with a small abrupt peak in the middle).

\begin{figure}[h!]
\caption{Original data distribution ($1^{st}$ plot) vs distribution for generated data from MS-VAE ($2^{nd}$ plot), $\beta$-TCVAE ($3^{rd}$ plot), and TVAE ($4^{th}$ plot) for column \textbf{Utilization}}
\label{fig:reconstructed_data_utilization}
\centering
\includegraphics[scale=0.2]{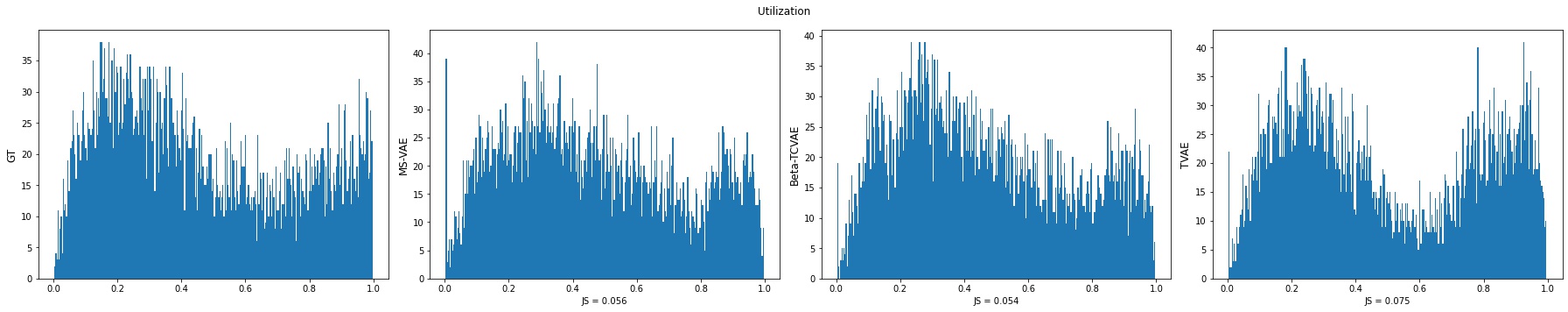}
\end{figure}

\textbf{Multi-mode continuous variables.} \textbf{Utilization} is a feature with multi-mode distribution. From Figure \ref{fig:reconstructed_data_utilization}, $\beta$-TCVAE performs best among all models for this type of distribution. On the other hand, MS-VAE produces unwanted artifacts at the left end, while TVAE's reconstructed distribution is more noisy than expected. For complex distributions as in this case, further research is needed to produce improvement.   

\section{Model interpretability}
Interpretability of latent space is especially important in finance. If we want to train a financial machine learning model using the data generated by VAE models, we need to explain clearly what are the sources (or the latent variables) used for the data generation process, and describe the generation process in details. Taking a vector of real numbers and then feeding into the black-box decoder as in a traditional VAE certainly doesn't offer enough interpretability for compliance and regulation purposes.

\subsection{Correlation of latent variables and original data}
One advantage of $\beta$-TCVAE and MS-VAE over TVAE is that they allow the disentanglement of variables of latent space thanks to the total correlation penalty term. From the disentanglement, we can hope to associate individual coordinate of the latent variables to the underlying factors from which the input distribution can be recovered. To check the effectiveness of this disentanglement with the end goal of recovering the hidden underlying factors, we measure the correlation between the model latent space's individual coordinate and the features of the original data. For our setting with 9-dimensional latent space, for each coordinate $z_1, \cdots, z_9$, we look at its correlations with each of the 19 original input features from tabular data. We define a correlation as \textbf{significant} if its value is at least $0.75$. We only retain significant correlations for each coordinate. Table \ref{table:correlation_table} shows the correlations for 2 models $\beta$-TCVAE and $TVAE$. We note that as the performance of $\beta$-TCVAE in Table \ref{table:correlation_table} is precisely the first stage of our multi-stage model, it carries the main interpretabilty information for MS-VAE.

\begin{table}
    \caption{Correlation between coordinates of latent vectors produced by different models and the original input features. We only include correlations that is significant (higher than 0.75).}
    \label{table:correlation_table}
    \centering
    \vspace{3mm}
    \begin{tabular}{lll}
    \toprule
    Coordinate type & $\beta$-TCVAE latent correlation with input & TVAE latent correlation with input\\
     & (Stage one of MS-VAE) & \\
    \midrule
    $z_1$ & \text{ } No significant correlation
          & No significant correlation \\
    \midrule
    $z_2$ & \begin{tabular}{ll} Mortgage & 0.910502 \\ \end{tabular}
        & No significant correlation \\
    \midrule
    $z_4$ & 
        \begin{tabular}{ll} Credit Inquiry\_12m & 0.898419 \\
                            Open Trade\_12m & 0.890393 \\ 
        \end{tabular}
          & No significant correlation \\
    \midrule
    $z_5$ & 
        \begin{tabular}{ll} Credit Inquiry\_6m & 0.808494 \\
                            Open Trade\_6m & 0.832294 \\
        \end{tabular}
          & No significant correlation \\
    \midrule
    $z_6$ & 
        \begin{tabular}{ll} Utilization & 0.904766 \\
                            Balance & 0.879174 \\
        \end{tabular}
          & No significant correlation \\
    \midrule
    $z_7$ & 
        \begin{tabular}{ll} Balance\_3m & 0.758381 \\
                            Balance\_6m & 0.849974 \\
                            Balance\_12m & 0.851672 \\
        \end{tabular}
          & No significant correlation \\
    \midrule
    $z_9$ & 
        \begin{tabular}{ll} Delinquency Status & 0.725198 \\
                            Delinquency Status\_3m & 0.877644 \\
                            Delinquency Status\_6m & 0.930754 \\
                            Delinquency Status\_12m & 0.928545 \\
        \end{tabular}
          & No significant correlation \\
    \bottomrule
    \end{tabular}
\end{table}

From the correlation table, we observe that for $\beta$-TCVAE model, each latent variable clearly corresponds to a separate group of features in the original data. More specifically, $z_2$ corresponds to the mortgage feature, while $z_4$ and $z_5$ variables are correlated to credit inquiry and open trade features for two separate periods 12 months and 6 months. On the other hand, $z_9$ is mainly correlated to delinquency status variables. Finally, $z_6$ corresponds to \textbf{Utilization} and \textbf{Balance} variables, while $z_7$ is correlated to the group of remaining 3 balance variables with different recent month periods. Each of these separate groups depends on a separate subset of underlying factors, suggesting that $\beta$-TCVAE did a reasonable job in separating the underlying factors. However, as the latent dimension is less than the input dimension and the group of input features may all depend on a single factor, it is very challenging to ask the model to separate entirely all features (each coordinate only has a significant relation with one feature).

Meanwhile, correlations for TVAE latent space's coordinates are relatively insignificant and spread out over all input features. A more detailed table without the 0.75 threshold is provided in the Appendix \ref{additional} Table \ref{table:additional_correlation}.

\subsection{Effect of individual latent coordinate perturbation on generated data}
\begin{figure}[h!]
    \centering
    \includegraphics[scale=1]{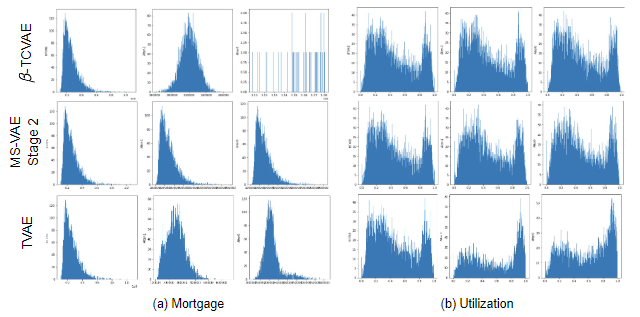}
      \caption{Effect of perturbation of the second coordinate of latent vectors on generated data. The left panel is for \textbf{Mortgage} feature, and the right is for \textbf{Utilization} feature. For each panel, in the first row, we perturb the second coordinate of the latent vector from $\beta$-TCVAE model. The second row is for the correlated latent factor $Z$ learned in the second stage of MS-VAE (from residual between original data and constructed data in stage one), and the third is for TVAE. In each panel, the first column is the original reconstruction data. In the second and third columns, we replace the second coordinate of the latent vector by zero vector and by constant vector $v_5$ respectively. Here $v_5$ is the vector with all entries equal $5$.}
      \label{fig:perturbation_effect}
\end{figure}

To visually estimate the effect of disentangled representation of the latent space for $\beta$-TCVAE and MS-VAE, we perturb a fixed coordinate of the latent vectors by a reasonably large margin. We then investigate the changes within the column features and compare with the changes produced by TVAE model under the same perturbation. Figure \ref{fig:perturbation_effect} shows the plot of the reconstructed data from different models when the corresponding latent vector's $2^{nd}$ coordinate is altered. Modifying the latent vector of $\beta$-TCVAE in the $2^{nd}$ coordinate only affects the feature \textbf{Mortgage} widely, while keeping the \textbf{Utilization}'s distribution nearly the same. This suggests a possible one-to-one relation between the latent vector's $2^{nd}$ coordinate and the \textbf{Mortgage} feature. In other words, the $2^{nd}$ coordinate seems to be a factor that dictates the \textbf{Mortgage} feature. Their correlation shown in Table \ref{table:correlation_table} further solidifies this hypothesis.

Varying the second component of the correlated factor $Z$ in the MS-VAE's second stage kept \textbf{Mortgage} and \textbf{Utilization} almost intact. The factors controlling these two features are thus learned via $\beta$-TCVAE in the first stage. As the core learning is done in the first stage, the factor $Z$ will additionally improve the reconstruction by taking into account other features that haven't been captured well in the first stage.

On the other hand, changing the latent variable from TVAE model has a slight impact on both features \textbf{Mortgage} and \textbf{Utilization}, but there seems to be no strong relation between these variables. This contrast indicates that $\beta$-TCVAE (first stage of MS-VAE) can potentially separate and recover distinct, informative factors of variations from credit data distribution, while TVAE does not achieve this effect.

\section{Conclusion}
\label{sec:conc}
In this work, we proposed a novel graphical model for \DR learning that, by virtue of being D-separated, overcomes the trade-off between reconstruction and disentanglement. To implement this graphical model, we introduced a multi-stage deep generative model, \simgan, that learns the independent and correlated latent variables in two separate stages of training. We showed that this model dramatically improves the reconstruction of preexisting \DR learning methods while maintaining the same level of disentanglement. Furthermore, our experiments for \simgangan and \simganflow (Appendices \ref{sec:gan_impl} and \ref{sec:flow_impl}, respectively) demonstrated that our framework is agnostic to both the level of supervision (e.g. unsupervised, weakly supervised, fully supervised) and to the underlying model (e.g. VAEs, GANs, normalizing flows). 

We have implemented the multi-stage model for generating synthetic credit tabular datasets as well. Compared to other leading benchmark models, MS-VAE maintains a harmonious balance between providing high-quality data and ease of understanding. The clarity regarding latent space features it provides is crucial for applications where transparent decision-making is key. In summary, this innovative model marks a notable advancement in simulating synthetic tabular data, merging enhanced performance with transparent interpretability.

While \simgan's two stage training is effective in alleviating the trade-off between learning disentangled representations and reconstruction quality in existing DR learning methods, in future, we aim to develop a single coherent DR model that does not require two separate training steps. 


\bibliography{citation}
\bibliographystyle{icml2021}

\begin{enumerate}

\item For all authors...
\begin{enumerate}
  \item Do the main claims made in the abstract and introduction accurately reflect the paper's contributions and scope?
    \answerYes{}
  \item Did you describe the limitations of your work?
    \answerYes{}
  \item Did you discuss any potential negative societal impacts of your work?
    \answerYes{}
  \item Have you read the ethics review guidelines and ensured that your paper conforms to them?
    \answerYes{}
\end{enumerate}

\item If you are including theoretical results...
\begin{enumerate}
  \item Did you state the full set of assumptions of all theoretical results?
    \answerNA{}
	\item Did you include complete proofs of all theoretical results?
    \answerNA{}
\end{enumerate}

\item If you ran experiments...
\begin{enumerate}
  \item Did you include the code, data, and instructions needed to reproduce the main experimental results (either in the supplemental material or as a URL)?
    \answerYes{}
  \item Did you specify all the training details (e.g., data splits, hyperparameters, how they were chosen)?
    \answerYes{}
	\item Did you report error bars (e.g., with respect to the random seed after running experiments multiple times)?
    \answerYes{}
	\item Did you include the total amount of compute and the type of resources used (e.g., type of GPUs, internal cluster, or cloud provider)?
    \answerNA{}
\end{enumerate}

\item If you are using existing assets (e.g., code, data, models) or curating/releasing new assets...
\begin{enumerate}
  \item If your work uses existing assets, did you cite the creators?
    \answerYes{}
  \item Did you mention the license of the assets?
    \answerNA{}
  \item Did you include any new assets either in the supplemental material or as a URL?
    \answerYes{}
  \item Did you discuss whether and how consent was obtained from people whose data you're using/curating?
    \answerNA{}
  \item Did you discuss whether the data you are using/curating contains personally identifiable information or offensive content?
    \answerNA{}
\end{enumerate}

\item If you used crowdsourcing or conducted research with human subjects...
\begin{enumerate}
  \item Did you include the full text of instructions given to participants and screenshots, if applicable?
    \answerNA{}
  \item Did you describe any potential participant risks, with links to Institutional Review Board (IRB) approvals, if applicable?
    \answerNA{}
  \item Did you include the estimated hourly wage paid to participants and the total amount spent on participant compensation?
    \answerNA{}
\end{enumerate}

\end{enumerate}

\clearpage
\newpage

\appendix
\onecolumn
\section{Traversal Results: Figures}
\label{sec:trf}
\begin{figure*}[hbt!]
\centering
\subfloat[Reconstruction on \textit{Cars3D}. Starting from top left, each triplet consists of (Target, $\beta$-TCVAE and \simgan) images.]{%
  \centering
  \begin{minipage}[b]{0.48\textwidth}
  \centering
  \includegraphics[width=\textwidth]{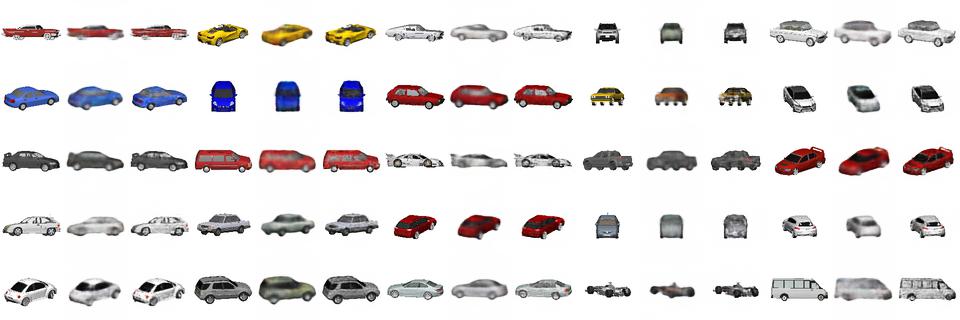}
  \end{minipage}
}\hfill
\subfloat[ Reconstruction on \textit{Small}NORB. Starting from top left, each triplet consists of (Target, $\beta$-TCVAE and \simgan) images.]{%
  \begin{minipage}[b]{0.48\textwidth}
  \centering
  \includegraphics[width=\textwidth]{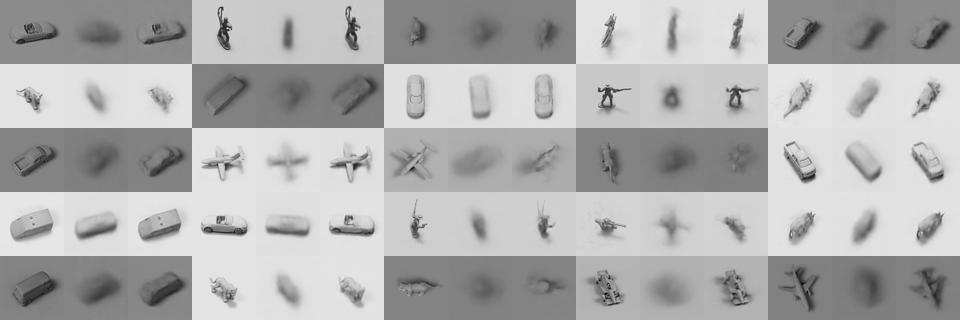}
  \end{minipage}
}

\subfloat[Traversal on \textit{Cars3D}. Odd row $\beta$-TCVAE, even row \simgan.]{%
  \begin{minipage}[b]{0.3\textwidth}
  \centering
  \includegraphics[width=\textwidth]{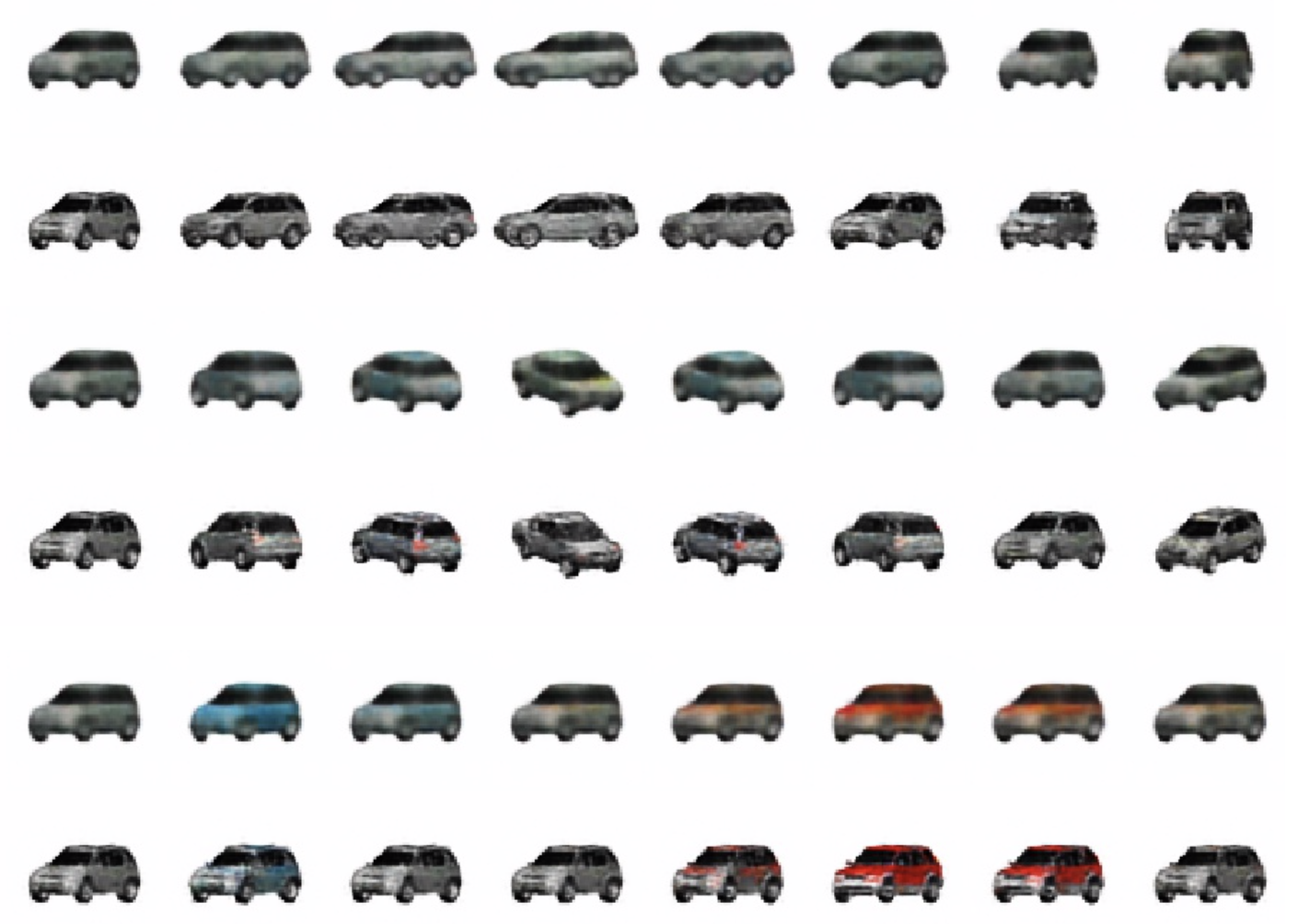}
  \end{minipage}
}\hfill
\subfloat[Traversal on \textit{Small}NORB. Odd row $\beta$-TCVAE, even row \simgan.]{%
  \begin{minipage}[b]{0.275\textwidth}
  \centering
  \includegraphics[width=\textwidth]{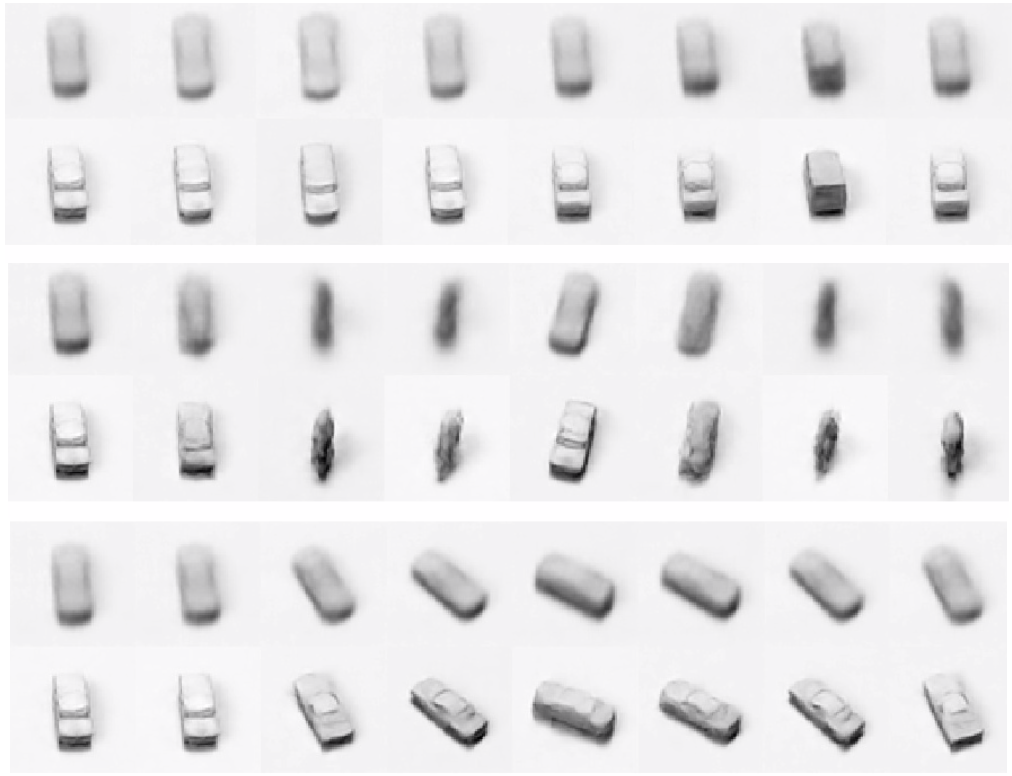}
  \end{minipage}
}\hfill
\subfloat[Z perturbation on \textit{Cars3D}.]{%
  \begin{minipage}[b]{0.3\textwidth}
  \centering
  \includegraphics[width=\textwidth]{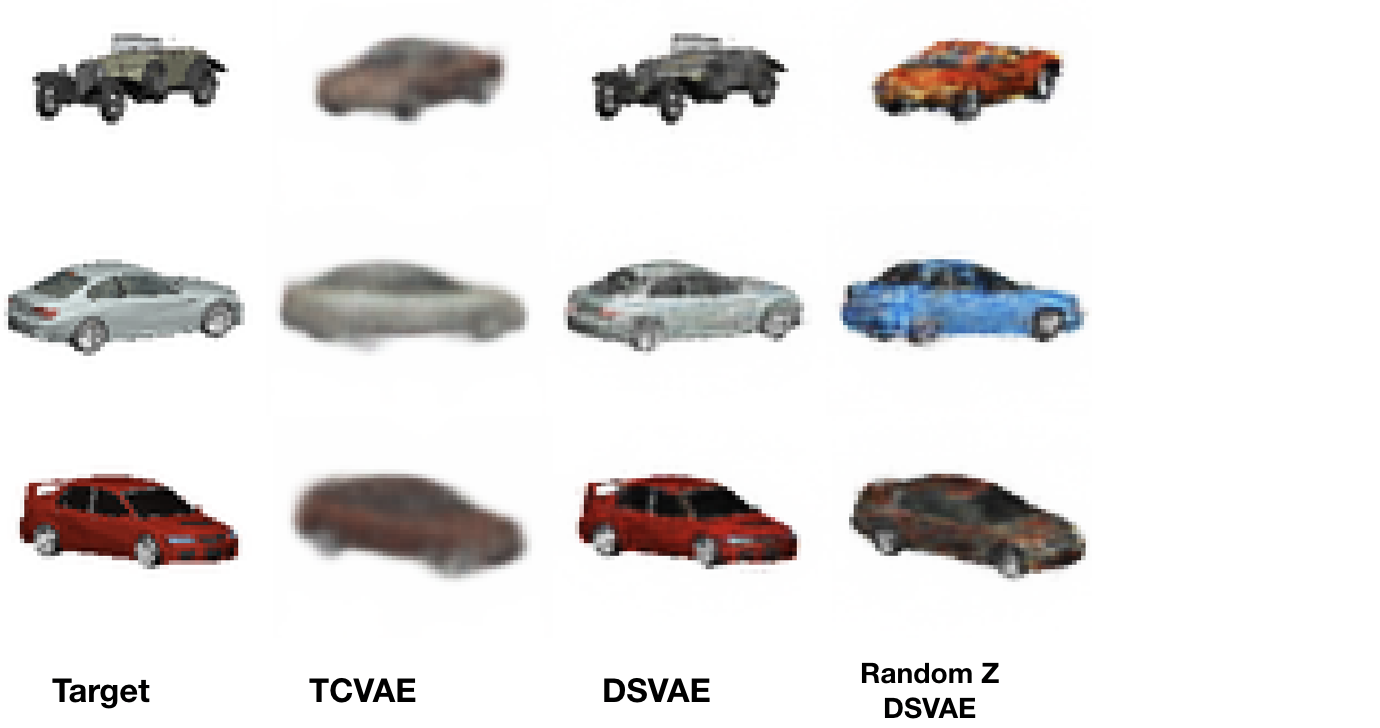}
  \end{minipage}
}

\caption{Qualitative results on the \textit{Cars3D} and \textit{Small}NORB datasets for $\beta$-TCVAE and \simgan.} 
\label{fig:qual}

\end{figure*}

\section{Choozing $C$ and $Z$}
\label{sec:c0}

We study the impact of the dimensionality of $C$ and $Z$ on the reported results in this section. To do so, we fix a value of $\beta$ and sweep over multiple values for $C$ and $Z$. 

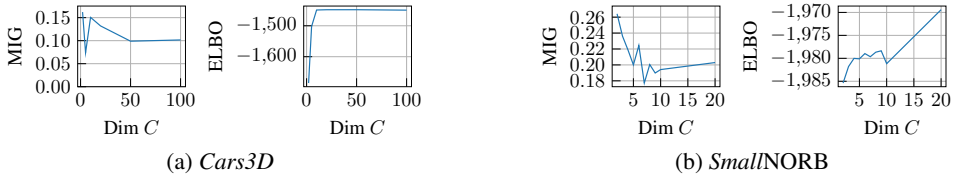
\begin{figure*}[hbt!]
\centering

\subfloat[\textit{Cars3D}]{%
  \centering
  \begin{minipage}[b]{0.5\textwidth}
  \centering
  \label{fig:sweep_c}
   \scalebox{0.75}{
\begin{tikzpicture}

\definecolor{color0}{rgb}{0.12156862745098,0.466666666666667,0.705882352941177}
\definecolor{color1}{rgb}{1,0.498039215686275,0.0549019607843137}
\definecolor{color2}{rgb}{0.172549019607843,0.627450980392157,0.172549019607843}
\definecolor{color3}{rgb}{0.83921568627451,0.152941176470588,0.156862745098039}

\begin{groupplot}[
    group style={
        group size= 2 by 1,
        horizontal sep=0.3\linewidth
    }, 
    width=\linewidth,
]

    \nextgroupplot[
width=0.5\linewidth,
y tick label style={
        /pgf/number format/.cd,
            fixed,
            fixed zerofill,
            precision=2,
        /tikz/.cd
    },
tick pos=left,
x grid style={white!69.0196078431373!black},
xlabel={Dim $C$},
xmajorgrids,
xmin=-2.9, xmax=104.9,
xtick style={color=black},
y grid style={white!69.0196078431373!black},
ylabel={MIG},
ymajorgrids,
ymin=0,
ymax=0.175185583601943,
ytick style={color=black}
]
\addplot [semithick, color0]
table {%
2 0.162069634006412
5 0.0738265213638457
10 0.150525104305477
20 0.131993554068622
50 0.0988466141613275
100 0.101574796266581
};

    \nextgroupplot[
width=0.5\linewidth,
tick pos=left,
x grid style={white!69.0196078431373!black},
xlabel={Dim $C$},
xmajorgrids,
xmin=-2.9, xmax=104.9,
xtick style={color=black},
y grid style={white!69.0196078431373!black},
ylabel={ELBO},
ymajorgrids,
ymin=-1697.03364257812, ymax=-1436.22783203125,
ytick style={color=black}
]
\addplot [semithick, color0]
table {%
2 -1685.17883300781
5 -1502.18664550781
10 -1448.99719238281
20 -1448.15539550781
50 -1448.08264160156
100 -1449.47338867188
};
\end{groupplot}

\end{tikzpicture}}
  \end{minipage}
}
\subfloat[\textit{Small}NORB]{%
  \begin{minipage}[b]{0.5\textwidth}
  \centering
  \label{fig:sweep_sn}
  \scalebox{0.75}{
\begin{tikzpicture}

\definecolor{color0}{rgb}{0.12156862745098,0.466666666666667,0.705882352941177}
\definecolor{color1}{rgb}{1,0.498039215686275,0.0549019607843137}
\definecolor{color2}{rgb}{0.172549019607843,0.627450980392157,0.172549019607843}
\definecolor{color3}{rgb}{0.83921568627451,0.152941176470588,0.156862745098039}

\begin{groupplot}[
    group style={
        group size= 2 by 1,
        horizontal sep=0.3\linewidth
    }, 
    width=\linewidth,
]

    \nextgroupplot[
width=0.5\linewidth,
y tick label style={
        /pgf/number format/.cd,
            fixed,
            fixed zerofill,
            precision=2,
        /tikz/.cd
    },
tick pos=left,
x grid style={white!69.0196078431373!black},
xlabel={Dim $C$},
xmajorgrids,
xmin=1.1, xmax=20.9,
xtick style={color=black},
y grid style={white!69.0196078431373!black},
ylabel={MIG},
ymajorgrids,
ymin=0.172616542957597, ymax=0.273957241192525,
ytick style={color=black}
]
\addplot [semithick, color0]
table {%
2 0.264183580238954
3 0.236120073032813
4 0.218848946888343
5 0.200051825475284
6 0.224414740691178
7 0.177222938331912
8 0.20041249342135
9 0.189834531913335
10 0.194140014950762
20 0.202951411844625
};

    \nextgroupplot[
width=0.5\linewidth,
tick pos=left,
x grid style={white!69.0196078431373!black},
xlabel={Dim $C$},
xmajorgrids,
xmin=1.1, xmax=20.9,
xtick style={color=black},
y grid style={white!69.0196078431373!black},
ylabel={ELBO},
ymajorgrids,
ymin=-1986.18184204102, ymax=-1968.6112487793,
ytick style={color=black}
]
\addplot [semithick, color0]
table {%
2 -1985.38317871094
3 -1981.79235839844
4 -1980.01708984375
5 -1980.11267089844
6 -1978.9638671875
7 -1979.64794921875
8 -1978.66613769531
9 -1978.35876464844
10 -1981.16381835938
20 -1969.40991210938
};
\end{groupplot}

\end{tikzpicture}}
  \end{minipage}
}
\caption{MIG and ELBO for different dimensionality of $C$ for \textit{Cars3D} and \textit{Small}NORB at $\beta = 4$}
\end{figure*}

{\bf Choosing C:} For $C$, we sweep over the range $[ 1-10, 20, 50 ]$ for both the datasets. Figure \ref{fig:sweep_c} and \ref{fig:sweep_sn} show the MIG and ELBO for the same. We can see that $| C| = 10$ achieves a good combination of MIG and ELBO for the \textit{Cars3D} dataset. While there is high variance in the MIG for \textit{Small}NORB, $| C| = 10$ performs reasonably well. This is also consistent with the large scale study in \citep{icmlbest}, which suggests that a value of 10 for the dimensionality of $C$ is optimal for \DR learning.

\begin{figure*}[htb!]
\centering
\subfloat[\textit{Cars3D}]{%
  \centering
  \begin{minipage}[b]{\textwidth}
  \centering
  \label{fig:cz_c}
    \scalebox{0.75}{
\begin{tikzpicture}

\definecolor{color0}{rgb}{0.12156862745098,0.466666666666667,0.705882352941177}
\definecolor{color3}{rgb}{1,0.498039215686275,0.0549019607843137}
\definecolor{color1}{rgb}{0.172549019607843,0.627450980392157,0.172549019607843}
\definecolor{color2}{rgb}{0.83921568627451,0.152941176470588,0.156862745098039}

\begin{groupplot}[
    group style={
        group size= 3 by 1,
        horizontal sep=0.1\linewidth
    }, 
    width=\linewidth,
]

    \nextgroupplot[
width=0.33\linewidth,
tick pos=left,
x grid style={white!69.0196078431373!black},
xlabel={Dim $Z$},
xmajorgrids,
xmin=0.75, xmax=94.25,
xtick style={color=black},
y grid style={white!69.0196078431373!black},
ylabel={FID},
ymajorgrids,
ymin=12.9120583191342, ymax=154.282348139237,
ytick style={color=black}
]
\addplot [semithick, color0]
table {%
5 138.520178747568
10 137.925926899141
50 147.846591049483
90 147.856425874687
};
\addplot [semithick, color1]
table {%
5 68.8872185341214
10 55.8256370145332
50 53.5214820282886
90 75.5850636940193
};
\addplot [semithick, color2]
table {%
5 22.0580487374536
10 22.8694422818883
50 19.6491627897661
90 19.3379805836843
};

    \nextgroupplot[
    width=0.33\linewidth,
    legend cell align={left},
    legend columns=10,
    legend style={fill opacity=0.8, draw opacity=1, text opacity=1,
        /tikz/every even column/.append style={column sep=0.5cm},
        at={(-0.18\linewidth,1.1)}, anchor=south west, draw=white!80!black},
tick pos=left,
x grid style={white!69.0196078431373!black},
xlabel={Dim $Z$},
xmajorgrids,
xmin=0.75, xmax=94.25,
xtick style={color=black},
y grid style={white!69.0196078431373!black},
ylabel={MIG},
ymajorgrids,
ymin=0.0885738435465964, ymax=0.165436480881288,
ytick style={color=black}
]
\addplot [semithick, color0]
table {%
5 0.12798620918838
10 0.127021726789503
50 0.150024759548161
90 0.137018629587691
};
\addlegendentry{$\beta$-TCVAE}
\addplot [semithick, color1]
table {%
5 0.0920675997890824
10 0.105739455658182
50 0.130593561130993
90 0.0762672632098107
};
\addlegendentry{\simgan (Random $Z$)}
\addplot [semithick, color2]
table {%
5 0.131971883740099
10 0.13118463390344
50 0.156109284289745
90 0.142575857986845
};
\addlegendentry{\simgan}

    \nextgroupplot[
width=0.33\linewidth,
tick pos=left,
x grid style={white!69.0196078431373!black},
xlabel={Dim $Z$},
xmajorgrids,
xmin=0.75, xmax=94.25,
xtick style={color=black},
y grid style={white!69.0196078431373!black},
ylabel={MI},
ymajorgrids,
ymin=0.379542583107672, ymax=0.715623359031943,
ytick style={color=black}
]
\addplot [semithick, color0]
table {%
5 0.641165025358618
10 0.617534227589804
50 0.631424225292956
90 0.629882131089608
};
\addplot [semithick, color1]
table {%
5 0.427770325835723
10 0.482230713890059
50 0.486571487777313
90 0.39481898201332
};
\addplot [semithick, color2]
table {%
5 0.700346960126295
10 0.677581224414183
50 0.683315513153944
90 0.684570985760045
};

\end{groupplot}

\end{tikzpicture}}
  \end{minipage}
}\\
\subfloat[\textit{Small}NORB]{%
  \begin{minipage}[b]{\textwidth}
  \centering
  \label{fig:cz_sn}
  \scalebox{0.75}{
\begin{tikzpicture}

\definecolor{color0}{rgb}{0.12156862745098,0.466666666666667,0.705882352941177}
\definecolor{color3}{rgb}{1,0.498039215686275,0.0549019607843137}
\definecolor{color1}{rgb}{0.172549019607843,0.627450980392157,0.172549019607843}
\definecolor{color2}{rgb}{0.83921568627451,0.152941176470588,0.156862745098039}

\begin{groupplot}[
    group style={
        group size= 3 by 1,
        horizontal sep=0.1\linewidth
    }, 
    width=\linewidth,
]

    \nextgroupplot[
width=0.33\linewidth,
tick pos=left,
x grid style={white!69.0196078431373!black},
xlabel={Dim $Z$},
xmajorgrids,
xmin=-2.4, xmax=94.4,
xtick style={color=black},
y grid style={white!69.0196078431373!black},
ylabel={FID},
ymajorgrids,
ymin=50.8740897918193, ymax=212.463021390033,
ytick style={color=black}
]
\addplot [semithick, color0]
table {%
2 202.30172161472
5 202.400245740936
10 202.581020117056
20 202.283687970578
50 202.154035752617
90 202.483333034989
};
\addplot [semithick, color1]
table {%
2 130.883114301188
5 131.795864081081
10 133.330247754586
20 149.647131967225
50 189.438285539984
90 205.118069953751
};
\addplot [semithick, color2]
table {%
2 86.3463975307788
5 64.7294834428646
10 58.2190412281018
20 76.8802955699292
50 95.3069049657454
90 96.4063417958834
};
    
    \nextgroupplot[
    width=0.33\linewidth,
    legend cell align={left},
    legend columns=4,
    legend style={fill opacity=0.8, draw opacity=1, text opacity=1, at={(-0.1\linewidth,1.1)}, anchor=south west, draw=white!80!black},
tick pos=left,
x grid style={white!69.0196078431373!black},
xlabel={Dim $Z$},
xmajorgrids,
xmin=-2.4, xmax=94.4,
xtick style={color=black},
y grid style={white!69.0196078431373!black},
ylabel={MIG},
ymajorgrids,
ymin=0.116458781075177, ymax=0.240924229486803,
ytick style={color=black}
]
\addplot [semithick, color0]
table {%
2 0.193635726781265
5 0.192976157270337
10 0.193254751531815
20 0.193847311829718
50 0.194140014950762
90 0.193291356061669
};
\addplot [semithick, color1]
table {%
2 0.136177686631725
5 0.155620603756356
10 0.150132886494635
20 0.14406356719492
50 0.134580256335974
90 0.122116301457524
};
\addplot [semithick, color2]
table {%
2 0.194228127284206
5 0.197363039876536
10 0.199506093623367
20 0.194430122763156
50 0.1966394233463
90 0.196921706604307
};

    \nextgroupplot[
width=0.33\linewidth,
tick pos=left,
x grid style={white!69.0196078431373!black},
xlabel={dim z},
xmajorgrids,
xmin=-2.4, xmax=94.4,
xtick style={color=black},
y grid style={white!69.0196078431373!black},
ylabel={MI},
ymajorgrids,
ymin=0.182890892593078, ymax=0.433052744953614,
ytick style={color=black}
]
\addplot [semithick, color0]
table {%
2 0.320456565687966
5 0.321746745640198
10 0.319831391145855
20 0.319457654294678
50 0.32181531056121
90 0.321191904461734
};
\addplot [semithick, color1]
table {%
2 0.224564617527215
5 0.23593419971572
10 0.235828089831823
20 0.227947222878996
50 0.225119592218419
90 0.194261885882193
};
\addplot [semithick, color2]
table {%
2 0.353843981901443
5 0.405927891504079
10 0.421681751664499
20 0.365233482009323
50 0.338683413901801
90 0.34649899504271
};

\end{groupplot}

\end{tikzpicture}}
  \end{minipage}
}
\label{fig:sweep-z}
\caption{MIG,FID and MI plots for \simgan as a function of the dimensionality of $Z$ for \textit{Cars3D} and \textit{Small}NORB ($\beta=4$).}
\end{figure*}
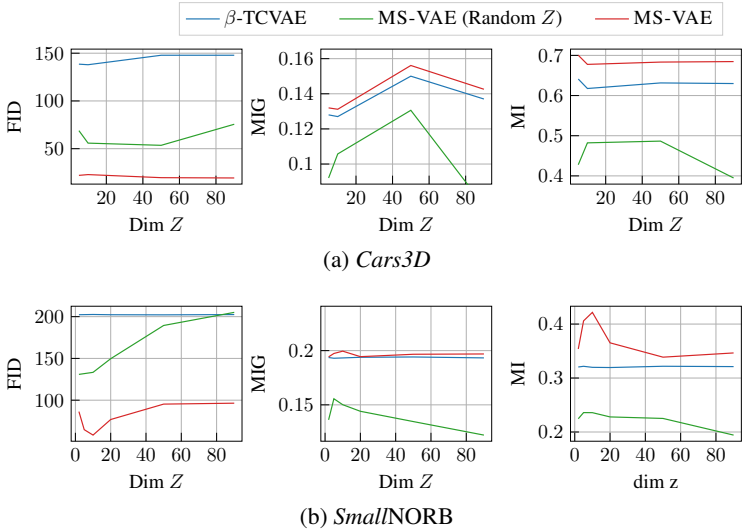

{\bf Choosing Z:} Similarly for $Z$, we sweep over a set of candidate values $[ 5, 10, 50, 90]$ and evaluate the FID and MIG. For evaluating the FID, we consider both reconstructions (by sampling $Z$ from the posterior) and 'Random Z' (by sampling a random $Z$ from the prior). We also look at the MI \ref{sec:mig_desc}. We find smaller values, 5 and 10 for \textit{Cars3D} and \textit{Small}NORB, respectively, achieve a good combination of all the three quantities of interest. The results are shown in \ref{fig:sweep-z}.

\begin{figure}[H]
	\centering
	\includegraphics[width=1\linewidth]{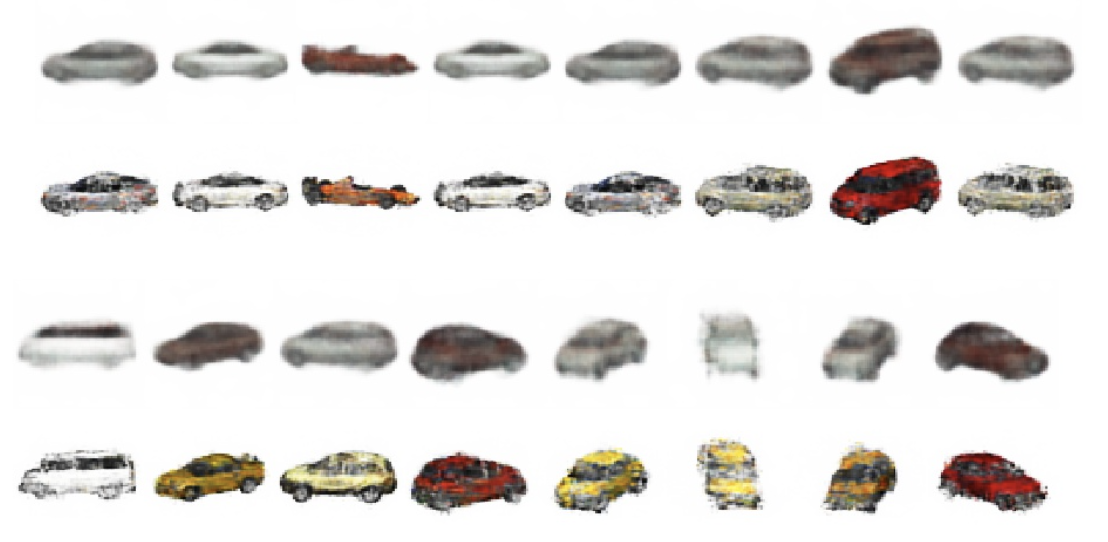}
	\caption{Traversals for cars for small $C$ (c = 2, z = 8 and $\beta$ = 4). Odd rows $\beta$-TCVAE and even rows \simgan. We can see the $Y$ images are blurry and various factors are entangled.}
	\label{fig:tr_cars_c2_z8}
\end{figure}

\begin{figure}[H]
	\centering
	\includegraphics[width=1\linewidth]{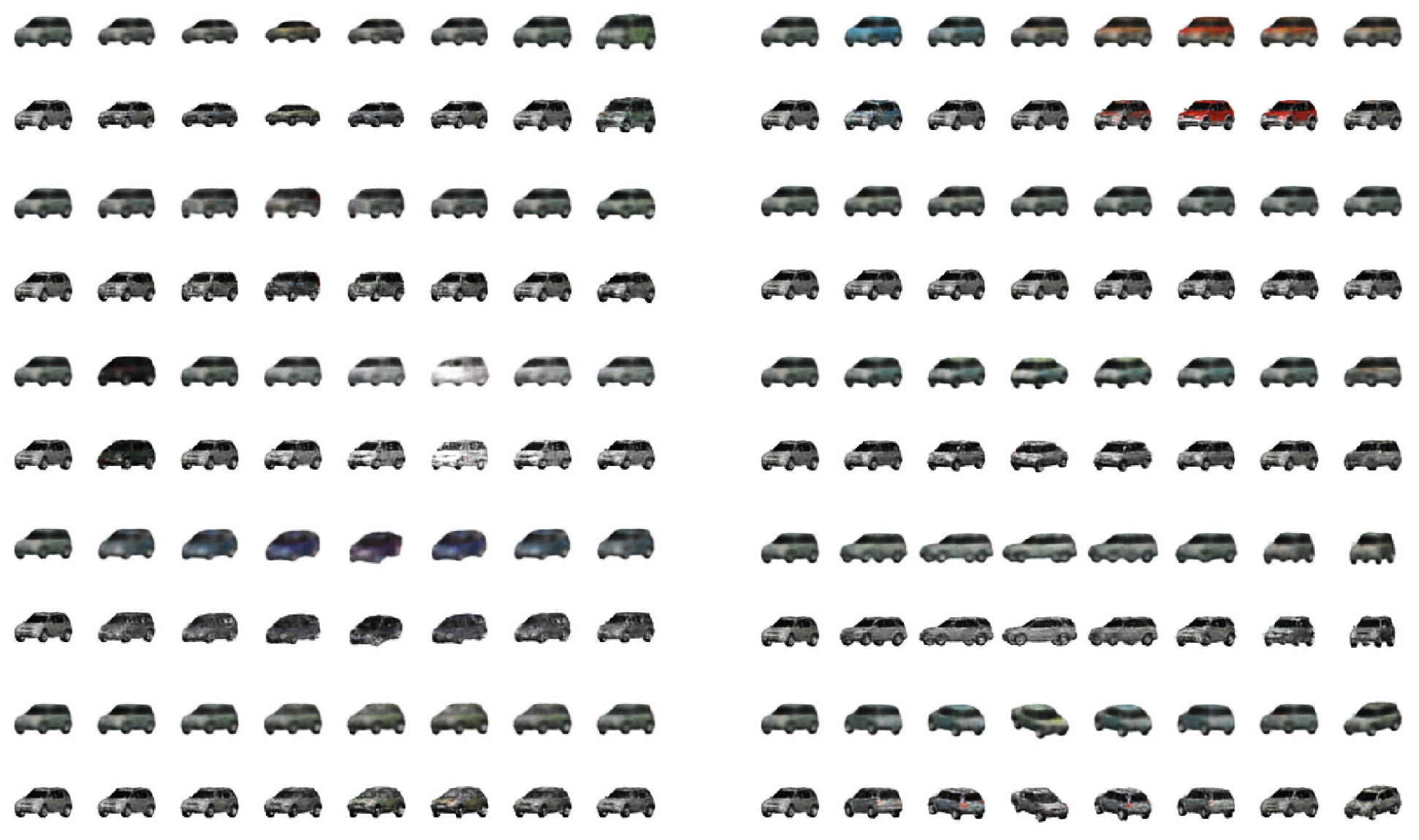}
	\caption{Traversals for cars large $C$ (c = 10, z = 5 and $\beta$ = 4). Odd rows $\beta$-TCVAE and even rows \simgan. Larger $C$ achieves greater disentanglement and $Z$ further refines the images.}
	\label{fig:tr_cars_c10_z5}
\end{figure}

\begin{figure}[H]
	\centering
	\includegraphics[width=1\linewidth]{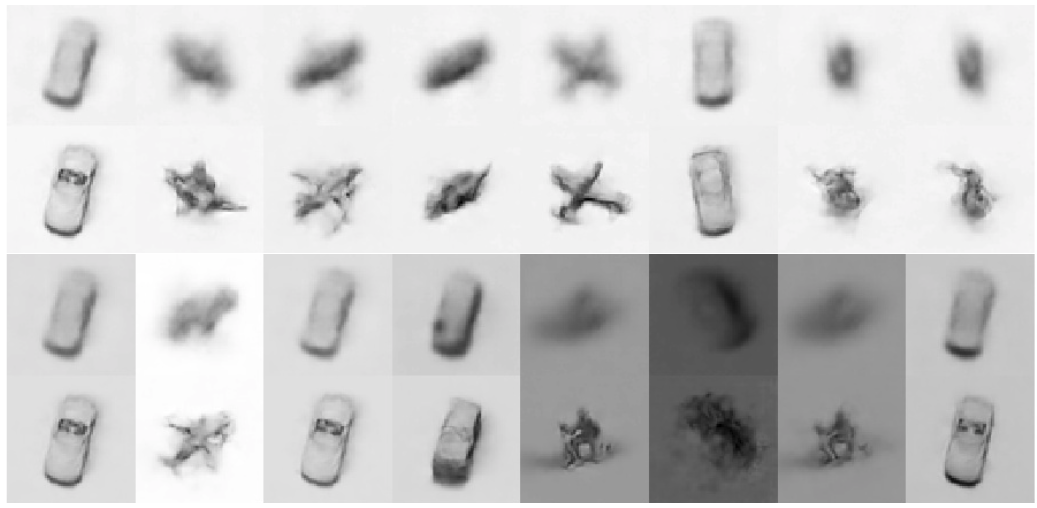}
	\caption{Traversals for SmallNORB c = 2, z = 8 and $\beta$ = 4. Odd rows $\beta$-TCVAE and even rows \simgan.}
	\label{fig:tr_norb_c2_z8}
\end{figure}

\begin{figure}[H]
	\centering
	\includegraphics[width=1\linewidth]{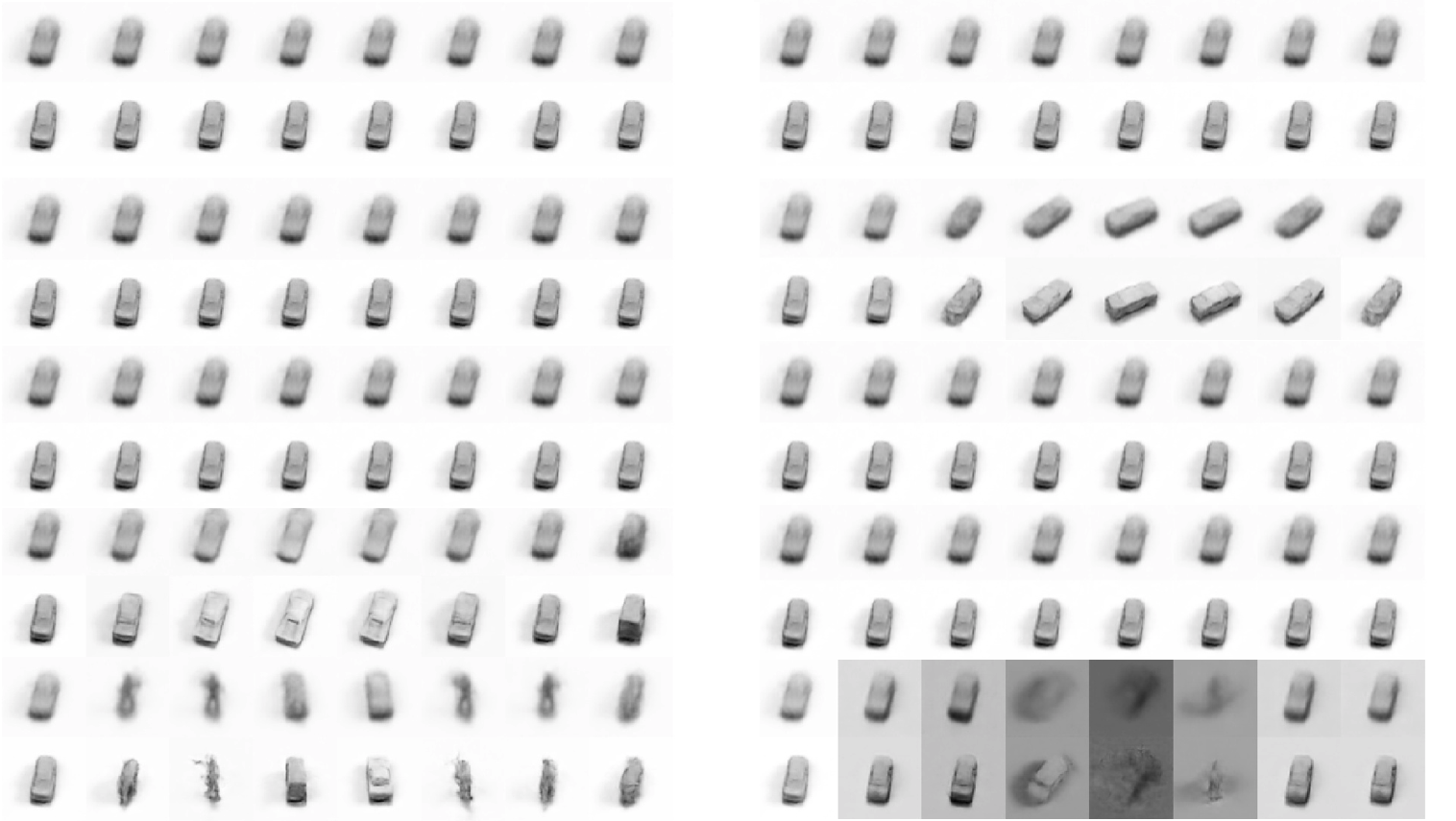}
	\caption{Traversals for SmallNORB c = 10, z = 10 and $\beta$ = 6. Odd rows $\beta$-TCVAE and even rows \simgan.}
	\label{fig:tr_norb_c10_z10}
\end{figure}

Qualitatively, Figure \ref{fig:tr_cars_c2_z8} and Figure \ref{fig:tr_cars_c10_z5} show the effect of varying the dimensionality of  $C$ and $Z$. For a small $C$ (Figure \ref{fig:tr_cars_c2_z8}), the underlying disentanglement method does a poor job at disentangling the meaningful latent variables, which further results in blurry images. In this case, $Z$ has substantial control over the image---it adds various missing details to $Y$ such as color. In contrast, Figure \ref{fig:tr_cars_c10_z5} shows traversals for a high value of $C$. Now, each individual dimension of C controls fewer independent factors, the $Y$ images are relatively sharper, and adding $Z$ improves them further. Similar conclusions hold for \textit{SmallNORB} as shown in Fig. \ref{fig:tr_norb_c10_z10} and \ref{fig:tr_norb_c2_z8}.

\section{CelebA Results}
\label{sec:celeba}
\begin{figure*}[h]
	\centering
	\includegraphics[scale=.7]{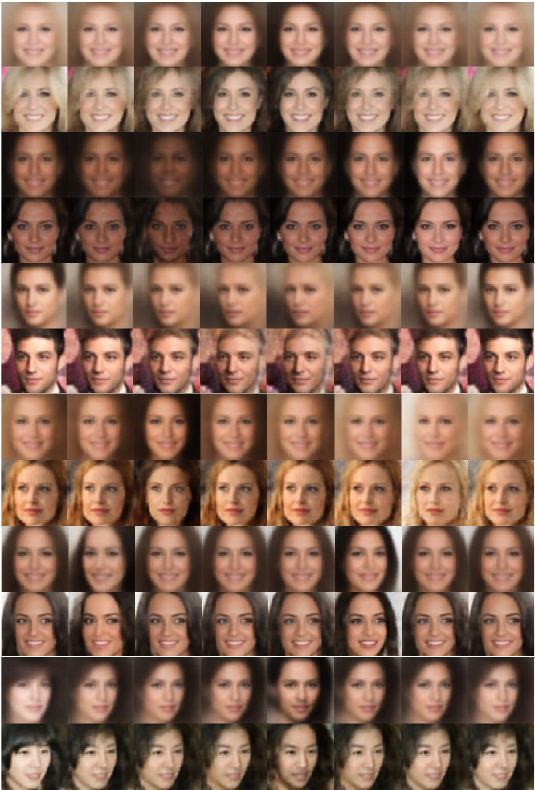}
	\caption{CelebA Latent traversal on $C$. Odd rows are $\beta$-TCVAE and even rows are \simgan.}
	\label{fig:celeba_traversal}
\end{figure*}

\begin{figure*}[h]
	\centering
	\includegraphics[scale=.35]{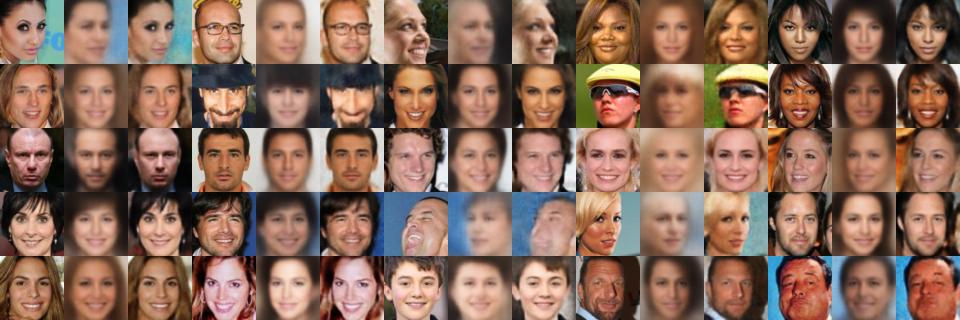}
	\caption{CelebA Reconstruction plots.}
	\label{fig:celeba_reconstruction}
\end{figure*}

Please see Figure \ref{fig:celeba_traversal} and Figure \ref{fig:celeba_reconstruction}.

\section{\simgan for Simple Pendulum}

\label{sec:pen}
A simple, 2D pendulum is comprised of a bob with mass $m$ attached to a string of length $L$ where the mass of the string is negligible in comparison to $m$. The primary forces acting upon the bob are the gravitational force, the tension force of the string, and the damping force from air resistance. Let $\theta$ be the angular displacement of the weight (the corresponding angle with respect to the y-direction). The pendulum's motion can be completely described as $\frac{d^2\theta}{dt^2} = -\frac{b}{m}\frac{d\theta}{dt} -\frac{g}{L}\sin{\theta}$,
where $b$ is the damping coefficient and $g$ is the gravitational acceleration constant (the initial angular displacement/velocity are constant). 

In this toy example, we assume that, given the length $L$, the damping coefficient $B$, and the mass $M$, the angular displacement of a pendulum is generated by a noiseless simulator $S: (L, M, B) \rightarrow \theta$ where $\theta \coloneqq \{\theta_0, \theta_1, ..., \theta_T\}$, $\theta_i \in [-\pi,\pi]$, and $T = 100$. Using samples from $S$ where only $L$ and $B$ are known a priori, we aim to learn a hierarchical, generative model for the pendulum where \textit{all generative factors are completely disentangled}. With \simgan, we can achieve this by learning the graphical model in Figure \ref{fig:pendulum_graphical}.
We define $Y_0$ as the undamped angular displacement and $Y_1$ as the damped angular displacement with fixed mass ($M=1$). 

We train our model in three separate steps. First, we use paired data $\{L^i, {y^i}_0\}$ to learn the subgraph $L \rightarrow Y_0$ as a conditional Gaussian distribution (how the pendulum length affects the angular displacement). Second, we use the paired data $\{b^i, {y^i}_0, {y^i}_1\}$ to learn $(B, Y_0) \rightarrow Y_1$ as a conditional Gaussian distribution (how the damping coefficient affects the angular displacement for a fixed mass). Finally, we use the paired data $\{{y^i}_1, x^i\}$ to learn the subgraph $(Z, {y^i}_1) \rightarrow \theta$, where $Z$ models the residual between $Y_1$ and the true observations $\theta$ (this subgraph is realized as a VAE). Through construction and training, the observed variable $L$ should control the length of the pendulum, the observed variable $B$ should control the damping coefficient, and the latent variable $\boldsymbol{Z}$ should be correlated with the mass $M$. \footnote{For simplicity, we don't model the data as time series explicitly but rather as a fixed length, 1D vector.} Note that while we have labeled data for the observed variables $L$ and $B$, we never have access to the mass $M$. Despite this, the latent variable $Z$ can still learn to model the effect of $M$ by modeling the residual.

\begin{figure*}[htp]

\subfloat[Traversing the length $l$ from 1.0 to 3.0 while fixing decay constant $b$ and $z$]{%
  \centering
  \begin{minipage}[b]{\textwidth}
  \centering
  \includegraphics[height=2in, width=6in]{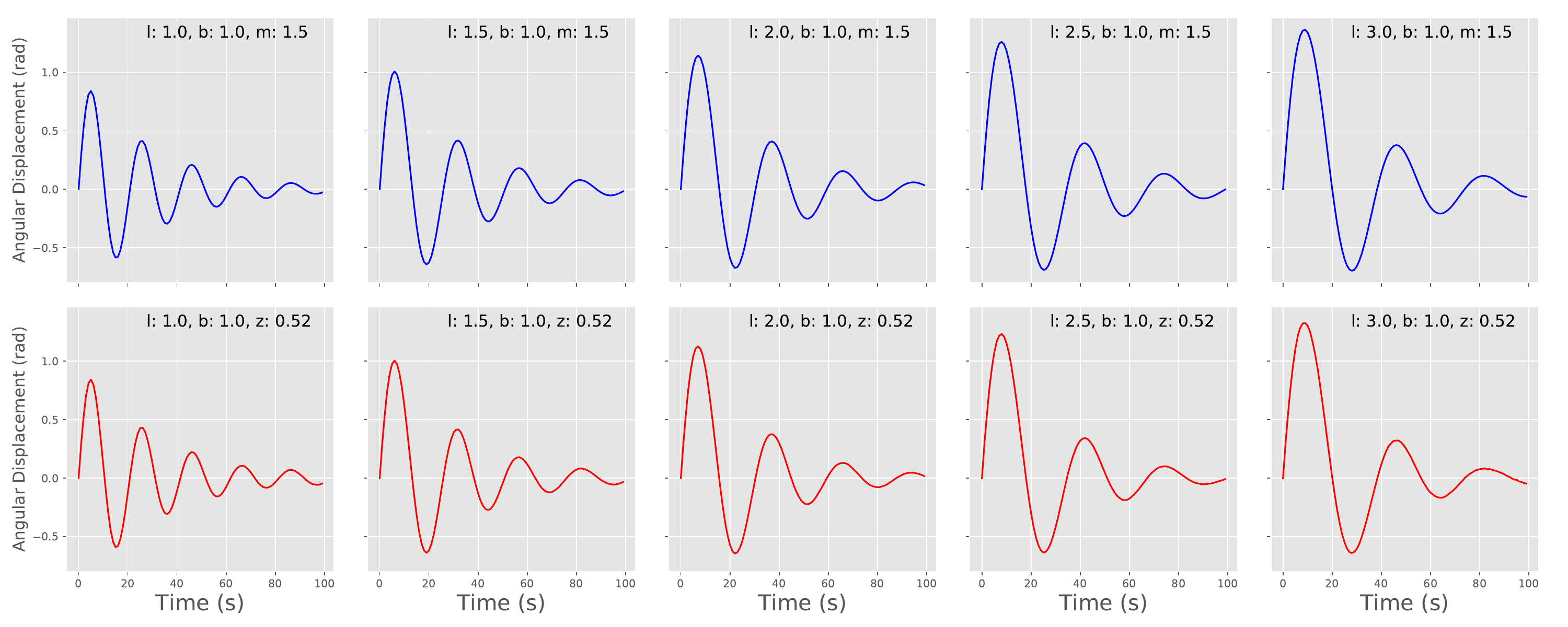}
  \end{minipage}
}

\subfloat[Traversing decay constant $b$ from .1 to 2.0 while fixing the length $l$ and $z$]{%
  \begin{minipage}[b]{\textwidth}
  \centering
  \includegraphics[height=2in, width=6in]{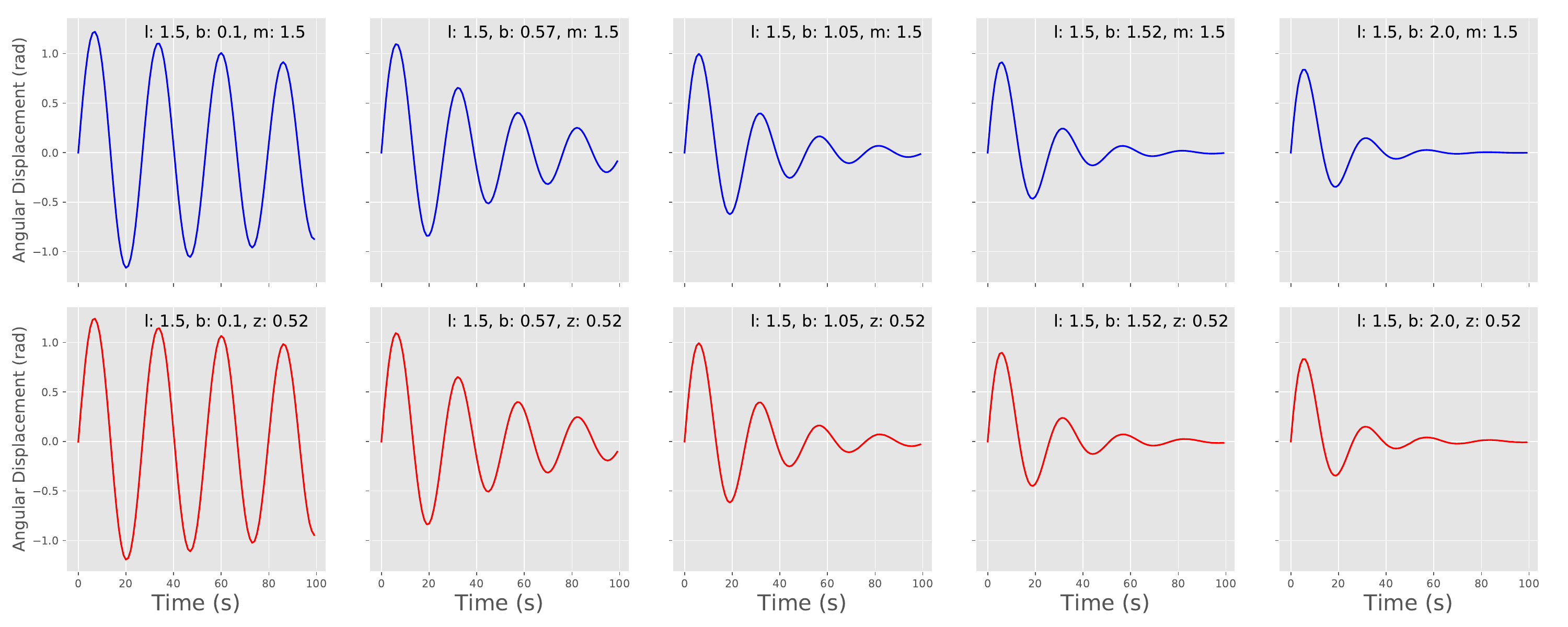}
  \end{minipage}
}

\subfloat[Traversing the mass $m$ from .1 to 2.0 (top) and traversing the corresponding $z$ values (-1.5 to 1.44) while fixing the length $l$ and decay constant $b$.]{%
  \begin{minipage}[b]{\textwidth}
  \centering
  \includegraphics[height=2in, width=6in]{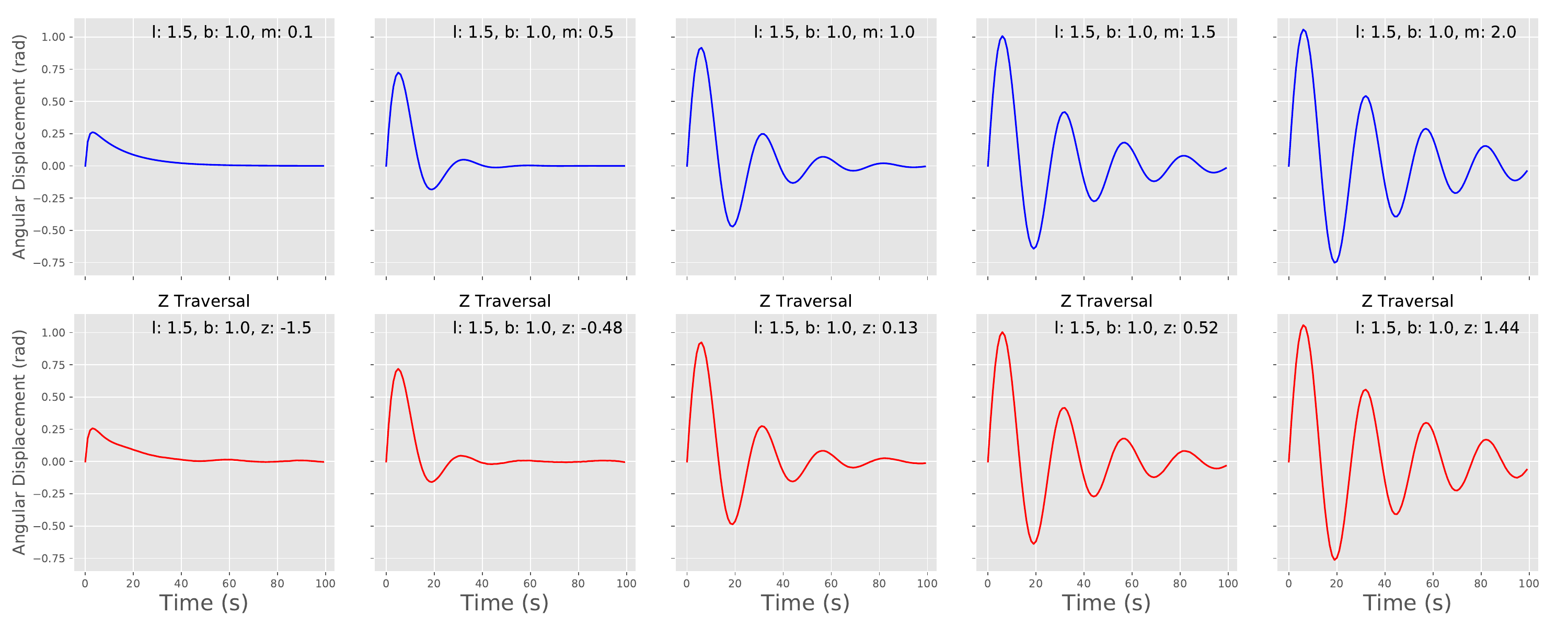}
  \end{minipage}
}
\caption{Latent traversals for the pendulum model. Odd rows (blue) are actual data from the simulator and even rows (red) are from \simgan. In this example, \simgan learns each of the three latent variables in a disentangled manner without any supervision for the mass control variate. We choose the $z$ values that are plotted by examining the posterior over $z$ and choosing $z$ values that correspond to the given masses ($z$ is 1-Dimensional so this is straightforward).} 
\label{fig:pendulum}
\end{figure*}

\begin{figure*}[t!]
	\centering
	\includegraphics[scale=.5]{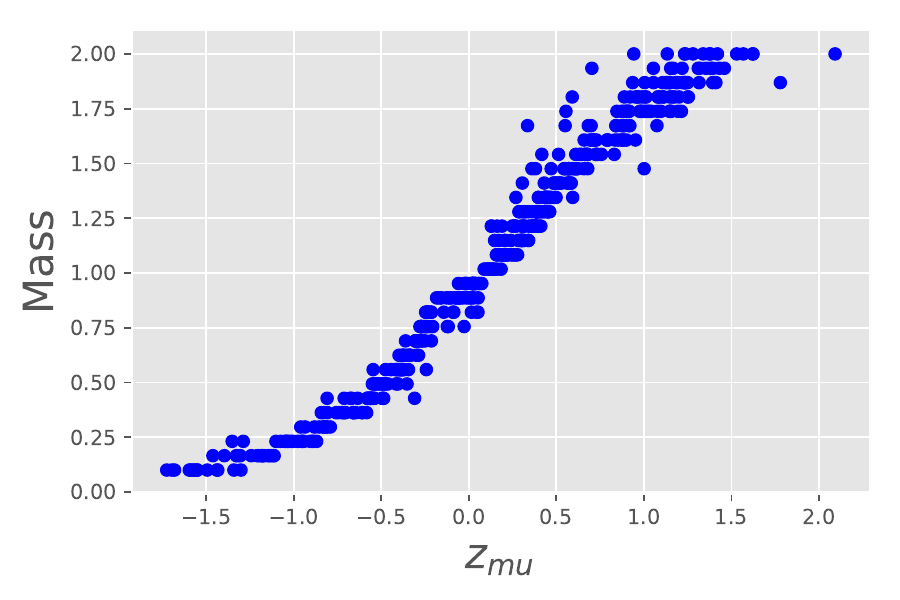}
	\caption{Correlation between $z$ and the mass when modeling a simple pendulum. In this plot, we show 500 1-Dimensional posterior means for different $\theta$ observations to illustrate how mass and the latent variable Z are heavily correlated after training.}
	\label{fig:z_regression}
\end{figure*}

The results are shown in Figures \ref{fig:pendulum} and \ref{fig:z_regression}. As can be seen, \simgan is able to learn completely disentangled generative factors using a combination of supervision and residual modeling. The final learned latent variable Z is shown to be heavily correlated with the mass of the pendulum, highlighting how \simgan's learned latent variables can be directly related with informative physical parameters. This simple toy example highlights how \simgan's training paradigm can be used to model physics-based simulators. 

\begin{figure}[h!]
    \centering
    \begin{tikzpicture}

  \node[obs]                                  (x)   {$\boldsymbol{\theta}$};
  \node[obs,     left=of x,   xshift=0.6cm]   (y_1) {$\mathbf{y_1}$};
  \node[obs,     left=of y_1, xshift=0.6cm]   (y_0) {$\mathbf{y_0}$};
  \node[latent, above=of x,   yshift=-0.6cm]  (z)   {$\mathbf{z}$};
  \node[obs, above=of y_1, yshift=-0.6cm]  (b)   {$\mathbf{b}$};
  \node[obs, above=of y_0, yshift=-0.6cm]  (L)   {$\mathbf{L}$};

  \edge {y_1,z} {x} ; %
  \edge {y_0,b}   {y_1} ; %
  \edge {L}   {y_0} ; %

  \plate {} {(x)(y_1)(y_0)(z)(b)(L)} {$N$} ;

\end{tikzpicture}
    \caption{\simgan graphical model of a simple pendulum. $\theta$ is the angular displacement; $L$ is the length; $B$ is the damping coefficient. After learning and by construction, the latent variable Z is correlated with the pendulum's mass.}\label{fig:pendulum_graphical}
\end{figure}

\section{FLOW-based Implementation (\simganflow)}
\label{sec:flow_impl}

\begin{figure}[htb!]
	\centering
	\includegraphics[width=0.5\linewidth]{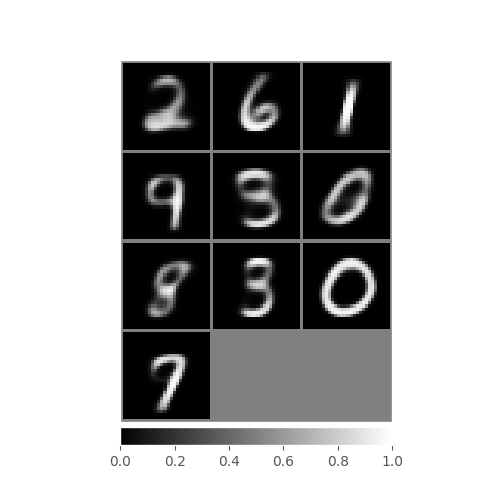}
	\caption{The means of the Gaussian mixture part of \simganflow on MNIST.}
	\label{fig:ds_flow-means}
\end{figure}

\begin{figure}[htb!]
	\centering
	\includegraphics[scale=.5]{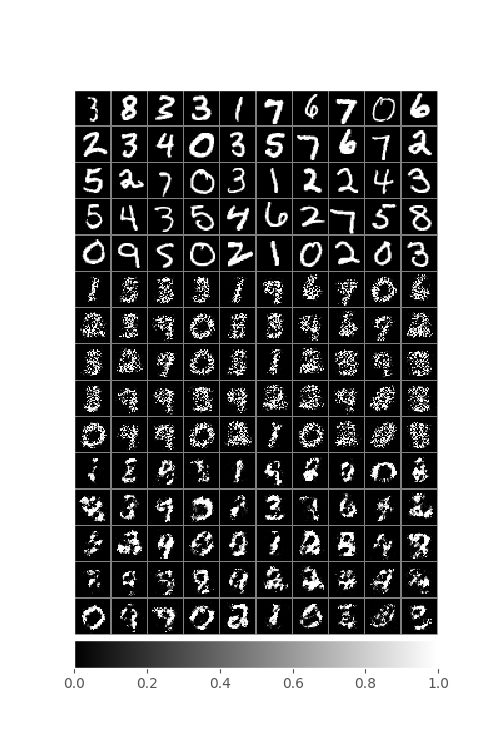}
	\caption{\simganflow on MNIST. First 5 rows are real data, next five rows are samples from the Gaussian mixture and last five rows are from the conditional flow.}
	\label{fig:ds_flow-recon}
\end{figure}

As an illustration, in this section we present the \simganflow model on the MNIST digit dataset. Here the subgraph $C \rightarrow Y$ is realized as a Gaussian mixture model and the subgraph $(Y, Z) \rightarrow X$ is implemented as a conditional coupling flow \citep{dinh2016density}. We fit a mixture of 10 Gaussians whose means are shown in Figure~\ref{fig:ds_flow-means}. 
Figure~\ref{fig:ds_flow-recon} shows the real data, samples from the Gaussian mixture and the conditional flow.
As expected, the log density of the overall model improves from -1585.086 to -1356.089 once the FLOW is applied to the output of the mixture model. All while maintaining the ability of controllable sampling.

\subsection{Training using Means or Samples}

\begin{figure}
    \centering
    \scalebox{0.75}{\input{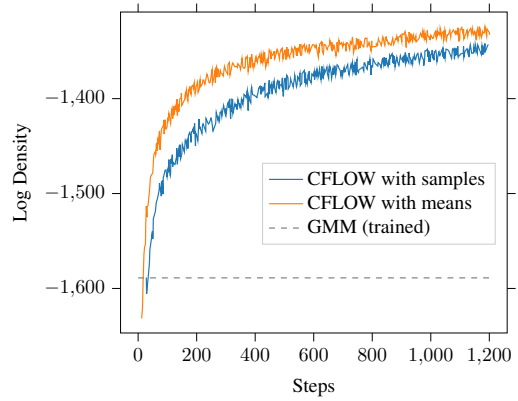}}
    \caption{Conditional log density during training for different types of $Y$ (samples v.s. means).}
    \label{fig:samples-vs-means}
\end{figure}

As mentioned in previous sections,
when training the conditional model,
we can use either the means from the likelihood model of the $C \rightarrow Y$ sub-graph, or the samples from it. 
We fit the \simganflow model with both strategies and show the change of log density during training in Figure~\ref{fig:samples-vs-means}
Using the means makes the training converge much faster than using the samples. 
It is not surprised at all, because using samples leads to noisy gradients during training.
Thus, for faster convergence, 
we use means rather than samples in all of our experiments, 
even though it make the estimates slightly biased.

\section{GAN-based Implementation (\simgangan)}
\label{sec:gan_impl}
Equation \ref{eq:sim_elbo} also allows for likelihood-free inference with a little manipulation. Simply by adding and subtracting the log of the true conditional density $\log p(X|Y)$ in \eqref{eq:sim_elbo}, we get the lower bound,
\begin{align}
    \label{eq:sim_elbo_gan} 
    &\log p_{\{\phi,\theta,\theta_Z,\Theta\}}(x,y) \geq - \mathcal{KL}[q_\phi(c|y) \Vert p(c)] + \nonumber \\ 
    &\int q_\phi(c|y)\log p_\theta(y|c) dc  
    - \mathcal{KL}[q_{\theta_Z}(z|x,y) \Vert p(z)] +\nonumber \\  &\underbrace{ \int q_{\theta_Z}(z|x,y)\log \frac{p_{\Theta}(x|z,y)}{p(x|y)} dz}_{\text{CGAN-based }(Y,Z) \rightarrow X} + \log p(x|y)
\end{align}

Here, the log density ratio $\log \frac{p_{\Theta}(x|z,y)}{p(x|y)}$ can be estimated using adversarial learning with a binary discriminator function \citep{sugiyama2012density,srivastava2017veegan}. In practice, since we are learning conditional distributions, the decoder is implemented as a conditional GAN (CGAN) \cite{mirza2014conditional} This implementation is referred to as \simgangan. 

\subsection{Preliminaries: Generative Adversarial Networks}
\label{subsec:CGAN}

Generative adversarial networks (GAN) \citep{gan} represent the current state of the art in likelihood-free generative modeling. In GANs, a generator network $G_\theta$ is trained to produce samples that can fool a discriminator network $D_\omega$ that is in turn trained to distinguish samples from the true data distribution $p(x)$ and the generated samples $G_\theta(z) | z \sim p_z(z)$. Here, $p_z$ is usually a low dimensional easy-to-sample distribution like standard Gaussian. A variety of tricks and techniques need to be employed to solve this min-max optimization problem. For our models, we employ architectural constraints proposed by DC-GAN \citep{dcgan} that have been widely successful in ensuring training stability and improving generated image quality.



Conditional GANs (CGAN) \citep{mirza2014conditional} adapt the GAN framework for generating class conditional samples by jointly modeling the observations with their class labels. In CGAN, the generator network $G_\theta$ is fed class labels $c$ to produce fake conditional samples and the discriminator $D_\omega$ is trained to discriminate between the samples from the joint distribution of true conditional and true labels $p(x|c)p(c)$ and the fake conditional and true labels $p_\theta(x|c)p(c)$.

\subsection{Learning $(Y,Z) \rightarrow X$ using CGAN}
Vanilla GANs can only model the marginal data distribution i.e. they learn $p_\theta$ to match $p_x$ and in doing so they use the input to the generator ($G_\theta$) only as a source of stochasticity. 
Therefore we start with a conditional GAN model instead, to preserve the correspondence between $Y$ and $X$. As shown in section \ref{subsec:CGAN}, this framework trains $G_\theta$ such that the observation $X$ is maximally explained by the conditioning variable $Y$. One major deviation from the original model is that the conditioning variable in our case is the same type and dimensionality as the observation. That is, it is an image, albeit a blurry one. This setup has previously been used by \citet{isola2017image} in the context of image-to-image translation. With that, $Z$ is incorporated using AdaIN as in the case of \simgan.

\section{Experiments for \simgangan}
In this section, we provide a comprehensive set of qualitative results to demonstrate how \simgangan is clearly able to not only disentangle $C$ from $Z$ in both supervised and unsupervised settings but also ensure that independent components of $C$ stay disentangled after training. Additionally, we show how in unsupervised settings \simgangan can be used to discover disentangled latent factors when $C$ is not explicitly provided. 

We evaluate \simgangan on a variety of image generation tasks which naturally involve observed attributes $C$ and unobserved attributes $Z$. To that end, we generate three 3D image datasets of faces, chairs, and cars with explicit control variables. Chairs and cars datasets are derived from ShapeNet \citep{shapenet}. We sample 100k images from the full yaw variation and a pitch variation of 90 degrees. We used the straight chair subcategory with 1968 different chairs and the sedan subcategory with 559 different cars. We used Blender to render the ShapeNet meshes scripted with the Stanford ShapeNet renderer. For faces, we generated 100k images from the Basel Face Model 2017 \citep{bfm17}. We sample shape and color (first 50 coefficients), expressions (first 5 coefficients), pose (yaw -90 to 90 degrees uniformly, pitch and roll according to a Gaussian with variance of 5 degrees) and the illumination from the Basel Illumination Prior \citep{Egger2018IJCV}. For the generation of the faces dataset, we use the software provided by \citet{kortylewski2019analyzing}. For the stated datasets we have complete access to $C$, however, we also include unsupervised results on celebA \citep{liu2015faceattributes} with unconstrained, real images. All our datasets are built from publicly available data and tools.

We use the DCGAN architecture ~\citep{dcgan} for all neural networks involved in all the experiments in this work and provide a reference implementation with exact architecture and hyperparameter settings at \textit{https://github.com/AnonymousAuthors000/DS-VAE}. 

\subsection{Supervised Setting}
\begin{figure*}[htb!]
\centering
\subfloat[\simgangan car rotation]{
  \includegraphics[height=0.84in]{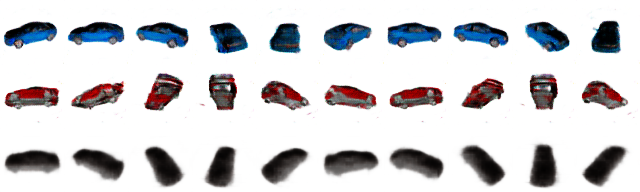}
}
\subfloat[\simgangan car elevation]{
  \includegraphics[height=0.84in]{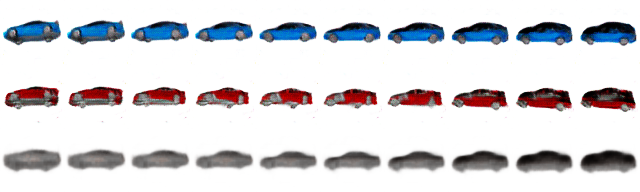}
}
\\
\subfloat[\simgangan chair rotation]{
  \includegraphics[height=0.84in]{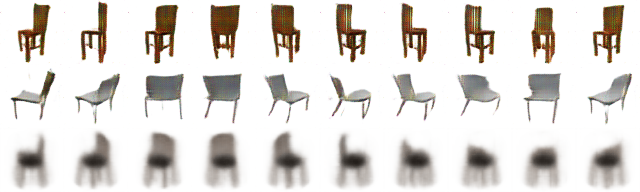}
}
\subfloat[\simgangan chair elevation]{
  \includegraphics[height=0.84in]{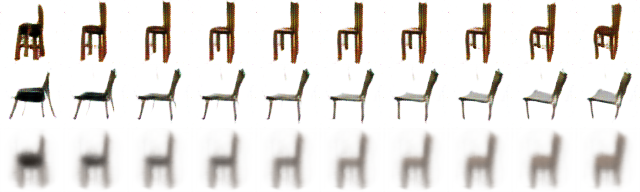}
}
\\
\subfloat[CGAN car rotation]{
  \includegraphics[height=0.56in]{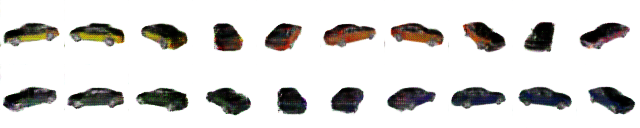}
}
\subfloat[CGAN car elevation]{
  \includegraphics[height=0.56in]{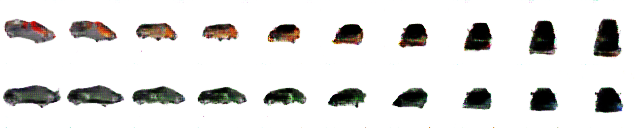}
}
\\
\subfloat[CGAN chair rotation]{
  \includegraphics[height=0.56in]{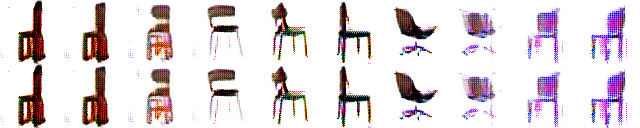}
}
\subfloat[CGAN chair elevation]{
  \includegraphics[height=0.56in]{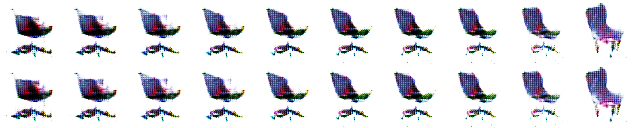}
}
\caption{Latent traversal on cars and chairs. Third rows in \simgangan results show $Y$.} 
\label{fig:car_chair}

\end{figure*}

In the supervised setting we compare \simgangan to CGAN qualitatively. To evaluate the level of disentanglement between $C$ and $Z$, we vary each individual dimension of $C$ over its range while holding $Z$ constant. We plot the generated images for both models on car and chair datasets in Figure \ref{fig:car_chair}. Notice that \simgangan allows us to vary the control variates without changing the identity of the object, whereas CGAN does not. In addition, we find that for CGAN, the noise $Z$ provides little to no control over the identity of the chairs. This is potentially due to the internal stochasticity introduced by the BatchNorm. The last rows for the \simgangan figures provide the visualization of $Y$. It can be seen how $Y$ is clearly preserving $C$ (pose information) but averaging the identity related details. 

We also qualitatively evaluate \simgangan on the more challenging faces dataset that includes 10 control variates. As shown in Figure \ref{fig:face_pose_color} in the appendix, \simgangan is not only able to model the common pose factors such as rotation and azimuth but also accurately captures the principal shape component of Basel face model that approximates the width of the forehead, the width of jaw etc. Compared to CGAN, \simgangan does a qualitatively better job at keeping the identity constant. 

\subsection{Unsupervised Setting}
We now test the performance of \simgangan in the unsupervised setting, where disentangled components of $C$ needs to be discovered, using $\beta$-VAE, as part of learning the mapping $C \rightarrow Y$. For our purpose, we use a simple version of the original $\beta$-VAE method with a very narrow bottleneck (6D for faces and 2D for cars and chairs) to extract $C$. 

\begin{figure}[h!]
\begin{center}
\includegraphics[height=.9in]{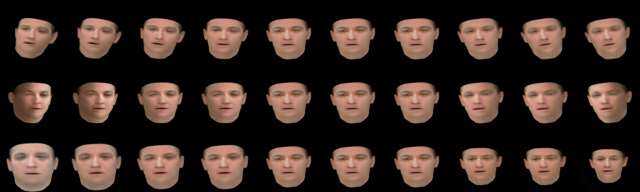}
\end{center}
\caption{Latent traversal on faces (unsupervised \simgangan). The three latent variables capture the rotation, azimuth, and distance respectively.}
\label{fig:unsupervised_face}
\end{figure}

\begin{figure}[h!]
\begin{center}
\includegraphics[height=.9in]{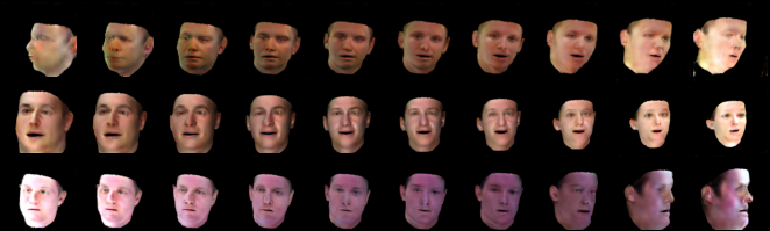}
\end{center}
\caption{Latent traversal of InfoGAN on faces. The latent variables are able to capture some pose changes but the pose changes are highly entangled with other pose factors as well as the face shape.}
\label{fig:InfoGAN}
\end{figure}

\begin{figure}[]
\begin{center}
\includegraphics[height=1.5in]{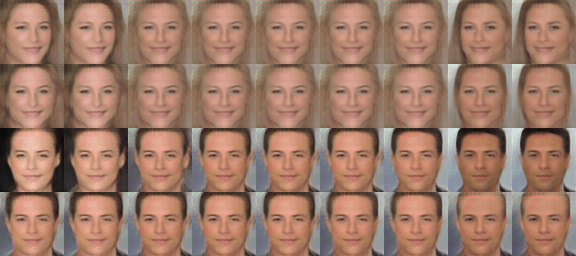}
\end{center}
\caption{Latent traversal on CelebA (unsupervised \simgangan). The latent variables consistently capture the azimuth, hair-style, gender and hair color respectively while maintaining good image quality.}
\label{fig:CelebA}
\end{figure}

The latent traversals for the faces dataset are presented in Figure \ref{fig:unsupervised_face}. Unsupervised discovery is able to recover rotation as well as translation variation present in the dataset. For comparison, we evaluate InfoGAN \citep{info} and present the results in Figure \ref{fig:InfoGAN} where it is evident that \simgangan clearly outperforms InfoGAN on both disentanglement and generative quality. More traversal results are provided in the appendix. 
We further test our method on the CelebA dataset \citep{liu2015faceattributes}, where pose information is not available. This traversal plot is shown in Figure \ref{fig:CelebA}. Traversal plots for cars and chairs dataset are provided in the Figures \ref{fig:unsupervised_old_chair} and \ref{fig:unsupervised_car}.


\subsection{Additional experiment results}

\begin{figure}[htb!]
\centering
  \subfloat[\simgangan face pose]{
  \begin{minipage}[b]{0.45\textwidth}
  \label{fig:P1}
  \includegraphics[width=.95\textwidth,height=1.8in]{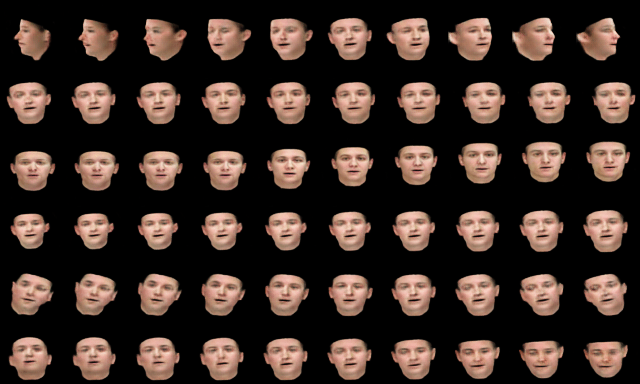}
  \end{minipage}
  }
  \\
  \subfloat[CGAN face pose]{
  \centering
  \begin{minipage}[b]{0.45\textwidth}
  \centering
  \label{fig:P2}
  \includegraphics[width=.95\textwidth,height=1.8in]{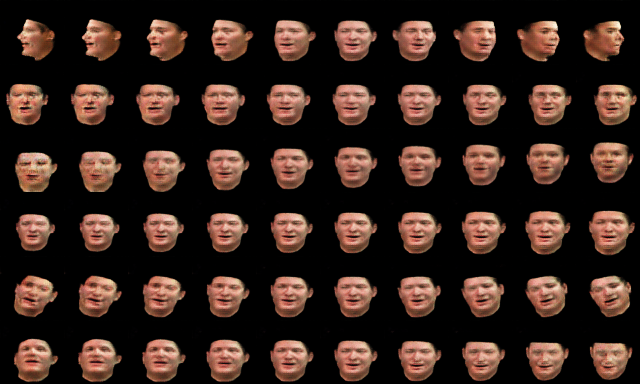}
  \end{minipage}
  }
  \\
  \subfloat[\simgangan face shape]{
  \begin{minipage}[b]{0.45\textwidth}
  \label{fig:P3}
  \includegraphics[width=.95\textwidth, height=1.2in]{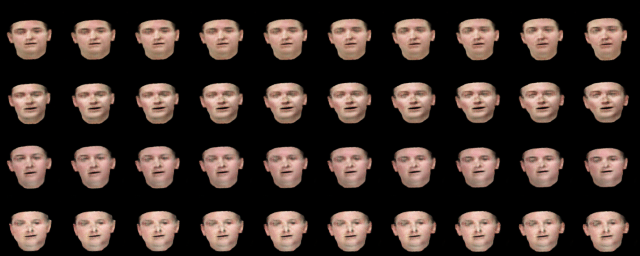}
  \end{minipage}
  }
  \\
  \subfloat[CGAN face shape]{
  \centering
  \begin{minipage}[b]{0.45\textwidth}
  \centering
  \label{fig:P4}
  \includegraphics[width=.95\textwidth, height=1.2in]{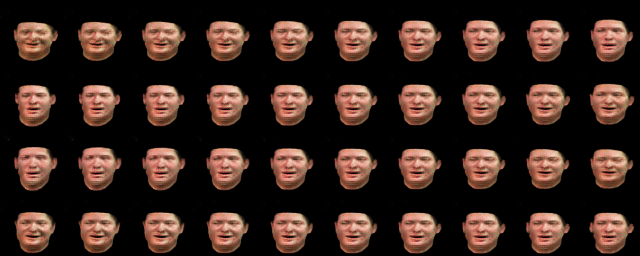}
  \end{minipage}
  }
\caption{Latent traversal of \simgangan and CGAN on faces. The pose variations are azimuth, horizontal translation, vertical translation, distance, rotation, and elevation from top to bottom. The shape variations show the difference in face height, forehead, jaw, and ear from top to bottom.}
\label{fig:face_pose_color}
\end{figure}

\begin{figure}[h] 

  \centering
  \begin{minipage}[b]{0.5\textwidth}
  \centering
  \label{fig:P1}
  \includegraphics[width=.95\textwidth, height=1.2in]{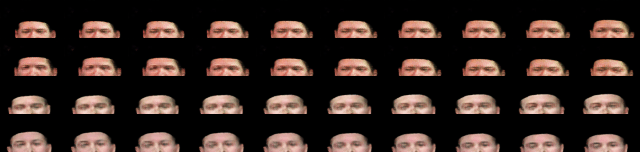}
  \end{minipage}
\caption{Zoomed-in comparison on face shape. Row 1: CGAN forehead variation; Row 2: CGAN jaw variation; Row 3: \simgangan forehead variation; Row 4: \simgangan jaw variation. Row 1 and Row 3 should have a bigger forehead from left to right while Row 2 and Row 4 should have a consistent forehead. CGAN and \simgangan show good forehead variation in Row 1 and Row 3, respectively, but \simgangan is better at keeping the forehead the same while another factor is changing (Row 4 vs. Row 1).}
\label{fig:zoomedin}
\end{figure}

Comparison of \simgangan and CGAN on face dataset is shown in Figure \ref{fig:face_pose_color}. CGAN not only produces blurry faces but also shows more undesired identity changes. In order to show the shape variation clearly, we provide a zoomed-in view in Figure \ref{fig:zoomedin}.

\begin{figure}[h]
  \centering
  \begin{minipage}[b]{0.5\textwidth}
  \centering
  \includegraphics[width=.95\textwidth,height=1.8in]{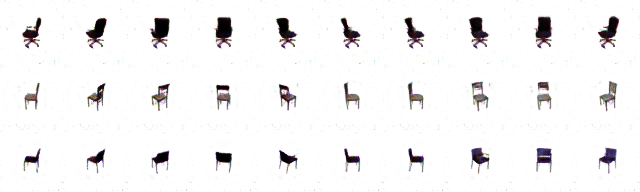}
  \includegraphics[width=.95\textwidth,height=0.6in]{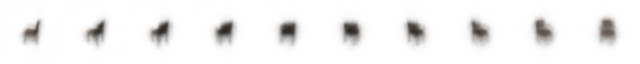}
  \end{minipage}
\caption{Latent traversal on chairs of \simgangan. The first three rows show the effect of the variable in $C$ that controls rotation. The last row is the corresponding $Y$.}
\label{fig:old_chair}
\end{figure}

\begin{figure}[hbt!]

  \centering
  \begin{minipage}[b]{0.45\textwidth}
  \centering
  \includegraphics[width=.95\textwidth,height=1.8in]{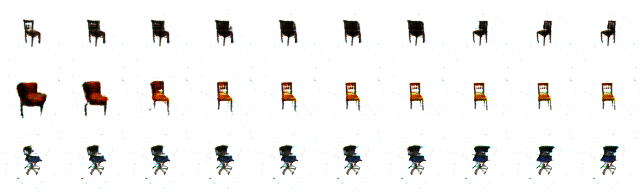}
  \end{minipage}
\caption{Latent traversal on chairs of unsupervised \simgangan.}
\label{fig:unsupervised_old_chair}
\end{figure}

We provide additional results for supervised and unsupervised results on the chair dataset from ~\citet{aubry2014seeing} in Figure \ref{fig:old_chair} and Figure \ref{fig:unsupervised_old_chair} respectively. The observation is the same with the previous one. \simgangan varies the control variables without changing the shape of chairs. In the first row in Figure \ref{fig:old_chair}, the leg of the chairs are visually indistinguishable showing an excellent disentanglement between $C$ and $Z$. For the results in unsupervised setting showing in Figure \ref{fig:unsupervised_old_chair}, \simgangan is able to disentangle the rotation of chairs without any label.

\begin{figure}[h]

  \centering
  \begin{minipage}[b]{0.5\textwidth}
  \centering
  \includegraphics[width=.95\textwidth,height=1.8in]{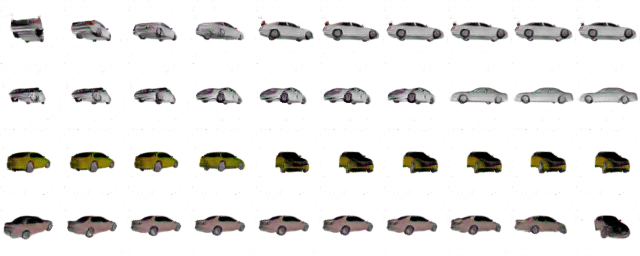}
  \end{minipage}
\caption{Latent traversal of unsupervised \simgangan on cars.}
\label{fig:unsupervised_car}
\end{figure}

Additional results of latent traversal of \simgangan in the unsupervised setting is provided in Figure \ref{fig:unsupervised_car}. The model is able capture the rotation but the translation is not very smooth.

\begin{figure}[]
  \centering
  \begin{minipage}[b]{0.5\textwidth}
  \centering
  \includegraphics[width=.95\textwidth,height=.60in]{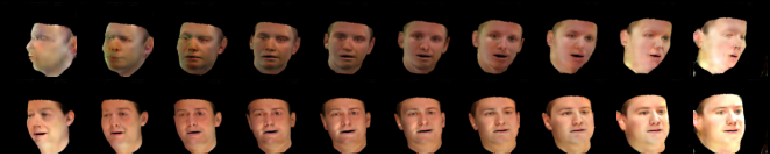}\\
  \includegraphics[width=.95\textwidth,height=.60in]{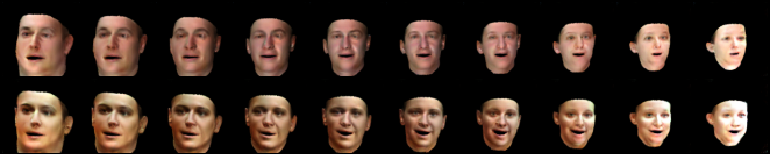}\\
  \includegraphics[width=.95\textwidth,height=.60in]{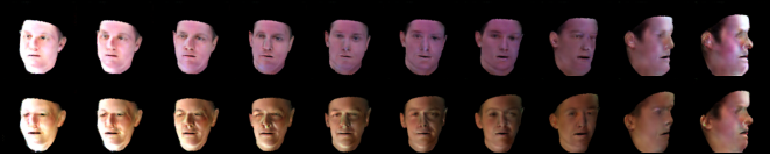}\\
  \includegraphics[width=.95\textwidth,height=.60in]{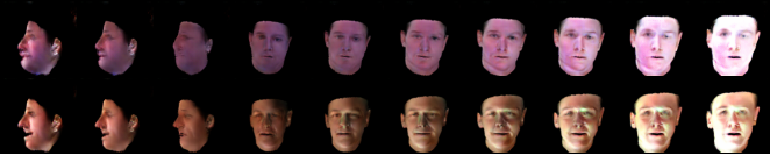}
  \end{minipage}
\caption{Latent traversal of InfoGAN on faces dataset.}
\label{fig:InfoGAN_4_variables}
\end{figure}

\begin{figure}[]
  \centering
  \begin{minipage}[b]{0.5\textwidth}
  \centering
  \label{fig:P1}
  \includegraphics[width=.95\textwidth,height=1.8in]{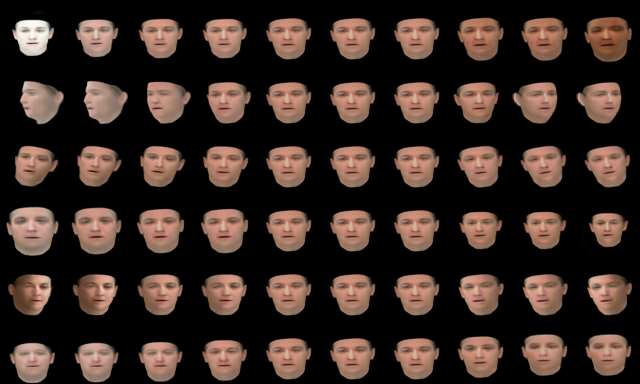}
  \end{minipage}
\caption{Latent traversal of unsupervised \simgangan on face dataset.}
\label{fig:unsupervised_face_all}
\end{figure}

Figure \ref{fig:InfoGAN_4_variables} provides the InfoGAN result on the face dataset. Compared with unsupervised \simgangan result in Figure \ref{fig:unsupervised_face_all}, clearly InfoGAN discovers some control variables but the effect is highly entangled.

\section{Metric Formulation}
\label{sec:mig_desc}
The normalized mutual information (MI) reported in  Figure \ref{fig:cars}(iii) and Figure \ref{fig:sn}(iii) is calculated as follows.

\begin{equation}\label{eq:mig}
MI[C_R,C]=\frac{1}{n}\sum_{i=1}^n \frac{ I(c^{i};c^i_{R})}{H(c^i_{R})}
\end{equation}
where $n$ is the dimension of $C$ and $C_R$ and $I(c^{i};c^i_{R})$ is the mutual information between $c^{i}$ and $c^i_{R}$.

For every ground truth factor, we compute its mutual information with each one of the learned dimensions of $C$, normalized by its entropy. For each factor, we take the difference, in mutual information, between the top two learned variables with which it has the highest mutual information. MIG is defined to be the average of that value over all ground truth factors.



\section{Comparison of \simgan and Big-$\beta$-TCVAE}
\label{sec:big-tcvae}
Table~\ref{tab:BigTCVAE} shows the comparison of \simgan with Big-$\beta$-TCVAE which is a $\beta$-TCVAE that has the same capacity in terms of the number of parameters and the latent dimensionality. Our experiments reveal that \simgan outperforms the Big-$\beta$-TCVAE in terms of both disentanglement representation learning quality and reconstruction quality.

\begin{table*}[htb!]
\caption[]{Comparion of \simgan with Big $\beta$-TCVAE.}
\centering
\begin{tabular}{lccc}
\toprule
Dataset & Model &
FID & MIG \\
\midrule
\multirow{1}{*}[-7pt]{\small{Cars}}  & \small{Big-$\beta$-TCVAE} &
35.24$\pm$1.28 & 0.09$\pm$0.04 \\
\cline{2-4}
& \small{\simgan} &
\textbf{22.72$\pm$3.66} & \textbf{0.10$\pm$0.03} \\
\midrule
\multirow{1}{*}[-7pt]{\small{\textit{Small}NORB}} & \small{Big-$\beta$-TCVAE} &
152.19$\pm$2.68 & 0.16$\pm$0.02 \\
\cline{2-4}
& \small{\simgan} &
\textbf{59.16$\pm$4.52} & \textbf{0.22$\pm$0.01} \\
\bottomrule
\end{tabular}
\label{tab:BigTCVAE}
\end{table*}

\section{Sampling Multiple C and Y during Training}
In our quantitative experiments for \simgan, the paired training set $\mathcal{D}_{\text{paired}}$ was constructed as follows: $x^i \sim p(x)$, c $\sim q_\phi(c|x)$, $y^i = \mathbb{E}_{y}[p_\theta(y|c)]$. In this construction, only one $c$ and, consequentially, only one $y$ is sampled for training. For the qualitative experiments with celebA, we sampled 10 different $c$'s from the $\beta$-TCVAE for each $x_i$ such that the final dataset had 10 different $y_i$'s for each $x_i$. 

\begin{figure*}[!h]
	\centering
	\includegraphics[scale=.25]{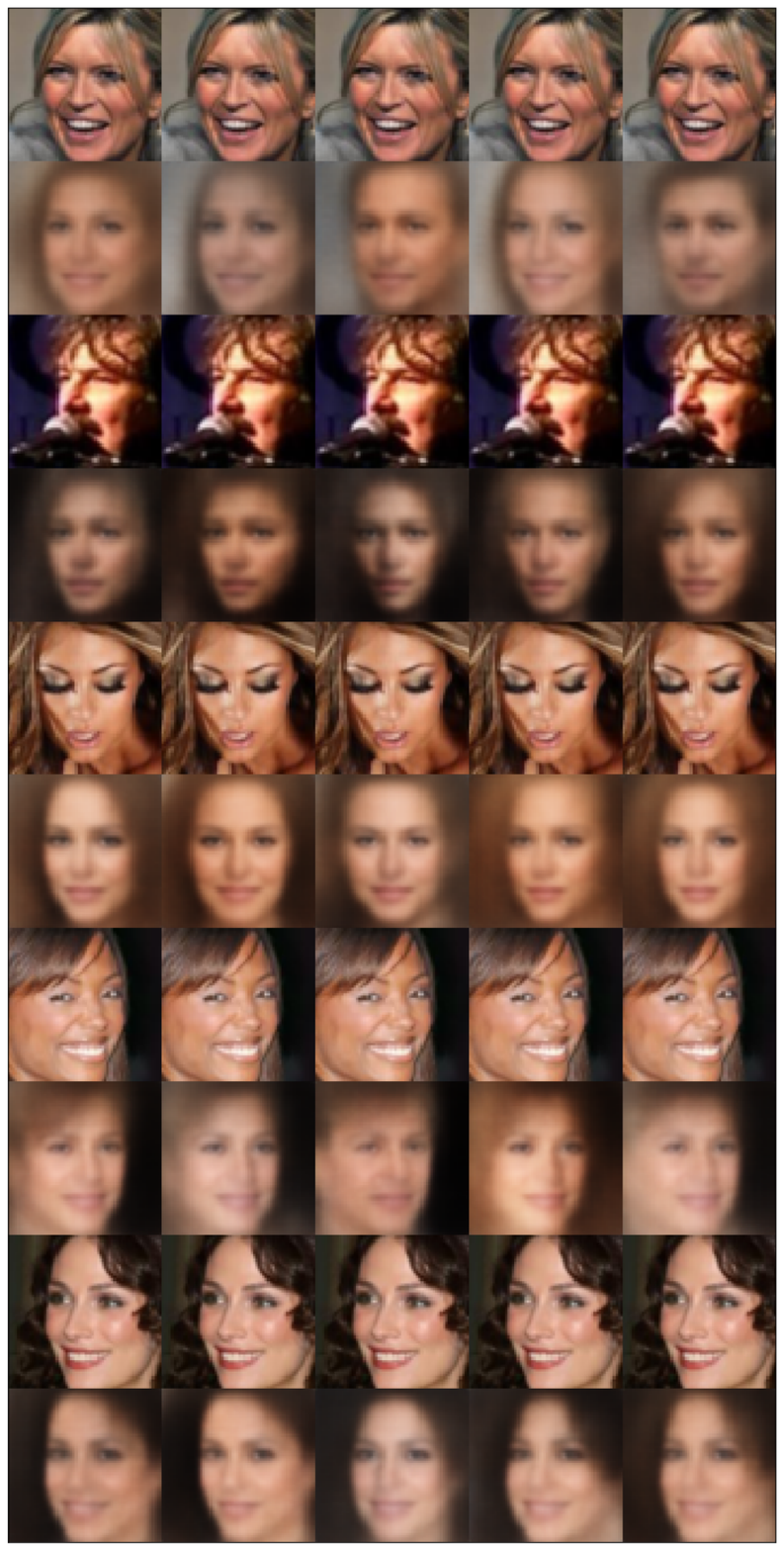}
	\caption{Even rows are samples from CelebA (the same sample is plotted five times) and odd rows are corresponding samples from the $\beta$-TCVAE. As can be seen, for the same celebA image, the reconstruction from the $\beta$-TCVAE is highly variable, illustrating the high variance of the learned posterior for $\beta$-TCVAE.}
	\label{fig:tcvae_var}
\end{figure*}

\begin{figure*}[]
	\centering
	\includegraphics[scale=.4]{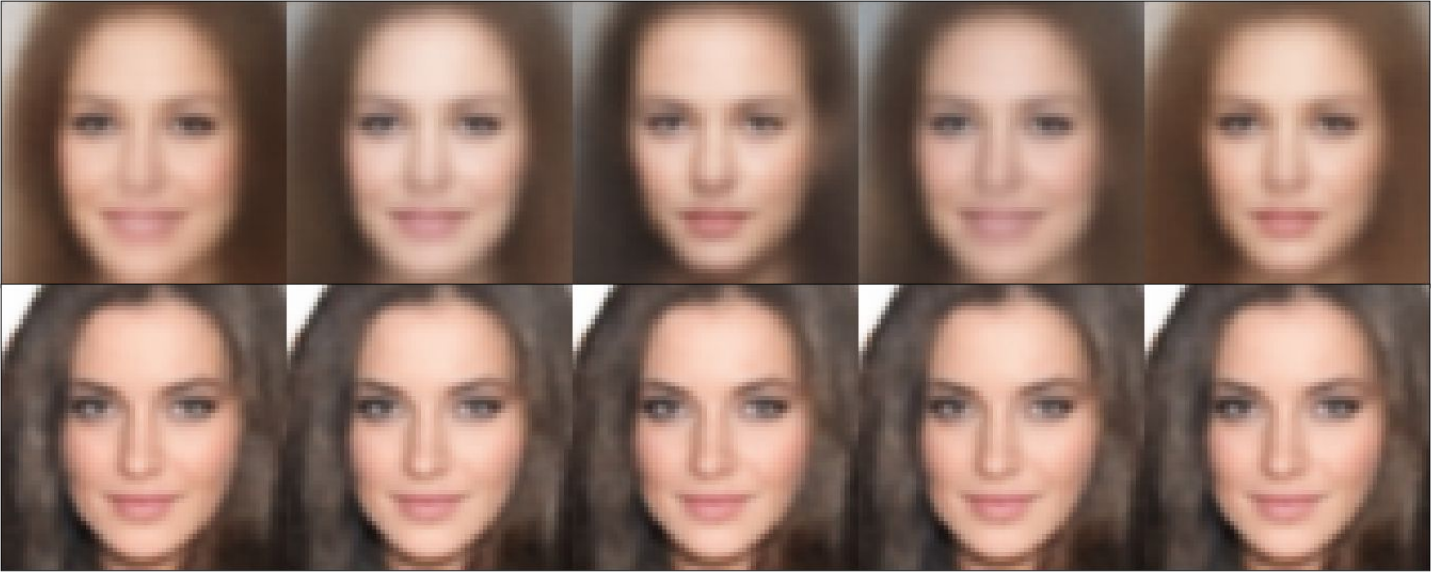}
	\caption{The top row is the same sample from the $\beta$-TCVAE and the bottom row is the corresponding reconstruction for \simgan. As can be seen in the figure, despite variability in the sampled Y's, \simgan is able to reconstruct the celebA images with little variance.}
	\label{fig:sampled_reconstruction}
\end{figure*}

The reason we sampled more $y_i$'s for each $x_i$ is that there was high variability of the sampled $y_i$'s given the same $x_i$. To visualize this effect, we plot five Y samples for five X samples from a trained $\beta$-TCVAE ($\beta=15, C=15$) in figure \ref{fig:tcvae_var}. As can be seen from in the figure, there is high variability in the sampled $y_i$'s for each $x_i$, highlighting the the high variance of $\beta$-TCVAE's posterior distribution over C. Training with multiple $y_i$'s for each $x_i$ allows the \simgan to naturally adapt to this variability as shown in \ref{fig:sampled_reconstruction}. We plan to explore this effect in future work.


\section{Pixel-wise Reconstruction Loss}
\label{sec:pw_recon}
\begin{figure*}[htb!]
\centering
\subfloat[\textit{Cars3D} dataset]{%
  \centering
  \label{fig:cars}
    \scalebox{.5}{
\begin{tikzpicture}

\definecolor{color0}{rgb}{0.12156862745098,0.466666666666667,0.705882352941177}
\definecolor{color1}{rgb}{1,0.498039215686275,0.0549019607843137}
\definecolor{color2}{rgb}{0.172549019607843,0.627450980392157,0.172549019607843}
\definecolor{color3}{rgb}{0.83921568627451,0.152941176470588,0.156862745098039}
\definecolor{color4}{rgb}{0.580392156862745,0.403921568627451,0.741176470588235}
\definecolor{color5}{rgb}{0.549019607843137,0.337254901960784,0.294117647058824}

\begin{axis}[
legend cell align={left},
legend style={fill opacity=0.8, draw opacity=1, text opacity=1,
    /tikz/every even column/.append style={column sep=0.5cm},
    at={(-0.15,1.08)}, anchor=south west, draw=white!80!black},
tick align=outside,
tick pos=both,
x grid style={white!69.0196078431373!black},
xlabel={\(\displaystyle \beta\)},
xmajorgrids,
xmin=0.444612903225807, xmax=10.5553870967742,
xtick style={color=black},
y grid style={white!69.0196078431373!black},
ylabel={L1},
ymajorgrids,
ymin=0.009, ymax=0.028,
ytick style={color=black}
]
\path [fill=color0, fill opacity=0.4]
(axis cs:1,0.0251425783436366)
--(axis cs:1,0.0247145490003464)
--(axis cs:2,0.0256910437825322)
--(axis cs:4,0.0236931529655395)
--(axis cs:6,0.022348218690029)
--(axis cs:8,0.0227314000004572)
--(axis cs:10,0.0265414983812229)
--(axis cs:10,0.026830446166999)
--(axis cs:10,0.026830446166999)
--(axis cs:8,0.0228049982938809)
--(axis cs:6,0.0224783414605669)
--(axis cs:4,0.0237890428312059)
--(axis cs:2,0.0260296067156647)
--(axis cs:1,0.0251425783436366)
--cycle;

\path [fill=color3, fill opacity=0.4]
(axis cs:1,0.0178447494991043)
--(axis cs:1,0.0172026063763338)
--(axis cs:2,0.0188316957126893)
--(axis cs:4,0.0152248206245515)
--(axis cs:6,0.0111439984085228)
--(axis cs:8,0.0128279906090773)
--(axis cs:10,0.0207088235270903)
--(axis cs:10,0.0209462249426108)
--(axis cs:10,0.0209462249426108)
--(axis cs:8,0.0129181293500997)
--(axis cs:6,0.0113023482074982)
--(axis cs:4,0.0157613019119694)
--(axis cs:2,0.0193152160171727)
--(axis cs:1,0.0178447494991043)
--cycle;

\addplot [only marks, mark=*, draw=color2, fill=color2, colormap/viridis]
table{%
x                      y
1 0.0184746
10 0.0198152
};
\addlegendentry{Lezama}

   \addplot [only marks, mark=*, draw=color4, fill=color4, colormap/viridis]
    table{%
    x                      y
    1 0.012
    };
    \addlegendentry{Big-VAE}

\path [fill=color1, fill opacity=0.4]
(axis cs:1,0.0215983683291771)
--(axis cs:1,0.0214756595407308)
--(axis cs:2,0.022245194151631)
--(axis cs:4,0.0237349102465112)
--(axis cs:6,0.0249844414040734)
--(axis cs:8,0.0257806773526097)
--(axis cs:10,0.0265527136071877)
--(axis cs:10,0.0268557995113913)
--(axis cs:10,0.0268557995113913)
--(axis cs:8,0.0260401119861566)
--(axis cs:6,0.0251209801250869)
--(axis cs:4,0.0238859714833833)
--(axis cs:2,0.0224111972465751)
--(axis cs:1,0.0215983683291771)
--cycle;

\addplot [semithick, color0]
table {%
1 0.0249285636719915
2 0.0258603252490984
4 0.0237410978983727
6 0.0224132800752979
8 0.022768199147169
10 0.026685972274111
};
\addlegendentry{TCVAE}

\addplot [semithick, color3]
table {%
1 0.017523677937719
2 0.019073455864931
4 0.0154930612682605
6 0.0112231733080105
8 0.0128730599795885
10 0.0208275242348505
};
    \addlegendentry{\simgan}
\addplot [semithick, color1]
table {%
1 0.0215370139349539
2 0.022328195699103
4 0.0238104408649472
6 0.0250527107645802
8 0.0259103946693831
10 0.0267042565592895
};
    \addlegendentry{TCVAE-L}
\end{axis}

\end{tikzpicture}}
}
\subfloat[\textit{Small}NORB dataset]{%
  \centering
  \label{fig:sn}
  \scalebox{.5}{
\begin{tikzpicture}

\definecolor{color0}{rgb}{0.12156862745098,0.466666666666667,0.705882352941177}
\definecolor{color1}{rgb}{1,0.498039215686275,0.0549019607843137}
\definecolor{color2}{rgb}{0.172549019607843,0.627450980392157,0.172549019607843}
\definecolor{color3}{rgb}{0.83921568627451,0.152941176470588,0.156862745098039}
\definecolor{color4}{rgb}{0.580392156862745,0.403921568627451,0.741176470588235}
\definecolor{color5}{rgb}{0.549019607843137,0.337254901960784,0.294117647058824}

\begin{axis}[
legend cell align={left},
legend style={fill opacity=0.8, draw opacity=1, text opacity=1,
    /tikz/every even column/.append style={column sep=0.5cm},
    at={(-0.15,1.08)}, anchor=south west, draw=white!80!black},
tick align=outside,
tick pos=both,
x grid style={white!69.0196078431373!black},
xlabel={\(\displaystyle \beta\)},
xmajorgrids,
xmin=0.444612903225807, xmax=10.5553870967742,
xtick style={color=black},
y grid style={white!69.0196078431373!black},
ylabel={L1},
ymajorgrids,
ymin=0.009, ymax=0.0376413895665347,
ytick style={color=black}
]
\path [fill=color0, fill opacity=0.4]
(axis cs:1,0.030451005956369)
--(axis cs:1,0.0304134585885417)
--(axis cs:2,0.0219039432666883)
--(axis cs:4,0.0192053590609785)
--(axis cs:6,0.0297271249942535)
--(axis cs:8,0.0262456490573051)
--(axis cs:10,0.0287998993500362)
--(axis cs:10,0.0289919229801853)
--(axis cs:10,0.0289919229801853)
--(axis cs:8,0.0267011858115689)
--(axis cs:6,0.0302477029903175)
--(axis cs:4,0.0195871679225618)
--(axis cs:2,0.0222573411345071)
--(axis cs:1,0.030451005956369)
--cycle;

\path [fill=color3, fill opacity=0.4]
(axis cs:1,0.0157447431040167)
--(axis cs:1,0.0155614978035768)
--(axis cs:2,0.0207915540417401)
--(axis cs:4,0.0163095320555375)
--(axis cs:6,0.0158949070928066)
--(axis cs:8,0.0198156568382318)
--(axis cs:10,0.0163721736521119)
--(axis cs:10,0.0179661330091545)
--(axis cs:10,0.0179661330091545)
--(axis cs:8,0.0213301501005616)
--(axis cs:6,0.0161912907828016)
--(axis cs:4,0.0167319384882979)
--(axis cs:2,0.0214364055183746)
--(axis cs:1,0.0157447431040167)
--cycle;

\addplot [only marks, mark=*, draw=color2, fill=color2, colormap/viridis]
table{%
x                      y
1 0.0120054832050584
10 0.0124175013235788
};
\addlegendentry{Lezama}

   \addplot [only marks, mark=*, draw=color4, fill=color4, colormap/viridis]
    table{%
    x                      y
    1 0.015
    };
    \addlegendentry{Big-VAE}

\path [fill=color1, fill opacity=0.4]
(axis cs:1,0.0196333668529459)
--(axis cs:1,0.0193141532957012)
--(axis cs:2,0.0222134865043527)
--(axis cs:4,0.0262214212877725)
--(axis cs:6,0.0285866520871552)
--(axis cs:8,0.0298994928727456)
--(axis cs:10,0.0305770224656125)
--(axis cs:10,0.0310453207804956)
--(axis cs:10,0.0310453207804956)
--(axis cs:8,0.0305252055515628)
--(axis cs:6,0.0294171148341677)
--(axis cs:4,0.026640773014331)
--(axis cs:2,0.022348094576471)
--(axis cs:1,0.0196333668529459)
--cycle;

\addplot [semithick, color0]
table {%
1 0.0304322322724553
2 0.0220806422005977
4 0.0193962634917701
6 0.0299874139922855
8 0.026473417434437
10 0.0288959111651107
};
\addlegendentry{TCVAE}

\addplot [semithick, color3]
table {%
1 0.0156531204537968
2 0.0211139797800573
4 0.0165207352719177
6 0.0160430989378041
8 0.0205729034693967
10 0.0171691533306332
};
    \addlegendentry{\simgan}
\addplot [semithick, color1]
table {%
1 0.0194737600743236
2 0.0222807905404118
4 0.0264310971510518
6 0.0290018834606615
8 0.0302123492121542
10 0.0308111716230541
};
    \addlegendentry{TCVAE-L}
\end{axis}

\end{tikzpicture}}
}
\caption{\label{fig:l1} L1 (lower is better) comparison for $\beta$-TCVAE ($C \rightarrow Y$), $\beta$-TCVAE-L (latent dimensionality same as $C + Z$), and \simgan models.}
\end{figure*}
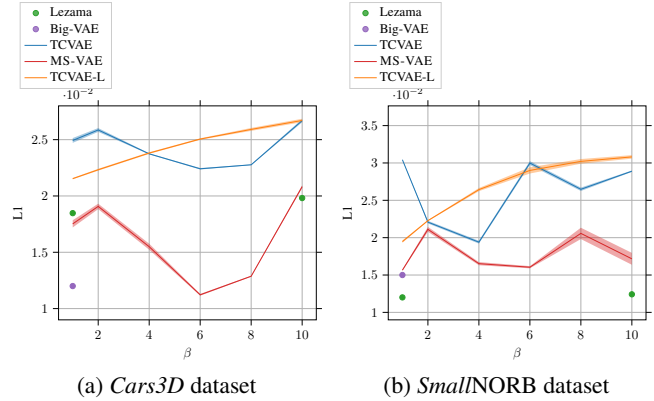

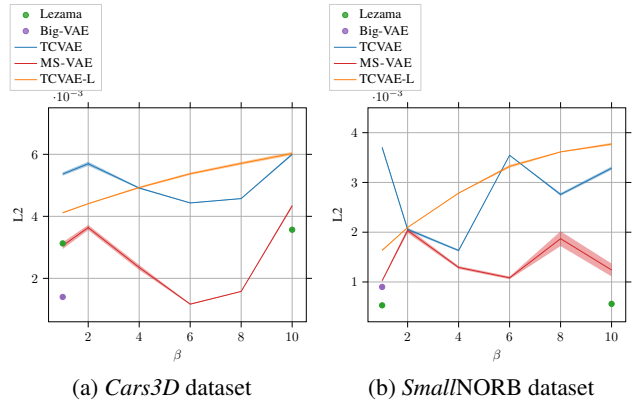
\begin{figure*}[htb!]
\centering
\subfloat[\textit{Cars3D} dataset]{%
  \centering
  \label{fig:cars}
    \scalebox{.5}{
\begin{tikzpicture}

\definecolor{color0}{rgb}{0.12156862745098,0.466666666666667,0.705882352941177}
\definecolor{color1}{rgb}{1,0.498039215686275,0.0549019607843137}
\definecolor{color2}{rgb}{0.172549019607843,0.627450980392157,0.172549019607843}
\definecolor{color3}{rgb}{0.83921568627451,0.152941176470588,0.156862745098039}
\definecolor{color4}{rgb}{0.580392156862745,0.403921568627451,0.741176470588235}
\definecolor{color5}{rgb}{0.549019607843137,0.337254901960784,0.294117647058824}

\begin{axis}[
legend cell align={left},
legend style={fill opacity=0.8, draw opacity=1, text opacity=1,
    /tikz/every even column/.append style={column sep=0.5cm},
    at={(-0.15,1.08)}, anchor=south west, draw=white!80!black},
tick align=outside,
tick pos=both,
x grid style={white!69.0196078431373!black},
xlabel={\(\displaystyle \beta\)},
xmajorgrids,
xmin=0.444612903225807, xmax=10.5553870967742,
xtick style={color=black},
y grid style={white!69.0196078431373!black},
ylabel={L2},
ymajorgrids,
ymin=0.0006, ymax=0.0075,
ytick style={color=black}
]
\path [fill=color0, fill opacity=0.4]
(axis cs:1,0.00541840409187192)
--(axis cs:1,0.00530903054000584)
--(axis cs:2,0.00562881707405459)
--(axis cs:4,0.00488849901079301)
--(axis cs:6,0.0044174811568598)
--(axis cs:8,0.00455953170914486)
--(axis cs:10,0.00596124851844446)
--(axis cs:10,0.00603940895229532)
--(axis cs:10,0.00603940895229532)
--(axis cs:8,0.00458520562365702)
--(axis cs:6,0.00444749484506035)
--(axis cs:4,0.00493723184583787)
--(axis cs:2,0.00576204299940258)
--(axis cs:1,0.00541840409187192)
--cycle;

\path [fill=color3, fill opacity=0.4]
(axis cs:1,0.0031648647325953)
--(axis cs:1,0.00294544299526324)
--(axis cs:2,0.00355344743157501)
--(axis cs:4,0.00226991467570183)
--(axis cs:6,0.00115052025867757)
--(axis cs:8,0.00156842822175)
--(axis cs:10,0.00429128725371563)
--(axis cs:10,0.00438507078291519)
--(axis cs:10,0.00438507078291519)
--(axis cs:8,0.00158846152126825)
--(axis cs:6,0.00118584141665705)
--(axis cs:4,0.00243393404364069)
--(axis cs:2,0.00372154569531742)
--(axis cs:1,0.0031648647325953)
--cycle;

\addplot [only marks, mark=*, draw=color2, fill=color2, colormap/viridis]
table{%
x                      y
1 0.00312710308096374
10 0.00356809532941452
};
\addlegendentry{Lezama}

   \addplot [only marks, mark=*, draw=color4, fill=color4, colormap/viridis]
    table{%
    x                      y
    1 0.0014
    };
    \addlegendentry{Big-VAE}

\path [fill=color1, fill opacity=0.4]
(axis cs:1,0.00413564851585966)
--(axis cs:1,0.00409334137358855)
--(axis cs:2,0.0043942398531705)
--(axis cs:4,0.00489764386320493)
--(axis cs:6,0.00533449148247089)
--(axis cs:8,0.00566041051371536)
--(axis cs:10,0.00597297645917918)
--(axis cs:10,0.00608015385595197)
--(axis cs:10,0.00608015385595197)
--(axis cs:8,0.00575548440860788)
--(axis cs:6,0.00541148272515356)
--(axis cs:4,0.00495819838956442)
--(axis cs:2,0.00442228338312835)
--(axis cs:1,0.00413564851585966)
--cycle;

\addplot [semithick, color0]
table {%
1 0.00536371731593888
2 0.00569543003672858
4 0.00491286542831544
6 0.00443248800096007
8 0.00457236866640094
10 0.00600032873536989
};
\addlegendentry{TCVAE}

\addplot [semithick, color3]
table {%
1 0.00305515386392927
2 0.00363749656344622
4 0.00235192435967126
6 0.00116818083766731
8 0.00157844487150913
10 0.00433817901831541
};
    \addlegendentry{\simgan}
\addplot [semithick, color1]
table {%
1 0.0041144949447241
2 0.00440826161814943
4 0.00492792112638467
6 0.00537298710381222
8 0.00570794746116162
10 0.00602656515756557
};
    \addlegendentry{TCVAE-L}
\end{axis}

\end{tikzpicture}}
}
\subfloat[\textit{Small}NORB dataset]{%
  \centering
  \label{fig:sn}
  \scalebox{.5}{
\begin{tikzpicture}

\definecolor{color0}{rgb}{0.12156862745098,0.466666666666667,0.705882352941177}
\definecolor{color1}{rgb}{1,0.498039215686275,0.0549019607843137}
\definecolor{color2}{rgb}{0.172549019607843,0.627450980392157,0.172549019607843}
\definecolor{color3}{rgb}{0.83921568627451,0.152941176470588,0.156862745098039}
\definecolor{color4}{rgb}{0.580392156862745,0.403921568627451,0.741176470588235}
\definecolor{color5}{rgb}{0.549019607843137,0.337254901960784,0.294117647058824}

\begin{axis}[
legend cell align={left},
legend style={fill opacity=0.8, draw opacity=1, text opacity=1,
    /tikz/every even column/.append style={column sep=0.5cm},
    at={(-0.15,1.08)}, anchor=south west, draw=white!80!black},
tick align=outside,
tick pos=both,
x grid style={white!69.0196078431373!black},
xlabel={\(\displaystyle \beta\)},
xmajorgrids,
xmin=0.444612903225807, xmax=10.5553870967742,
xtick style={color=black},
y grid style={white!69.0196078431373!black},
ylabel={L2},
ymajorgrids,
ymin=0.0002, ymax=0.0045,
ytick style={color=black}
]
\path [fill=color0, fill opacity=0.4]
(axis cs:1,0.00373343438938261)
--(axis cs:1,0.00367540157924241)
--(axis cs:2,0.00202519969971657)
--(axis cs:4,0.00161343269780477)
--(axis cs:6,0.00354231204423265)
--(axis cs:8,0.00272991293205568)
--(axis cs:10,0.00325146686304112)
--(axis cs:10,0.00332165779802038)
--(axis cs:10,0.00332165779802038)
--(axis cs:8,0.0027862656910604)
--(axis cs:6,0.00355029589058703)
--(axis cs:4,0.00165024810649564)
--(axis cs:2,0.00208758718358855)
--(axis cs:1,0.00373343438938261)
--cycle;

\path [fill=color3, fill opacity=0.4]
(axis cs:1,0.00103085241541021)
--(axis cs:1,0.00101639800824543)
--(axis cs:2,0.00198386426877607)
--(axis cs:4,0.00126170194695016)
--(axis cs:6,0.00105428753950067)
--(axis cs:8,0.00172066263190762)
--(axis cs:10,0.00111139402487632)
--(axis cs:10,0.00137063398867317)
--(axis cs:10,0.00137063398867317)
--(axis cs:8,0.00201888172183045)
--(axis cs:6,0.00111137729698355)
--(axis cs:4,0.00132112701122083)
--(axis cs:2,0.00208751619334694)
--(axis cs:1,0.00103085241541021)
--cycle;

\addplot [only marks, mark=*, draw=color2, fill=color2, colormap/viridis]
table{%
x                      y
1 0.000530235908940599
10 0.000560034956825866
};
\addlegendentry{Lezama}

   \addplot [only marks, mark=*, draw=color4, fill=color4, colormap/viridis]
    table{%
    x                      y
    1 0.0009
    };
    \addlegendentry{Big-VAE}

\path [fill=color1, fill opacity=0.4]
(axis cs:1,0.00164790756097218)
--(axis cs:1,0.00162144488754575)
--(axis cs:2,0.002094614495979)
--(axis cs:4,0.00277799660159743)
--(axis cs:6,0.00329310226775147)
--(axis cs:8,0.00360366211166251)
--(axis cs:10,0.00374487066044896)
--(axis cs:10,0.00379725828863017)
--(axis cs:10,0.00379725828863017)
--(axis cs:8,0.00362405789458795)
--(axis cs:6,0.00335307978818267)
--(axis cs:4,0.0027929808690542)
--(axis cs:2,0.00209677856064793)
--(axis cs:1,0.00164790756097218)
--cycle;

\addplot [semithick, color0]
table {%
1 0.00370441798431251
2 0.00205639344165256
4 0.0016318404021502
6 0.00354630396740984
8 0.00275808931155804
10 0.00328656233053075
};
\addlegendentry{TCVAE}

\addplot [semithick, color3]
table {%
1 0.00102362521182782
2 0.00203569023106151
4 0.00129141447908549
6 0.00108283241824211
8 0.00186977217686904
10 0.00124101400677475
};
    \addlegendentry{\simgan}
\addplot [semithick, color1]
table {%
1 0.00163467622425896
2 0.00209569652831347
4 0.00278548873532582
6 0.00332309102796707
8 0.00361386000312523
10 0.00377106447453956
};
    \addlegendentry{TCVAE-L}
\end{axis}

\end{tikzpicture}}
}
\caption{\label{fig:l2} L2 (lower is better) comparison for $\beta$-TCVAE ($C \rightarrow Y$), $\beta$-TCVAE-L (latent dimensionality same as $C + Z$), and \simgan models.}
\end{figure*}

For all the methods, we report the L1 and L2 reconstruction errors. As shown in Figures \ref{fig:l1} and \ref{fig:l2}, pixel-wise reconstruction loss can be misleading. In the case of SmallNorb, where \simgan clearly has significantly better image fidelity (as determined by the reconstruction FIDs and traversal plots), the L1 and L2 error are only slightly better than TCVAE. 

\section{D-Separation in \simgan: An Illustrative Example}
\label{sec:illustrative_example}
We provide a constructive example of the graphical model of \simgan shown in Figure~\ref{fig:ds_vae} to illustrate the d-separation of $Y$ as discussed in Section~\ref{sec:d-separate}.

\begin{figure*}[htb!]
\centering
\subfloat[$Y$]{
  \centering
  \includegraphics{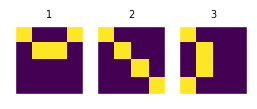}
  \label{fig:example-y}
}\\
\subfloat[$Z=0$]{
  \centering
  \includegraphics{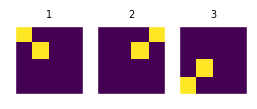}
  \label{fig:example-z0}
}
\subfloat[$Z=1$]{
  \centering
  \includegraphics{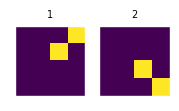}
  \label{fig:example-z1}
}
\\
\subfloat[$X$]{
  \centering
  \includegraphics{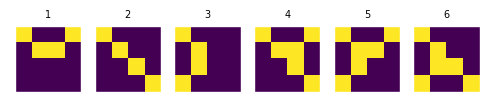}
  \label{fig:example-x}
}
\caption{A constructive example to illustrate d-separation in \simgan.}
\label{fig:example}
\end{figure*}

Suppose we are given a dataset as shown in Figure~\ref{fig:example-x} and we have trained a (disentanglement) model ($C \rightarrow Y$) where $C \sim \mathcal{U}\text{niform}([1,2,3])$ and $P(Y|C)$ is a deterministic mapping as shown in Figure~\ref{fig:example-y}, where the indices above each image are the corresponding $C$s.
As it can be seen, only a part of dataset (first three images) are modeled by the generative process of $C -> Y$.
Now, with $Y$ generated, we add details to recover back the entire dataset.
We first sample $Z \sim \mathcal{B}\text{ernoulli}(0.5)$ and add a mask to $Y$.
The mask is \textit{uniformly} sampled conditionally on $Z$ as in Figure~\ref{fig:example-z0} ($Z=0$) and Figure~\ref{fig:example-z1} ($Z=1$).
To be more specific, e.g., given that we have sampled $Z=0$, we apply a mask, uniformly chosen from Figure~\ref{fig:example-z0} to $Y$, which yield $X$.
It's not hard to check that this generative process will yield a distribution with support shown in Figure~\ref{fig:example-x}.
Also, note that $X$ and $Y$ are in the same space.
To simplify the discussion, we will represent the realizations of the random variable $X$ and $Y$ using the integer ($C$) above each image.

We want to check that $C$ and $Z$ are not conditionally independent given $X$ as implied by d-separation.
To do so, we pick $X = 4$ and check $P(C, Z | X = 4) \overset{?}{=} P(C | X = 4) P(Z | X = 4)$.
The probability distributions involved here can be represented by tables, as shown in Table~\ref{tab:example-condx}.
\begin{table}[htb!]
    \centering
    \subfloat[$P(C, Z | X = 4)$]{
        \centering
        \begin{tabular}{cc|c}
            \toprule
            C & Z & Probability \\
            \midrule
            1 & 0 & 0 \\
            1 & 1 & 3/8 \\
            2 & 0 & 1/4 \\
            2 & 1 & 3/8 \\
            3 & 0 & 0 \\
            3 & 1 & 0 \\
            \bottomrule
        \end{tabular}
    }
    $\neq$
    \subfloat[$P(C | X = 4)$]{
        \centering
        \begin{tabular}{c|c}
            \toprule
            C & Probability \\
            \midrule
            1 & 3/8 \\
            2 & 5/8 \\
            3 & 0 \\
            \bottomrule
        \end{tabular}
    }
    $\times$
    \subfloat[$P(Z | X = 4)$]{
        \centering
        \begin{tabular}{c|c}
            \toprule
            Z & Probability \\
            \midrule
            0 & 1/4 \\
            1 & 3/4 \\
            \bottomrule
        \end{tabular}
    }
    \caption{Conditioning on $X$ \textit{only}.}
    \label{tab:example-condx}
\end{table}
It can be verified that $P(C, Z | X = 4) \neq P(C | X = 4) P(Z | X = 4)$.

Now let's check the conditional independence if we \textit{additionally} condition on $Y$, e.g. $Y=1$.
Specifically, we check $P(C, Z | X = 4, Y = 1) \overset{?}{=} P(C | X = 4, Y = 1) P(Z | X = 4, Y = 1)$.
The probability distributions involved here can be represented by tables, as shown in Table~\ref{tab:example-condxy}.
\begin{table}[htb!]
    \centering
    \subfloat[$P(C, Z | X = 4, Y = 1)$]{
        \centering
        \begin{tabular}{cc|c}
            \toprule
            C & Z & Probability \\
            \midrule
            1 & 0 & 0 \\
            1 & 1 & 1 \\
            2 & 0 & 0 \\
            2 & 1 & 0 \\
            3 & 0 & 0 \\
            3 & 1 & 0 \\
            \bottomrule
        \end{tabular}
    }
    $=$
    \subfloat[$P(C | X = 4, Y = 1)$]{
        \centering
        \begin{tabular}{c|c}
            \toprule
            C & Probability \\
            \midrule
            1 & 1 \\
            2 & 0 \\
            3 & 0 \\
            \bottomrule
        \end{tabular}
    }
    $\times$
    \subfloat[$P(Z | X = 4, Y = 1)$]{
        \centering
        \begin{tabular}{c|c}
            \toprule
            Z & Probability \\
            \midrule
            0 & 0 \\
            1 & 1 \\
            \bottomrule
        \end{tabular}
    }
    \caption{Conditioning on $X$ \textit{and} $Y$.}
    \label{tab:example-condxy}
\end{table}
It can be verified that $P(C, Z | X = 4, Y = 1) = P(C | X = 4, Y = 1) P(Z | X = 4, Y = 1)$ holds.
In other words, conditioning on $X=4$ and $Y=1$ d-seperates $C$ and $Z$.

\section{Ablation Studies}
\label{app:ablation}
We now report the results of the ablation studies that we conducted to understand the contribution of: 1) Our two-step training paradigm, 2) Our use of $Y$ instead of $C$ for the second stage, 3) Our use of AdaIN to incorporate $Z$, 4) Our decision to learn $C$ before learning $Z$, in the following three subsections.
All experiments are performed on the \textit{Cars3D} dataset.

\subsection{Importance of Two-step Training}
\begin{figure*}[]
	\centering
	\includegraphics[scale=.5]{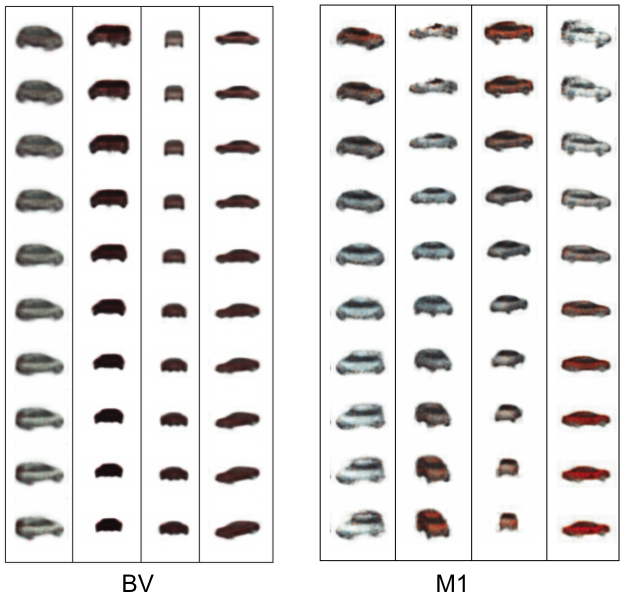}
    \caption{BV = $\beta$-VAE, M1 = $\beta$-VAE+Z. BV is trained with a dimensionality of $C=5$. M1 is trained with a dimensionality of $C=5$ and $Z=5$. For M1, the $\beta$-penalty is only applied to $C$ and $Z$ is still incorporated using AdaIN. In this figure, we show traversals of 4 of the learned disentangled factors $C$ for both BV and M1. As can be seen, traversing $C$ for BV captures different disentangled factors (e.g. azimuth, scale, elevation). Traversing $C$ for M1, however, illustrates that the disentangled factors are now entangled with other factors (e.g. identity, color).}
	\label{fig:aba1}
\end{figure*}

\simgan uses two-step training to ensure that the disentangled factors are captured in $C$ via the underlying $\beta$-VAE model and residual factors are captured in $Z$. We now show that the same will not happen if the latent space ($C$) of the underlying $\beta$-VAE is extended to incorporate $Z$ and then trained in an end-to-end fashion using the same AdaIN decoder. 
For this purpose we train a $\beta$-VAE model (BV) with the dimensionality of $C$ set to 5. As shown in Figure \ref{fig:aba1}, BV discovers disentangled factors that capture azimuth, scale and elevation. Next, we train a model (M1) which has 10 latent factors. We enforce a higher $\beta$ penalty for 5 of these latent factors ($C$) and a normal VAE $\beta$ penalty for the other 5 factors ($Z$). The contribution of $Z$ is still limited using the AdaIN decoder strucutre. As can be seen in Figure \ref{fig:aba1}, the disentangled factors $C$ in M1 are now entangled with other factors (e.g. identity, color). This experiment illustrates how end-to-end training cannot easily reproduce the two-step training scheme in \simgan which induces d-separation of $C$ and $Z$.

\subsection{Impact of using $Y$ over $C$ and AdaIN over Concatenation}
\begin{figure*}[]
	\centering
	\includegraphics[scale=.325]{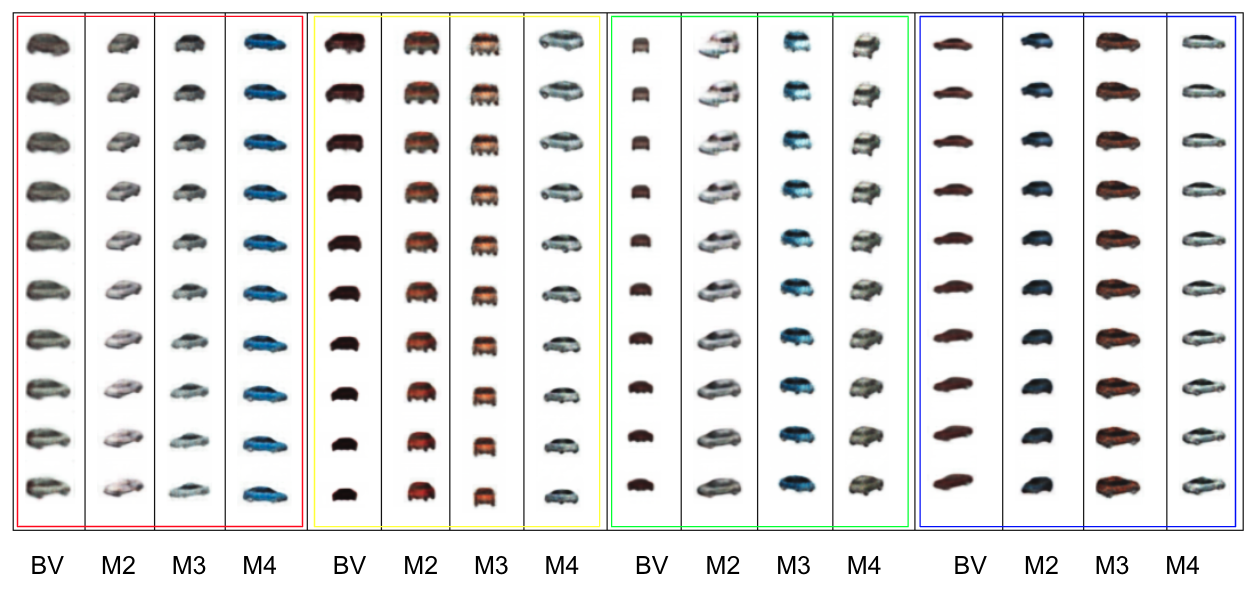}
    \caption{BV = $\beta$-VAE, M2 = \simgan-C, M3 = \simgan-C-IN, M4 = \simgan. BV is trained with a dimensionality of $C=5$. In this figure, we show traversals of 4 of the learned disentangled factors and how they are preserved in each version of \simgan (each color box is one disentangled factor). To perform traversals for each \simgan, we encode X using the $\beta$-VAE to get C and Y, then we extract Z using Y; We fix Z for the traversals of each \simgan. Clearly, M4 and M3 preserve the disentanglement from BV better than M2 (M2 changes the identity and color). Also, M4 has the best reconstruction of the three \simgan models as it utilizes Y to improve the reconstruction (rather than just using C). This ablation study illustrates that AdAIN is important for preserving disentanglement and using Y is important for improving reconstruction.}
	\label{fig:aba2}
\end{figure*}

\simgan utilizes AdaIN to improve the reconstruction $Y$ from the underlying disentangled representation learner with information from the correlated factors $Z$. To understand the importance of using AdaIN (rather than concatenation) and $Y$ (instead of $C$), we perform the following experiment: Using a $\beta$-VAE model (BV) as the $C \rightarrow Y$ sub-graph, we train three models \simgan-C (M2), \simgan-C-IN (M3) and \simgan (M4). In M2, we concatenate the inferred $C$ with $Z$ and use that as input to a DGM to reconstruct $X$. In M3, we still use the same $C$ but introduce $Z$ via AdaIN in the second stage DGM. In M4, we train \simgan as done in the paper using $Y$ and AdaIN. In Figure \ref{fig:aba2}, we show traversal plots of 4 of the disentangled factors for each of these models. For reconstruction quality, M4 outperforms both M3 and M2. This is expected as $Y$ is produced using both $C$ and the learned parameters of the first stage decoder $\theta$ and, therefore, contains much more semantic information about the observation $X$ than $C$ does. Using only $C$ requires the second stage DGM to relearn $\theta$ while also modelling the residual between $Y$ and $X$. For disentanglement, M4 and M3 both outperform M2. This illustrates how simply concatenating $Z$ and $C$ and inputting into into the second stage DGM will result in a non-linear entanglement of the two. This entanglement causes the network to fail to condition on $Y$ sufficiently and to use the entangled representation as a whole towards the reconstruction of $X$. Utilizing AdaIN to incorporate $Z$ (as done in M3 and M4), allows the network to use Z later in the generative process to model the residual information. This allows the network to better maintain the information in $Y$ (or $C$) while still incorporating $Z$.

\subsection{Ordering of $C$ and $Z$}
\begin{figure*}[]
	\centering
	\includegraphics[scale=.325]{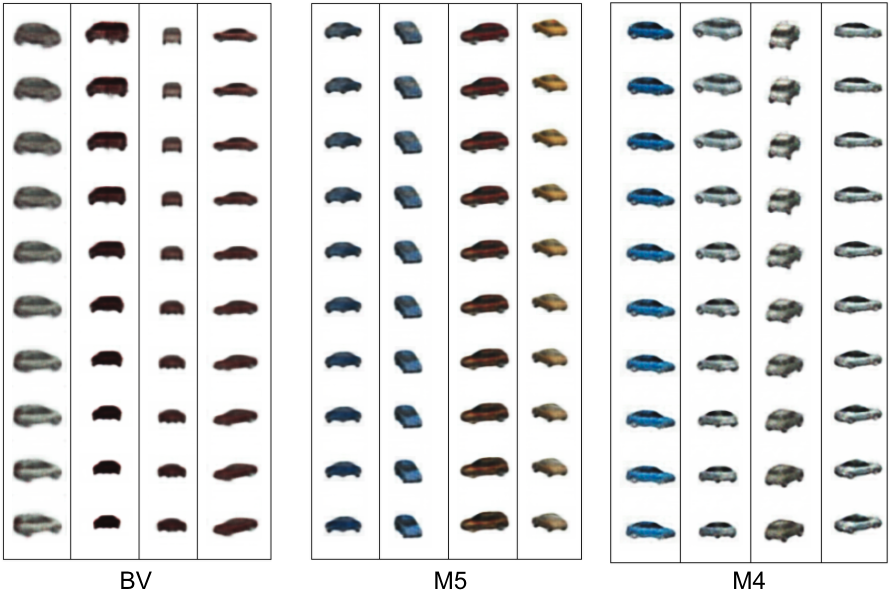}
    \caption{BV = $\beta$-VAE, M5 = \simgan-ZC, M4 = \simgan. BV is trained with a dimensionality of $C=5$. In this figure, we show traversals of 4 of the learned disentangled factors and how they are preserved in each of the three models. To perform traversals of \simgan-ZC, we encode X using a VAE to first get Z and Y, then extract C using X. To perform traversals for \simgan, we encode X using the $\beta$-VAE to get C and Y, then we extract Z using Y. We fix Z for the traversals in \simgan. Clearly, M4 preserves the disentanglement from BV better than M5 (M5 as expected, is not disentangled).}
	\label{fig:aba3}
\end{figure*}

\begin{figure*}[]
	\centering
	\includegraphics[scale=.5]{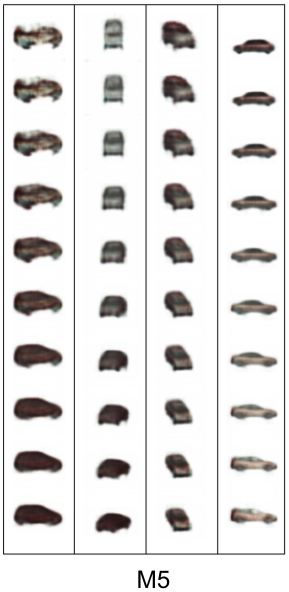}
    \caption{M5 = \simgan-ZC. In this figure, we show traversals of the $Z$ factor in \simgan-ZC model. Clearly, $Z$ is capturing factors such as scale, identity, color, rotation in an entangled fashion.}
	\label{fig:aba3_m5}
\end{figure*}

In \simgan, we first learn the disentangled factors $C$ and then we learn the residual factors $Z$. In this ablation, we evaluate how the order in which these latent variables are learned affects the final disentanglement. For this purpose, we train \simgan-ZC (M5) where, in the first stage, $Z$ is learned using a standard VAE. In the second stage DGM, the disentangled factors $C$ are learned by enforcing a KL-penalty and then concatenated with the learned $Z$ from the first stage to reconstruct $X$. In Figure \ref{fig:aba3}, we show traversals of 4 of the disentangled factors $C$. As can be seen, this new model M5 does not capture any disentangled factors in $C$. In contrast, \simgan (M4) and BV capture a variety of disentangled factors in $C$ (e.g. Azimuth, scale, elevation). We hypothesize that M5 fails to extract any meaningful disentangled factors because $Z$ already captured both the entangled and disentangled factors in the first part of the training. This is clearly shown in Figure \ref{fig:aba3_m5}. As such, the second stage will suffer from the shortcut problem where the decoder utilizes $Z$ and does not learn anything meaningful in $C$ (due to the high KL-penalty).

\label{sec:ablation}

\newpage

\section{Additional tables and plots}\label{additional}

\begin{table}[h]
    \caption{Correlation between coordinates of latent vectors produced by different models and the original input features. We include all possible correlations except very low correlations, which are less than 0.2.}
    \label{table:additional_correlation}
    \centering
    \vspace{3mm}
    \begin{tabular}{lll}
    \toprule
    Coordinate type & $\beta$-TCVAE latent correlation with input & TVAE latent correlation with input\\
    \midrule
    $z_1$ & 
        \begin{tabular}{ll} Amount Past Due & 0.218354 \\ 
                            \end{tabular}
          &
        \begin{tabular}{ll} Balance & 0.355599 \\
                            Amount Past Due & 0.534584 \\
                            Delinquency Status & 0.548507 \\
                            Credit Inquiry & 0.485710 \\
                            Open Trade & 0.426015 \\
                            Credit Inquiry\_3m & 0.339572 \\
                            Credit Inquiry\_6m & 0.457270 \\
                            Open Trade\_6m & 0.362006 \\
                            Delinquency Status\_3m & 0.566576 \\
                            Delinquency Status\_6m & 0.593434 \\
                            Delinquency Status\_12m & 0.581158 \\
                            \end{tabular} \\
    \midrule
    $z_2$ & 
        \begin{tabular}{ll} Mortgage & 0.910502 \\ \end{tabular}
        & No significant correlation \\
    \midrule 
    $z_4$ & 
        \begin{tabular}{ll} Credit Inquiry\_12m & 0.898419 \\
                            Open Trade\_12m & 0.890393 \\ 
        \end{tabular}
          & No significant correlation \\
    \midrule
    $z_5$ & 
        \begin{tabular}{ll} Credit Inquiry\_6m & 0.808494 \\
                            Open Trade\_6m & 0.832294 \\
        \end{tabular}
          & No significant correlation \\
    \midrule
    $z_6$ & 
        \begin{tabular}{ll} Utilization & 0.904766 \\
                            Balance & 0.879174 \\
                            Balance\_3m & 0.537878 \\
                            Balance\_6m & 0.428085 \\
                            Balance\_12m & 0.368590 \\
        \end{tabular}
          & No significant correlation \\
    \midrule
    $z_7$ & 
        \begin{tabular}{ll} Balance & 0.358759 \\
                            Balance\_3m & 0.758381 \\
                            Balance\_6m & 0.849974 \\
                            Balance\_12m & 0.851672 \\
        \end{tabular}
          &
        \begin{tabular}{ll} Balance\_3m & 0.285814 \\
                            Balance\_6m & 0.254979 \\
                            Balance\_12m & 0.233641 \\
        \end{tabular} \\
    \midrule
    $z_9$ & 
        \begin{tabular}{ll} Credit Inquiry & 0.503301 \\
                            Credit Inquiry\_3m & 0.573307 \\
                            Delinquency Status & 0.725198 \\
                            Delinquency Status\_3m & 0.877644 \\
                            Delinquency Status\_6m & 0.930754 \\
                            Delinquency Status\_12m & 0.928545 \\
        \end{tabular}
          &
        \begin{tabular}{ll} Credit Inquiry\_6m & 0.256394 \\
                            Open Trade\_6m & 0.209882 \\
        \end{tabular} \\
    \bottomrule
    \end{tabular}
\end{table}

\begin{figure}[h!]
\caption{JSD comparison by features: MS-VAE versus $\beta$-TCVAE}
\label{fig:js_compare1}
\centering
\includegraphics[scale=0.7]{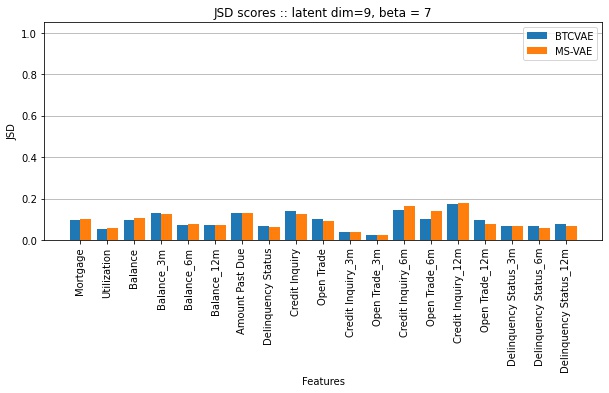}
\end{figure}

\begin{figure}[h!]
\caption{JSD comparison by features: MS-VAE versus TVAE}
\label{fig:js_compare2}
\centering
\includegraphics[scale=0.7]{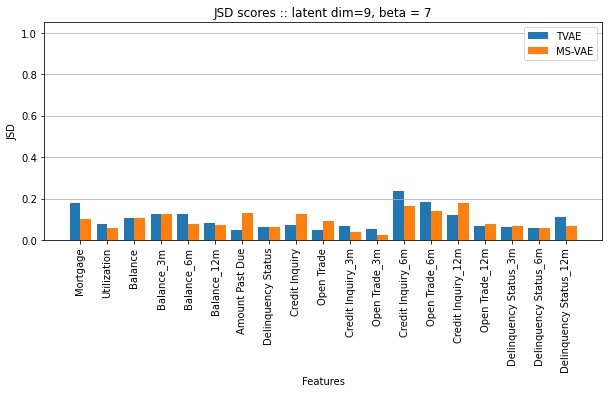}
\end{figure}

\end{document}